\documentclass{article}

\usepackage[utf8]{inputenc}
\usepackage[T1]{fontenc}
\usepackage[hyphens]{url}
\usepackage{authblk}
\usepackage{booktabs}
\usepackage[colorlinks=true, linkcolor=red, urlcolor=blue, citecolor=blue, anchorcolor=blue,backref=page]{hyperref}
\usepackage{amsfonts}
\usepackage{nicefrac}
\usepackage{microtype}
\usepackage{amsmath, amsthm, amssymb}
\usepackage{fancybox}
\usepackage{graphicx}
\usepackage{float}
\usepackage{algorithm}
\usepackage[noend]{algpseudocode}
\usepackage{color}
\usepackage{tikz}
\usepackage{pgfplots}
\usepackage{array}
\usepackage{caption}
\usepackage{subcaption}
\usepackage[textsize=scriptsize]{todonotes}
\usepackage{enumitem}
\usepackage{wrapfig}
\usepackage{enumitem}
\usepackage{etoolbox}
\usepackage{diagbox}
\usepackage{comment}
\usepackage{makecell}
\usepackage{multirow}
\usepackage{bbm, bm}
\usepackage{multicol}
\usepackage{mathpazo}

\usepackage{fullpage}
\setlist[itemize]{leftmargin=2.5em}
\setlist[itemize]{leftmargin=2.5em}
\setlist[enumerate]{leftmargin=2.5em}
\usepackage[framemethod=TikZ]{mdframed}
\mdfsetup{skipabove=5pt,
innertopmargin=0pt}

\newtheoremstyle{slplain}% name
  {.4\baselineskip\@plus.1\baselineskip\@minus.1\baselineskip}% Space above
  {.3\baselineskip\@plus.1\baselineskip\@minus.1\baselineskip}% Space below
  {\itshape}% Body font
  {}%Indent amount (empty = no indent, \parindent = para indent)
  {\bfseries}%  Thm head font
  {.}%       Punctuation after thm head
  { }%      Space after thm head: " " = normal interword space;
  {}%       Thm head spec
\theoremstyle{slplain} % italics

\newtheorem*{definition*}{Definition}
\newtheorem*{theorem*}{Theorem}
\newtheorem{theorem}{Theorem}[section]
\newtheorem{lemma}[theorem]{Lemma}
\newtheorem{proposition}[theorem]{Proposition}
\newtheorem*{unumproposition}{Proposition}

\makeatletter
\newtheorem*{rep@theorem}{\rep@title}
\newcommand{\newreptheorem}[2]{%
\newenvironment{rep#1}[1]{%
 \def\rep@title{#2 \ref{##1}}%
 \begin{rep@theorem}}%
 {\end{rep@theorem}}}
\makeatother

\newreptheorem{theorem}{Theorem}
\newreptheorem{lemma}{Lemma}
\newreptheorem{claim}{Claim}
\newreptheorem{corollary}{Corollary}
\newreptheorem{proposition}{Proposition}

\theoremstyle{definition}

\theoremstyle{plain} % choose from plain/definition/remark

\numberwithin{equation}{section}

\newcommand{\D}{\mathcal{D}}

\newcommand{\R}{\mathbb{R}}

\DeclareMathOperator{\E}{\mathbb{E}}

\renewcommand\bar\overline

\newcolumntype{C}[1]{>{\centering\let\newline\\\arraybackslash\hspace{0pt}}m{#1}}

\newcommand{\calN}{\ensuremath{\mathcal{N}}}

\newcommand{\bzero}{\ensuremath{\bm{0}}}

\newcommand{\bE}{\ensuremath{\bm{E}}}

\newcommand{\bx}{\ensuremath{\bm{x}}}

\newcommand{\bz}{\ensuremath{\bm{z}}}

\def\nd/{\textsuperscript{nd}}
\def\rd/{\textsuperscript{rd}}
\def\th/{\textsuperscript{th}}

\makeatletter
\def\nnil{\nil}
\newcounter{prob}

\newcounter{dual}

\makeatother

\newenvironment{prob*}{%
	\csname equation*\endcsname%
	\aligned%
}{%
	\endaligned%
	\csname endequation*\endcsname%
}

\usepackage{cleveref}

\title{Explicitly Encoding  Structural Symmetry is Key to Length Generalization in Arithmetic Tasks}

\author{Mahdi Sabbaghi}
\author{George Pappas}
\author{Hamed Hassani}
\author{Surbhi Goel}
\affil{University of Pennsylvania\\ \texttt{\{smahdi, pappasg, hassani, surbhig\}@seas.upenn.edu}}

\date{\today}
\begin{document}
\maketitle

\begin{abstract}
   Despite the success of Transformers on language understanding, code generation, and logical reasoning, they still fail to generalize over length on basic arithmetic tasks such as addition and multiplication. A major reason behind this failure is the vast difference in structure between numbers and text; For example, the numbers are typically parsed from right to left, and there is a correspondence between digits at the same position across different numbers. In contrast, for text, such symmetries are quite unnatural. In this work, we propose to encode these semantics explicitly into the model via modified number formatting and custom positional encodings. Empirically, our method allows a Transformer trained on numbers with at most 5-digits for addition and multiplication to generalize up to 50-digit numbers, without using additional data for longer sequences. We further demonstrate that traditional absolute positional encodings (APE) fail to generalize to longer sequences, even when trained with augmented data that captures task symmetries. To elucidate the importance of explicitly encoding structure, we prove that explicit incorporation of structure via positional encodings is necessary for out-of-distribution generalization. Finally, we pinpoint other challenges inherent to length generalization beyond capturing symmetries, in particular complexity of the underlying task, and propose changes in the training distribution to address them.
\end{abstract}

\section{Introduction}
Large language models (LLMs) \cite{brown2020language,openai2023gpt4}, powered by transformer architectures \cite{vaswani2017attention}, and extensive computational resources, have demonstrated emergent abilities in language understanding, code generation, and logical reasoning \cite{wei2022emergent}. Despite these capabilities, LLMs still struggle to generalize when handling unseen data in highly structured tasks like arithmetic \cite{dziri2023faith,liu2023exposing}. Specifically, they fail to extrapolate from shorter instances to longer ones, a challenge known as \emph{length generalization}. This issue is particularly evident in arithmetic, where models fail to generalize from smaller numbers to larger ones. \cite{nye2021,nogueira2021investigating,lee2023teaching}. 

It is now well understood that the major reasons for the failure due to structural issues are the use of positional encodings that do not scale with sequence length \cite{kazemnejad2023impact, shen2023positional}, and format of the data itself \cite{shen2023positional}. Though the Transformer can learn to solve this task for a fixed length of numbers given enough samples, the inductive bias of the architecture is insufficient to capture such structure across unseen positions, in order to generalize to longer lengths.

We posit that the challenges of length generalization in arithmetic tasks, such as addition and multiplication, can be broken down into two factors, (1) \textit{increase in complexity} of the task, (2) \textit{inability to capture the positional structure} in the tasks. For example, when we add two numbers, the positional structure corresponds to aligning each digit from \textit{right} to \textit{left} to compute the digit-wise addition, and the complexity of the task corresponds to the longest sequence of carry-overs. In contrast, when we perform multiplication of two numbers (say 1 digit number multiplied with $k$ digit number), the positional structure corresponds to aligning the first number with \textit{all} of the digits of the second number to compute digit-wise multiplication, and complexity corresponds similarly to longest sequence of induced carries that need to be resolved. 

We focus on rectifying the inability of Transformers to capture positional structure and provide strong evidence for the need to explicitly encode structural biases into the architecture for achieving length generalization. We show that with the correct data format and choice of positional encodings, a Transformer trained on only data with numbers of a bounded size (e.g., $\le 5$) can generalize to numbers of much longer length (e.g., $\ge 50$) on both addition and multiplication. We further show, empirically and theoretically, that standard tricks of implicitly inducing this positional structure, such as data augmentation with shifting contexts \cite{brown2020language}, are not sufficient to guarantee this generalization. Our key findings are summarized below:

\noindent\textbf{(F1) Positional encodings that encapsulate the structure enable length generalization for addition and multiplication.} For addition, relative position encoding (RPE) allows the Transformer to naturally capture digit-wise alignment in order to achieve length generalization from adding two numbers with up to 5-digit to two numbers with up to 50-digit (similar generalization was observed in \cite{jelassi-length}). For multiplication, our proposed new uniform positional encoding (UPE) that assigns each digit of a number the same position encoding, coupled with RPE, allows the Transformer to length generalize from multiplication of 3-digit numbers with 5-digit numbers to multiplication of a 3-digit number with 20-digit numbers. To our knowledge, this is the first positive result on length generalization for multiplication without using chain-of-thought \cite{wei2022ChainofThought} or data priming \cite{jelassi-length}.

\noindent\textbf{(F2) Data augmentation is insufficient for length generalization.} The most natural data augmentation of shifting the numbers by adding zeros to the right to implicitly induce the necessary alignment needed for both addition and multiplication helps with length generalization; However, the performance is significantly worse than the crafted positional encodings. To further elucidate this, we propose a simple linear setting that captures the relative position alignment of addition, and prove that a one-layer linear Transformer trained with gradient descent with RPE generalizes, and shifting with zeros to encode a similar bias does not help.

\noindent\textbf{(F3) Incorporating higher complexity shorter sequences improves generalization to longer sequences.}
% We show that increasing coverage of data from the shorter sequences leads to improvements in performance on longer sequences.
While our proposed positional encodings address the structural aspect, we also isolate the other major reason for failure on longer samples to be related to the increase in complexity of the task. 
% We identify that the key cause for this is the increased coverage of higher complexity examples which lie in the tails of the distribution over shorter sequences. 
We show that the increase in coverage of higher complexity examples, which lie in the tails of the distribution over shorter sequences, leads to improvements in performance on longer sequences. For the task of addition, we show the required length of shorter sequences needed to adequately cover complex examples for larger sequences under the uniform distribution on numbers is not very large. This allows us to cover sufficient complexity with 5-digit numbers to generalize to 50-digit numbers.
\paragraph{Related Work.} Despite significant advances, Transformer models such as GPT-4 \cite{openai2023gpt4} struggle with high-complexity arithmetic tasks \cite{dziri2023faith, yang2023gpt}. Investigations into these limitations have ranged from ineffective representation learned for numbers \cite{wu2016google, sennrich-etal-2016-neural, wallace2019nlp, thawani2021representing}, to the inability to achieve numeracy via unsupervised learning \cite{razeghi2022impact, li2022systematic, kim-etal-2021-seen, wang2021exploring, mahdavi2023better}. Attempts to mitigate these deficiencies such as fine-tuning pre-trained models\cite{geva2020injecting, lewkowycz2022solving}, scratch-padding\cite{nye2021work}, and leveraging chain-of thoughts reasoning \cite{wei2022ChainofThought, ouyang2022training} only provide marginal out-of-distribution (OOD) benefits \cite{anil2022exploring}. Specifically, \cite{anil2022exploring} shows that scratchpad fine-tuning fails to generalize to longer problems, and even with chain-of-thoughts, model are prone to error-propagation, learning shortcut solutions and insufficient features, etc \cite{dziri2023faith}.  Recent efforts have proposed modification to data and model structures to achieve length generalization \cite{jelassi2023length, shen2023positional}. This includes introducing new structures to positional vectors or leveraging positional information like \cite{rpe, huang2018music,su2023roformer,raffel2023exploring}. Some studies suggest removing positional vectors or using randomized encodings \cite{kazemnejad2023impact,ruoss2023randomized}. Despite these efforts, arithmetic tasks remain difficult. In this work, we assert that without explicitly encoding symmetries, generalization remains unattainable even with implicitly encoding symmetries via data augmentation (see \Cref{arith}). Building on prior work, we employ RPE \cite{rpe} motivated by the results of \cite{jelassi2023length} for addition, and design our custom embedding for multiplication in \Cref{arith}, providing a new framework to leverage symmetries effectively and give the first theoretical explanation for the effectiveness of this approach in \Cref{Section_RPE,label:theory}. 

Beyond architectural changes, modifying the training distribution has also been a prominent direction for improving performance \cite{lee2023teaching, shen2023positional}, particularly through curriculum learning \cite{wang2021survey, yang2023gpt}, and priming \cite{jelassi-length}. Our observations in \Cref{Section: complexity} suggest that exposing models to more complex in-distribution samples can enhance OOD performance.  We provide additional related works in \Cref{app:related-work}.

% Other work has found a language framework for tasks that are implementable with transformers \cite{weiss2021thinking, zhou2023algorithms}. 
% In this work, we posit the problem of length generalization as one case of generalization on the unseen (\textit{GOTU}) \cite{abbe-lengen}, and we assert that without explicitly encoding symmetries, generalization remains unattainable even with implicitly encoding symmetries via data augmentation (see \Cref{arith}). \surbhi{Add a line about how it differs from all the other approaches tried so far.}
\section{Preliminaries: Positional Encodings}\label{setup}
% In this paper, we focus on enabling Transformers to solve arithmetic tasks in an end-to-end manner, without the use of chain-of-thought or scratchpads, and in a non-autoregressive manner. We believe the end-to-end approach is more effective, and perhaps necessary for simple arithmetic tasks, since chain-of-thoughts necessitates additional supervision to dissect each task into simpler steps for the model. Our primary focus is to exploit the structure and symmetries present in the task to obtain length generalization. This goal will accomplished with explicitly incorporating the symmetries to the positional encodings.
To ground our design of positional encodings, we first describe a general formalism for positional encodings in this section.
% Then, we review the setup of our experiments.

\paragraph{Transformer Architecture.} A transformer model takes as input a sequence of $d_x$-dimensional vectors, denoted as $\bx = (x_1, \dots, x_n)$ where $x_i \in \R^{d_x}$. These vectors are embedded using an embedding matrix $\bE: \R^{d_x}\to \R^{d_z}$, producing $\bz^{(0)} = (z^{(0)}_1, z^{(0)}_2, \dots, z^{(0)}_n) = (\bE x_1, \dots, \bE x_n)$, where $d_z$ is the dimension of the embedded space. Each Transformer block consists of attention layers operating on $\bz^{(l-1)}$ to produce $\bz^{(l)}$ using $H$ heads, followed by a shared MLP layer. Each head computes attention scores: 
$$A^{(l)}_{i, j} = {\rm Softmax}_j\big((W_Q z^{(l-1)}_i)^T (W_K z^{(l-1)}_j)\big)$$
The output of each attention head is computed as follows:
$$b^{(l)}_i = \sum_j A^{(l)}_{i, j} (W_V z^{(l-1)}_j) \in \R^{d_z / H} $$ 
where $W_Q, W_K, W_V \in \R^{d_z/H \times d_z}$, are three matrices that together form the core of each head. The concatenated head outputs are processed by a feed-forward network to obtain $z^{(l)}_i$. 

In the formulation given above, the output at position $i$ remains invariant to the permutation of the rest of the sequence ($\{x_1, \dots, x_{i-1}, x_{i+1}, \dots, x_n\}$), which is a limitation for our tasks where the sequence order is indispensable. As a result,  positional encodings are employed.

\paragraph{Absolute Positional Encoding (APE).} One approach to overcome this limitation is to encode positional information by adding a collection of unique and fixed vectors to the input $x_i$'s: $$z^{(0)}_i = \bE x_i + p_i$$ 
This method is known as absolute positional encoding (APE), where the positional vectors can either be predefined and fixed or learned during training, adapting to specific task requirements.

\paragraph{Pairwise Positional Encoding.} More generally, since every layer of an attention model only considers pairwise interactions between elements, we can move the positional vectors inside the calculation of attention scores either by adding the pairwise vectors ($p_{ij}$) to the embedded keys, or to the embedded queries, or to the computed attention between a query and a key. In all experiments, we adhere to the method of adding them to the key, as suggested in \cite{rpe}. That is, 
\begin{equation}\label{pairwise}
    A_{i, j} = {\rm Softmax}((W_Q z_i)^T (W_K z_j + p_{i, j})).
\end{equation}
We propose tailoring positional vectors $p_{i,j}$ to the specific relationships between positions in a task. If the connection from position $i_1$ to $j_1$, denoted by $i_1 \to j_1$, has the same functionality as  $i_2 \to j_2$, we can set $p_{j_1, i_1}$ equal to $p_{j_2, i_2}$. This ensures $i_2$ influences $j_2$ similarly to how $i_1$ influences $j_1$. In general, if any positions have equivalent functionality, we will set their corresponding pairwise positional vectors to be the same. 

\paragraph{Relative Positional Encoding (RPE).} For tasks with translational symmetries, we have the following property: $\big(i \to j \big) \sim \big((i+t) \to (j+t) \big) \quad \forall t$, in which "$\sim$" stands for being equivalent in our notation. This symmetry implies that: $p_{i, j} = p_{i+t, j+t}$ for all $t$. Therefore, we can express positional vectors using the difference between $i$ and $j$: $$p_{i- j} \leftarrow p_{i ,j}$$
Consequently, computations are based on relative positions, i.e.  the output depends on the relative, not absolute, positions in the sequence, hence the term relative positional encoding (RPE) \cite{rpe}.

\section{Length Generalization for Arithmetic tasks}\label{arith}
In this paper, we focus on two arithmetic tasks: addition and multiplication. We first formalize the setup of our tasks and then describe our findings.

\paragraph{Data format.} The input consists of two integers represented per-digit as $x^{(1)}= (x^{(1)}_{l_1}, \dots, x^{(1)}_1), x^{(2)}= (x^{(2)}_{l_2}, \dots, x^{(2)}_1)$, and we ensure both sequences are of the same length ($l_1 = l_2 = l$) by adding "pad" tokens to both integers, as necessary. Consider for illustration the following example: "123 + 4095 = 4218", and $l = 20$:
\begin{flalign}\label{eq:format_sum}
        &\text{\textbf{Input}: }(\underbrace{\text{pad}, \dots, \text{pad}}_{\text{17 times}}, 1, 2, 3, +, \underbrace{\text{pad}, \dots, \text{pad}}_{\text{16 times}}, 4, 0, 9, 5) \nonumber \\
        & \text{\textbf{Output}: }  \hspace{0.72in} (\underbrace{\_, \dots, \_}_{\text{20 times}}, \underbrace{\text{pad}, \dots, \text{pad}}_{\text{17 times}}, 4, 2, 1, 8) 
\end{flalign}
For multiplication, we use the same input format with the difference that the multiplier is not padded and the "$+$" token is replaced with "$\times$":
\begin{flalign}\label{eq:format_mult}
        & \text{\textbf{Input}:} \ \hspace{0.43in} ( 5, 6, \times, \underbrace{\text{pad}, \dots, \text{pad}}_{\text{16 times}}, 4, 2, 9, 7) \nonumber \\
        & \text{\textbf{Output:}} \ (\_, \dots, \_, \underbrace{\text{pad}, \dots, \text{pad}}_{\text{16 times}}, 2, 4, 0, 6, 3, 2) 
\end{flalign}
Note that in the example above, the maximum length of the output can reach 22 digits, accounting for the worst-case scenario of multiplying a 20-digit number by a 2-digit number. 

Indexing from right to left for simplicity, the first digit of the answer corresponds to the first position of the input. The "$\_$" symbol in the output indicates neglected tokens that are not part of the supervision signal for the addition task. Note that our formatting approach is distinct from that of \cite{jelassi2023length} in that we pad each integer from the left to ensure that the relative coordinates of digits remain consistent regardless of the integers' lengths, and also from other prior approaches that reverse the digits \cite{shen2023positional, lee2023teaching}. The fixed length assumption is not unrealistic since numbers are typically represented with fixed bit-complexity in most computations. Consequently, the tokenizer can readily impose a fixed-length structure of any chosen length as sketched in \Cref{fig:LLM_sketch}. 
% Besides, since positional encodings are added per pair in the attention mechanism, we can embed this structure regardless of the numbers' positions.

\paragraph{Length generalization task.} For our tasks, the output consists of the integer resulting from the arithmetic operation on $x^{(1)}$ and $x^{(2)}$. We can model our length generalization problem as an instance of generalization on the unseen \cite{abbe-lengen}. We denote the domain of all pairs of integers with lengths $\le l$ by $\D := \Big\{ (x^{(1)}, x^{(2)}) \in \mathbb{Z}^2 : 0 \leq x^{(1)}, x^{(2)} < 10^{l} \Big\}$. In the problem of length generalization , we observe only a subset of this domain during training. Specifically, pairs of integers with lengths at most $l_s < l$ are used in training. We denote this \emph{seen domain} as $\D_s := \Big\{ (x^{(1)}, x^{(2)}) \in \mathbb{Z}^2 : 0 \leq x^{(1)}, x^{(2)} < 10^{l_s} \Big\}$, and denote the \emph{unseen domain}, which consists of at least one integer with longer length, as $\D_u := \Big\{ (x^{(1)}, x^{(2)}) \in \mathbb{Z}^2 : 0 \leq x^{(1)}, x^{(2)} < 10^l, \ 10^{l_s} \leq x^{(1)} \lor 10^{l_s} \leq x^{(2)} \Big\}$. We thus have: $\D = \D_s \cup \D_u$. We will focus on the following question: \textit{If we train the model using samples from $\D_s$ to near perfect in-distribution accuracy, what can be inferred about the accuracy on $\D_u$?} 

\paragraph{Model.} For all tasks, we utilize a BERT-based encoder-only attention model \cite{devlin2018bert}. We use an encoder-only architecture because causal maps are not beneficial for our tasks. For instance, to compute the second digit of a sum, the model must consider the first two digits of both integers. This entails both looking forward and backward in the input. We also chose to avoid autoregressive settings because they often require several tricks to function, such as reversing the digits order either in the product or the inputs \cite{shen2023positional, lee2023teaching}. Since arithmetic operations naturally start from the least significant digit, generating them from left to right in an auto-regressive manner offers no computational advantage. The number of attention layers that we employ will vary depending on the difficulty of the task. 

\paragraph{Training and Evaluation.} During training, samples are generated uniformly from $\D_s$. At test time, we evaluate the generalization performance based on prediction accuracy over a uniform distribution of numbers up to $l$ digits. An output is deemed correct only if all digits, including "pad" symbols, are correct; otherwise, it is considered false. Note that this is worst-case accuracy over digits but average-case accuracy over $l$-digit numbers.  We present results for different training checkpoints. As noted in \cite{jelassi-length}, out-of-distribution accuracy can fluctuate even as in-distribution accuracy improves. Therefore, we show performance across various checkpoints, using a small validation set with longer sequences to choose the best checkpoint. Further experimental details are in \Cref{exp_det}.
% \begin{flalign}\label{eq:format_mult}
%         & \text{\textbf{Input}:} \ \hspace{0.43in} ( 5, 6, \times, \underbrace{\text{pad}, \dots, \text{pad}}_{\text{16 times}}, 4, 2, 9, 7) \nonumber \\
%         & \text{\textbf{Output:}} \ (\_, \dots, \_, \underbrace{\text{pad}, \dots, \text{pad}}_{\text{16 times}}, 2, 4, 0, 6, 3, 2) 
% \end{flalign}
% Note that in the above, the maximum length of the output can be 22 since in the worst-case, it is a product of a 20-digit and a 2-digit number. 
%Therefore, the formatting includes the appropriate padding digits. 
%Note that we require the model to predict the pads correctly as well.

\subsection{Task: Addition}\label{Section:addition}
In a transformer model, characterized by parallel information processing, the common iterative way to perform multi-digit addition is not effective. Instead, the model must simultaneously identify all relevant positions contributing to each of output digits.
% An example of this is given in Figure \ref{sum_example}.

\paragraph{Addition in parallel.}\label{Section: parallel}
The result of \cite{quirke2024understanding} showcases that the transformer dissects the calculation of the sum into parallel streams of computation.
Refer to \Cref{sum_example}, which presents a parallel method for addition; First, it adds the corresponding digits across both integers, and then propagates the carries. 
% In order to compute the output at position $i$, the model has to identify some items: First, a pair $(a^{(1)}_i, a^{(2)}_i)$ where $a^{(1)}_i$ is the position in the input sequence that corresponds to the $i$-th digit of the first integer, and $a^{(2)}_i$ corresponds to the position of the $i$-th digit of the second integer. The second item is a list ${\rm C}_i = \left({\rm C}_{i, 1}, \cdots, {\rm C}_{i, m}\right)$ of all other positions in the input sequence whose values are needed to compute the carry-over. We will use $\sigma(i) =  \left(a^{(1)}_i, a^{(2)}_i, {\rm C}_{i, 1}, \cdots, {\rm C}_{i, m} \right)$ to denote the list of all the required positions for position $i$. As the next step, the model has to learn a function that computes the output at position $i$, i.e. the result of the sum at position $i$, given the values at these positions in $\sigma(i)$. For instance, to compute the output at position $i = 2$ for the sample given in Section \ref{setup}, we have $a^{(1)}_i = 2$, $a^{(2)}_i = 23$, and ${\rm C}_i = (1,22)$. Note that the indices are read from right to left. In general, to compute the sum of two $l$-digit integers in our format, $\sigma$ can be written as:
% \vspace{-0.2cm}
% \begin{align}\label{eq:2}
%     \sigma(i) =  \big(&a^{(1)}_i = i, a^{(2)}_i = i + l + 1, {\rm C}_i = (1, \cdots, i - 1, l+2, \cdots, i + l) \big)
% \end{align}
To compute the output at position $i$, the model must identify two items: First, the pair of positions in the input sequence corresponding to the $i$-th digit of the first and second integers. The second item is a list of all other positions in the input sequence required to compute the incoming carry-over. Having described these items, we will now synthesize them into $\sigma(i)$ to denote the list of all positions required for position $i$. Next, the model learns a function to compute the output at position $i$ based on the values at these positions in $\sigma(i)$. For instance, to compute the output at position $i = 2$ for the sample given in \Cref{eq:format_sum}, we have $\sigma(2) = (1, 2, 22, 23)$, pinpointing the first two digits of both integers. In general, for two $l$-digit integers in our format, $\sigma$ is written as $\sigma(i) =  \big(1, \cdots, i - 1, i, l+2, \cdots, i + l, i + l + 1 \big)$.
% \vspace{-0.2cm}
% \begin{align}\label{eq:2}
%     \sigma(i) =  \big(i, i + l + 1, 1, \cdots, i - 1, l+2, \cdots, i + l \big)
% \end{align}
The model then needs to learn and apply $\sigma(\cdot)$ to any pair of numbers from the unseen domain $\D_u$. 

\begin{figure}
\centering
\begin{minipage}[b]{0.35\textwidth}
    \centering
    \includegraphics[width=\linewidth]{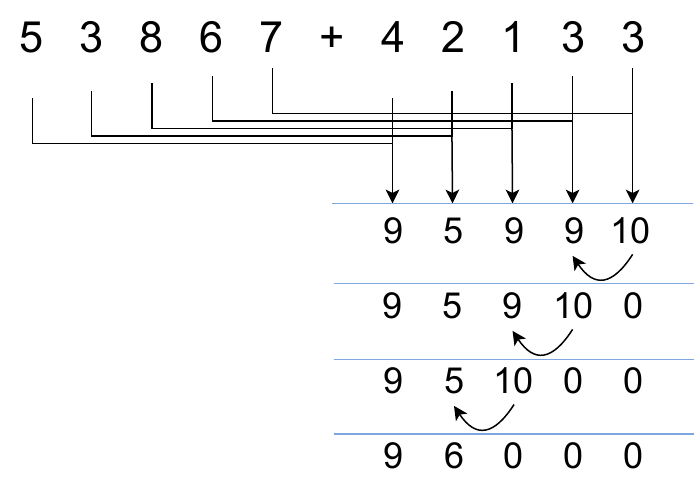}
    \subcaption{}
    % \label{sum_example}
\end{minipage}\hfill
\begin{minipage}[b]{0.55\textwidth}
    \centering
    
    \begin{algorithm}[H] % Use [H] to set the algorithm in place
    \scriptsize 
    \caption{Parallel Carry-Handle Addition}
    \begin{algorithmic}[1]
    \State $x^{(1)} = (x^{(1)}_{l}, \dots, x^{(1)}_1)$ \Comment{first integer}
    \State $x^{(2)} = (x^{(2)}_{l}, \dots, x^{(2)}_1)$ \Comment{second integer}
    % \State $s = (s_{l+1}, \dots, s_{1})$ \Comment{array of size $l+1$ initialized to 0}
    \State $s = (s_{l+1}, \dots, s_{1})$ \Comment{array of size $l+1$ initialized to 0}
    \For{$i \gets 1$ \textbf{to} $l$}
        \State $s_i \gets x^{(1)}_i + x^{(2)}_i$
    \EndFor
    \While{$\exists i, s_i \geq 10$ \textbf{for some} $i$}
        \For{$j \gets 1$ \textbf{to} $l$}
            \State $s'_{j+1} \gets s_{j+1} + \lfloor s_j / 10 \rfloor$ \Comment{Adjust one carry}
        \EndFor
         \For{$i \gets 1$ \textbf{to} $l$}
            \State $s_j \gets s'_j \mod 10$ \Comment{Remainder counts}
        \EndFor       

    \EndWhile
    \State \textbf{return} $s$
    \end{algorithmic}
    \end{algorithm}
    \subcaption{}
    % \label{alg:sum_example}
\end{minipage}
\setcounter{figure}{0}
\caption{Parallel Carry-Handle Addition. \textbf{(a)} An illustration of a parallel algorithm that handles cascading carries in addition. The algorithm first adds the corresponding digits of the two integers and then propagates the carries to the latter positions. \textbf{(b)} Pseudo-code explanation of the process. Note that the for loop computations can take place in parallel for each position.}
\label{sum_example}
\end{figure}

\paragraph{(A1) Failure of absolute positional encoding.} We first demonstrate that while APE can succeed in generalizing to in-distribution data, its performance degrades significantly on longer length sequences. \Cref{APE_5to20} shows the performance of our trained model (on the uniform distribution of numbers with $l_s= 5$) on a uniform distribution over domains with lengths varying from 6 to 20. As expected, performance drops drastically when the length increases from 5 to 6.
APE does not capture translational symmetries in the task of addition. In this regard, prior work \cite{abbe-lengen} has shown that it is in general impossible to obtain length generalization without providing extra information to the model, either by imposing task symmetries on the model or by using data augmentation in training. Next, we will see that encoding the translational symmetries through relative positional encoding (RPE) will lead to generalization on the unseen positions in the task of addition. 

% In our next experiment, we investigate whether the model can accurately sum longer samples constructed from shifting two samples from $\D_s$ to the left with zeros added to the end (clearly, the result of the sum is also shifted to the left). The purpose of this experiment is to see if the model can learn the translational equivariances in the task of addition; i.e. shifting the two input integers to the left should result in shifting their summation to the left by the same amount. Figure \ref{APE_shifting} shows that the model does not produce the same output after shifting, and hence, we conclude that the model, trained only on $\D_s$, has failed to recognize these translational equivariances. 

\begin{figure}
     \centering
     % You have to set the textwidths to sum to under 1 because Latex doesn't like it when the two are exactly 1. :(
     % This splits the figure into it to two half pages width figures.
     \begin{subfigure}[b]{0.32\textwidth}
         \centering
         % Linewidth is the now the unit for half a page.
         \includegraphics[width=1.0\linewidth]{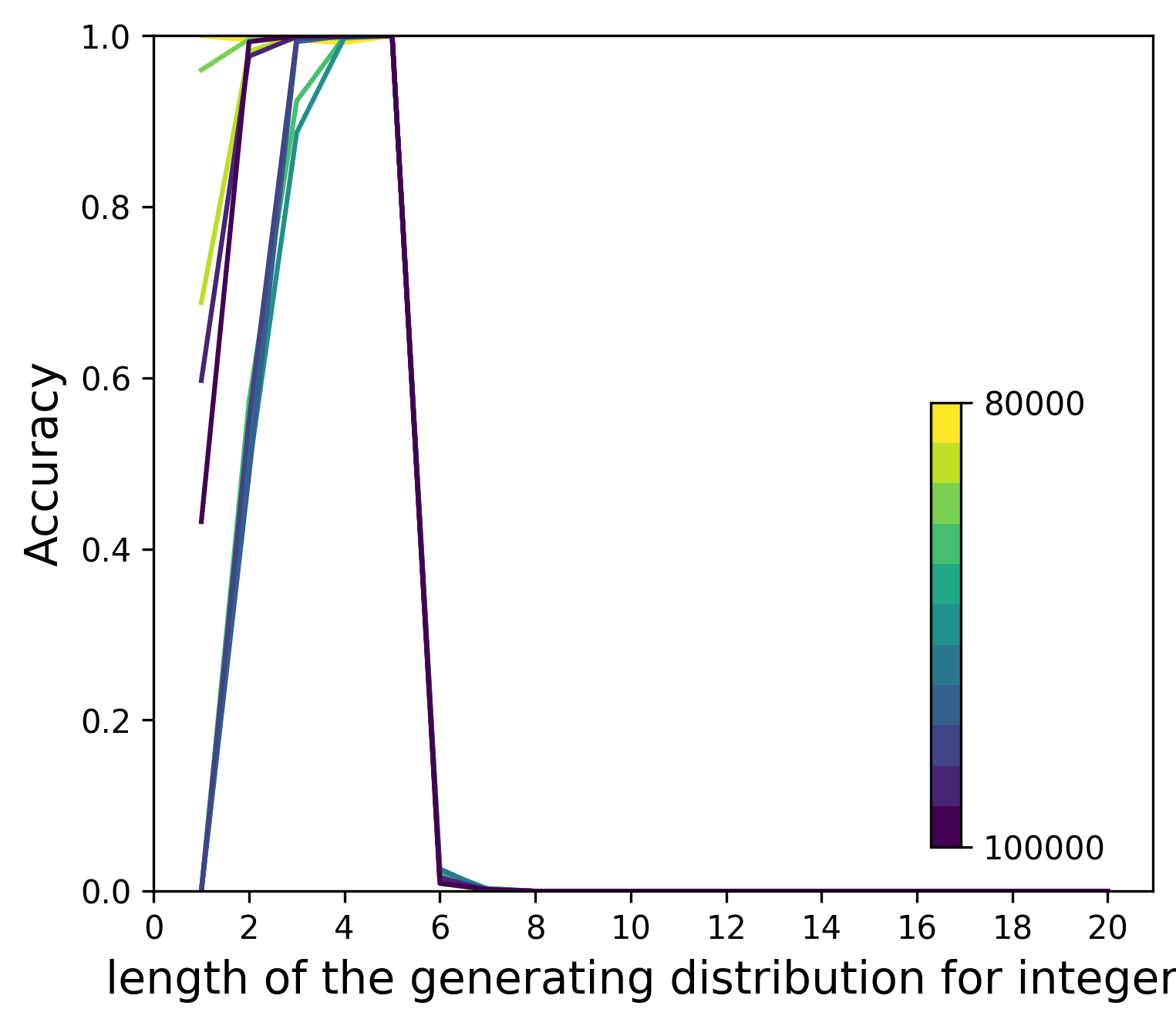}
        % \vspace{-2ex}
         \caption{}
         \label{APE_5to20}
     \end{subfigure}
     % \hfill
    % \hspace{0.15in}
     \begin{subfigure}[b]{0.32\textwidth}
         \centering
         % Linewidth is the now the unit for half a page.
         \includegraphics[width=1.0\linewidth]{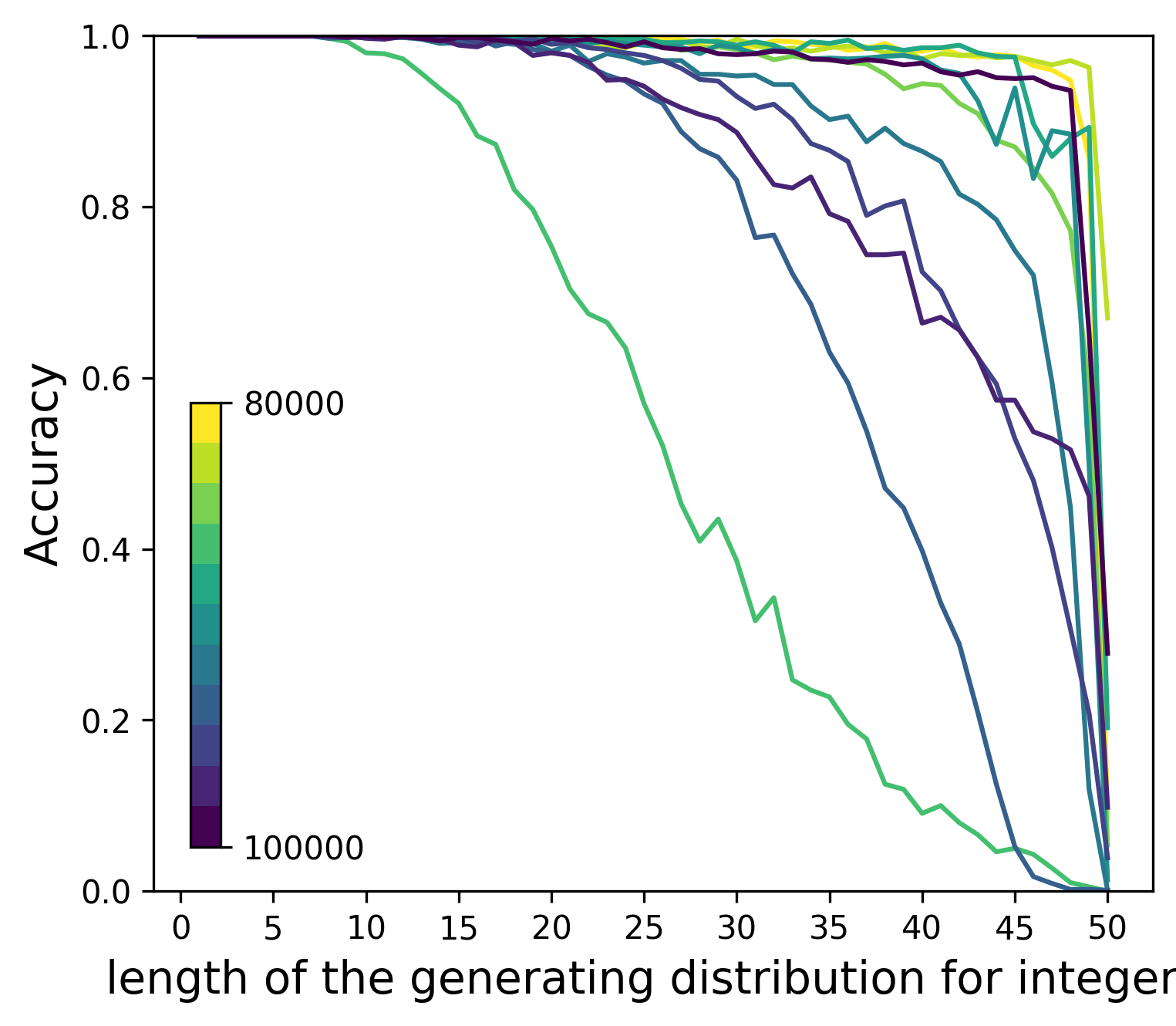}
     % \vspace{-2ex}
         \caption{}
         \label{RPE_5to50}
     \end{subfigure}
    % \hspace{0.15in}
     \begin{subfigure}[b]{0.32\textwidth}
        \centering
        \includegraphics[width=1.0\textwidth]{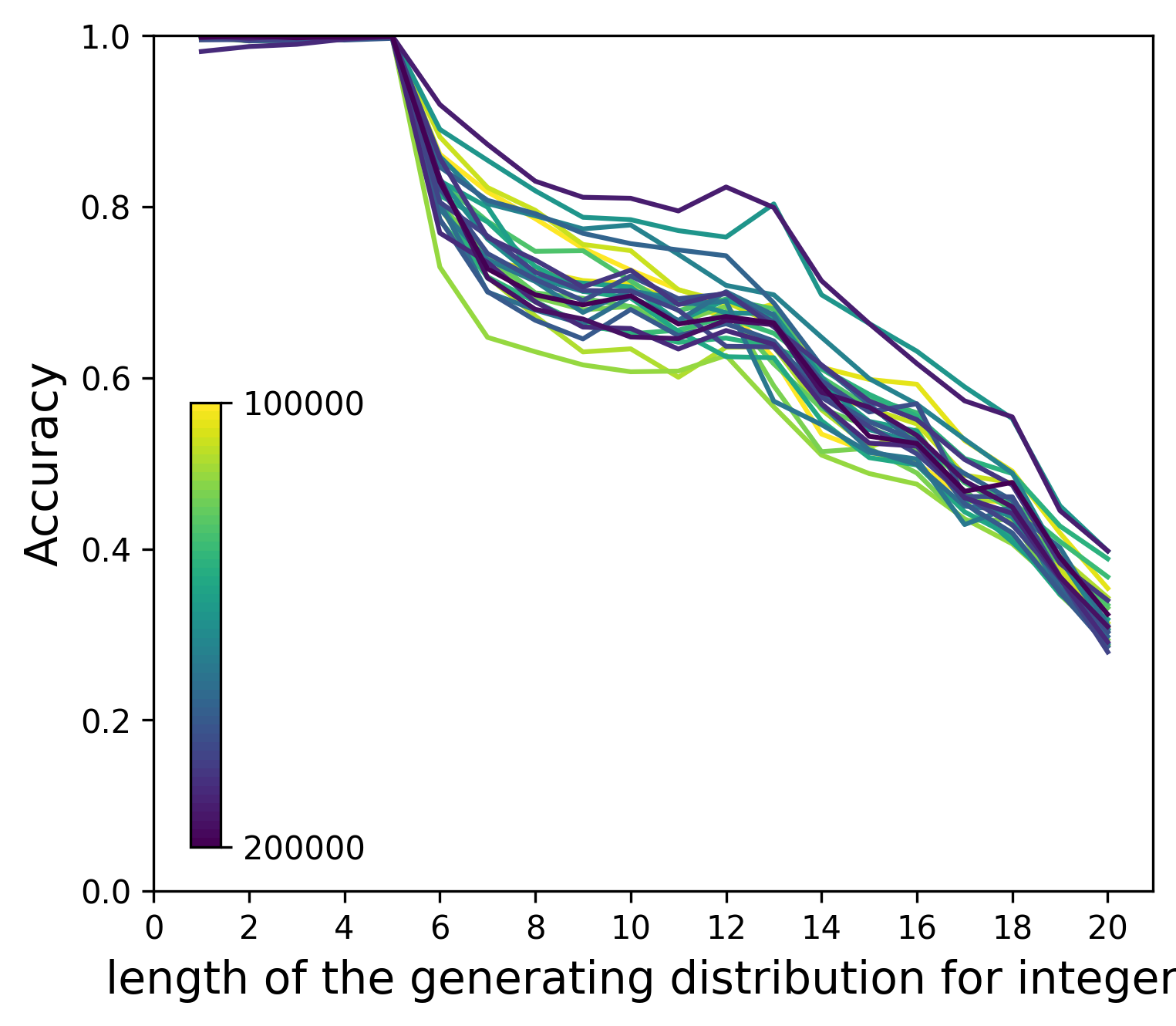} 
    % \vspace{-2ex}
        \caption{}
        \label{APEshifted_5to20}
    \end{subfigure}
    \caption{\textbf{(a)} Accuracy of the model using absolute positional encoding while increasing the length of the sequence. \textbf{(b)} The model with RPE trained with up to 5-digit sums and tested on up to 50-digit sum. While showing high fluctuations, the inductive bias of relativity makes generalization possible. \textbf{(c)} The model with APE when trained with augmented data, i.e., all shifted versions of samples from the dataset are given in training. 
    % Even though the models achieves full accuracy on augmented data, it still lacks the generalization ability. 
    }
\end{figure}

\paragraph{(A2) Relative position encoding enables length generalization.} \label{Section_RPE} 
% In this regard, we argue that RPE is capable of encoding translational equivariances for the task of addition.
% We conducted another experiment with BERT, this time incorporating relative positional vectors into the model. In order to demonstrate the advantages of our trained model, we tested it up to 50-digit sums. 
\Cref{RPE_5to50} confirms that RPE is capable of encoding translational equivariances for the task of addition. Adding RPE to the Transformers maintains its high performance at longer lengths. This effectiveness of RPE was first captured in \cite{jelassi-length}. Note that the abrupt fall in 50-digit sums is due to the input format. In all samples seen from $\D_s$, the two integers are separated with a "+" and several "pad" tokens, whereas when testing on 50-digit sums, there is no "pad" sign that separates them. We consider this to be a spurious correlation that the model has learned. the gradual decline in accuracy for longer sequences is a consequence of two factors. More importantly, the gradual decline in accuracy for longer sequences is caused by samples with higher complexities that are more likely among larger integers, according to the definition we give for complexity in \Cref{Section: complexity}.
% We will demonstrate how the model fails to sustain its accuracy on these more complex samples, causing the main reason for decrease in performance.

% \begin{figure}
%      \centering
%      % \begin{subfigure}[b]{0.235\textwidth}
%      %     \centering
%      %     \includegraphics[width=1.0\linewidth]{figs/APE_shifting.png}
%      %     \caption{}
%      %     \label{APE_shifting}
%      % \end{subfigure}
%      % \begin{subfigure}[b]{0.235\textwidth}
%          \centering
%          \includegraphics[width=1.0\linewidth]{figs/RPE_5to50_shifting.png}
%          % \caption{}
%          \label{RPE_shifting}
%      % \end{subfigure}
%         \caption{\textbf{(a)} Average accuracy of a model with APE trained with up to 5-digit numbers for samples inside the dataset when both numbers have been shifted with zeros. \textbf{(b)} Average accuracy for the model with RPE \surbhi{Combine this into one?.}}
%     \vspace{-0.5cm}
% \end{figure}

% We evaluated the accuracy of the model on shifted samples from the training dataset. As expected, the model performs perfectly on shifted samples.

\paragraph{(A3) Adding symmetries implicitly via augmented data is not enough.}\label{Aug_fails}
Knowing that the task has certain symmetries, another potential approach is to use data augmentation to implicitly encode the symmetries. For the task of addition, a natural data augmentation scheme consists of shifting a pair of integers in the training data by adding the same number of zeros from the right. \footnote{We are leveraging the neutral role of zero in a sum. Note that this augmentation does not introduce any new nonzero digits, avoiding the need for additional computations. It also does not append training samples that could potentially cover the entire unseen domain, which might artificially enhance performance by unfairly exploiting dependency constraints in the addition task that are oblivious to relative positional encoding.} We trained a BERT model, again with APE, on this augmented dataset, and the result can be seen in \Cref{APEshifted_5to20}. The figure shows that augmentation does not lead to generalization on the unseen positions. However, we remark that the in-distribution performance (on shifted test data) is near-perfect (above 0.999), indicating that it has learned translation by zeros. To understand the failure of the model trained with augmentation, we examined the model's output for 10-digit sums. Most errors occurred in the least significant digit, even though the model performs well on single-digit sums in \Cref{APEshifted_5to20}. The detailed explanation of this experiment can be found in \Cref{Aug:table}. Since augmentation only adds zeros when shifting, causing the model to learn spurious correlations between the first position and positions larger than $l_s$, which will not hold at test time. We will address this erroneous learning in second part of \Cref{alltheory}, where only some parameters are properly learned during training, while others can grow inadvertently. We show a similar spurious correlation appearing in our simplified theoretical setup in \Cref{label:theory}.

% \begin{table}[h!]
%      % \vskip 0.15in
%     \centering
%     \scriptsize
%     \begin{tabular}{lcccr}
%     \toprule
%     Test Case & Accuracy \\
%     \midrule
%     Shifted integers & 99.7 $\pm$ 0.1\\
%     Uniform distribution & 81 $\pm$ 1\\
%     Accuracy on the 1st digit & 90 $\pm$ 1 \\
%     Accuracy on the 2nd digit & 96 $\pm$ 1\\
%     Accuracy on the 3th digit & 98.8 $\pm$ 0.1\\
%     Accuracy on the 4th digit & 98.0 $\pm$ 0.1\\
%     Accuracy on the 5th digit & 98.8 $\pm$ 0.1\\
%     Accuracy on 6:11 digits & 96 $\pm$ 1\\
%     \bottomrule
%     \end{tabular}
%     %\end{sc}
%     \label{tab1}
%     \caption{10-digit sum's accuracy of the augmented model for in-distribution, out of distribution (uniform integers in $[1, 10^{10} -1]$), and isolating the performance at every coordinate of the output. Despite the acceptable accuracy of the model on most digits, the first digit errors hurt the model.}
% \end{table}

\subsection{Task: Multiplication}\label{Section:mult}
Efficient algorithms for multiplication, such as 'shift and add', cannot compute multiplication in completely parallel streams. However, a parallel algorithm is necessary to effectively induce the task structures into a transformer architecture. \Cref{mul_example} proposes a method that, under a limited length constraint for the multiplier, splits a multi-digit multiplication into the product of the multiplier with each digit of the multiplicand, and propagates the carries subsequently. To induce this mechanism, we introduce \emph{uniform positional encoding}, where each digit of the multiplier uniformly affects all digits of the multiplicand.

\begin{figure}
\centering
\begin{minipage}[b]{0.32\textwidth}
    \centering
    \includegraphics[width=1.0\linewidth]{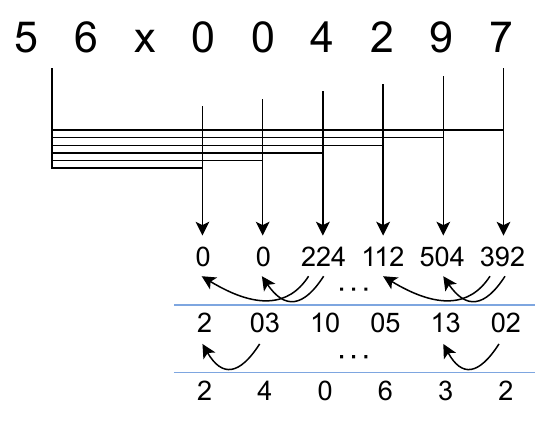}
    \subcaption{}
    \label{mul_example}
\end{minipage}\hfill
\begin{minipage}[b]{0.55\textwidth}
    \centering
    \begin{algorithm}[H] % Use [H] to set the algorithm in place
    \scriptsize
    \caption{Parallel Carry-Handle Multiplication}
    \begin{algorithmic}[1]
    \State $x^{(1)} = (x^{(1)}_{l_1}, \dots, x^{(1)}_1)$ \Comment{first integer}
    \State $x^{(2)} = (x^{(2)}_{l_2}, \dots, x^{(2)}_1)$ \Comment{second integer}
    \State $m = (m_{l_1+ l_2}, \dots, m_{1})$ 
    \For{$i \gets 1$ \textbf{to} $l$}
        \State $m_i \gets x^{(1)} * x^{(2)}_i$
    \EndFor
    \While{$\exists i, m_i \geq 10$ \textbf{for some} $i$}
        \For{$k \gets 1$ \textbf{to} $l_1$}
            \For{$j \gets 1$ \textbf{to} $l_1 + l_2 - k$}
                \State $m'_{j+k} \gets m_{j+k} + \lfloor m_j / 10^k \rfloor$ \Comment{Handle one carry}
            \EndFor
        \EndFor
        \For{$j \gets 1$ \textbf{to} $l_1 + l_2 -1$}
            \State $m_j \gets m'_j \mod 10$ \Comment{Remainder counts}
        \EndFor
    \EndWhile
    \State \textbf{return} $m$
    \end{algorithmic}
    \end{algorithm}
    \subcaption{}
    \label{alg:mul_example}
\end{minipage}
\setcounter{figure}{2}
\caption{\textbf{(a)} An illustration of a parallel algorithm that computes the multiplication, inspired from the uniform functionally of the multiplier in the absence of carries. Besides, carries still has a relative structure for propagation to following positions. \textbf{(b)} Pseudo-code explanation of the process assuming $l_1 \ll l_2$.}
\end{figure}

\paragraph{Uniform Positional Encoding (UPE).} 
% When we have positions that affect all the other positions in a uniform manner. This motivates our uniform PE: 
Starting from \Cref{pairwise}, consider a scenario where position $i$ uniformly affects every other position  $j \neq i$. In this case, the positional vector $p_{j, i}$ must remain constant for every $j$:
% \vspace{-0.05in}
\begin{equation}\label{upe}
    p_{j, i} = p_{j', i} = c_i \quad \forall j, j' \neq i
\end{equation}

Furthermore, if a task involves multiple positions with uniform functionality, assuming ${i_1, \dots, i_m}$ is the set of such positions, we assign them the positional vectors $c_1, \cdots, c_m$, respectively. If we denote those positions in the sequence whose outputs have to be computed as $\{j_1, \cdots, j_n\}$, the computation will need the following collection of positional vectors: $\{p_{i,j}; i,j \leq n \} \cup \{c_1, \dots, c_m\}$. Note that additional relationships may exist among ${j_1, \dots, j_n}$, such as translational properties, which would justify the use of relative positional encoding. For example, for multiplying a 3-digit number by a 20-digit number, we will use $c_1, c_2, c_3$ for the multiplier as well as relative positional vectors $\{p_{-21}, \cdots, p_{21}\}$ (length of the answer is at most 22 digits). 

\noindent\textit{Remark.} Multiplication is significantly more challenging than addition, even when the multiplier is a single-digit integer. The model has to deal with larger stack of carries compared to the addition task. This is because the probability of a carry occurring is higher at each stage, and unlike in addition, carries can exceed 1. See \Cref{carry_dist} for more explanation.  Consequently, we limit our experiments to 3-digit multipliers and focus on the generalization of the multiplicand. Even in this setting, we require larger networks compared to the task of addition to generalize.

\begin{figure}[t!]
    \centering
     % You have to set the textwidths to sum to under 1 because Latex doesn't like it when the two are exactly 1. :(
     % This splits the figure into it to two half pages width figures.
     \begin{subfigure}[b]{0.32\textwidth}
         \centering
         % Linewidth is the now the unit for half a page.
         \includegraphics[width=1.0\linewidth]{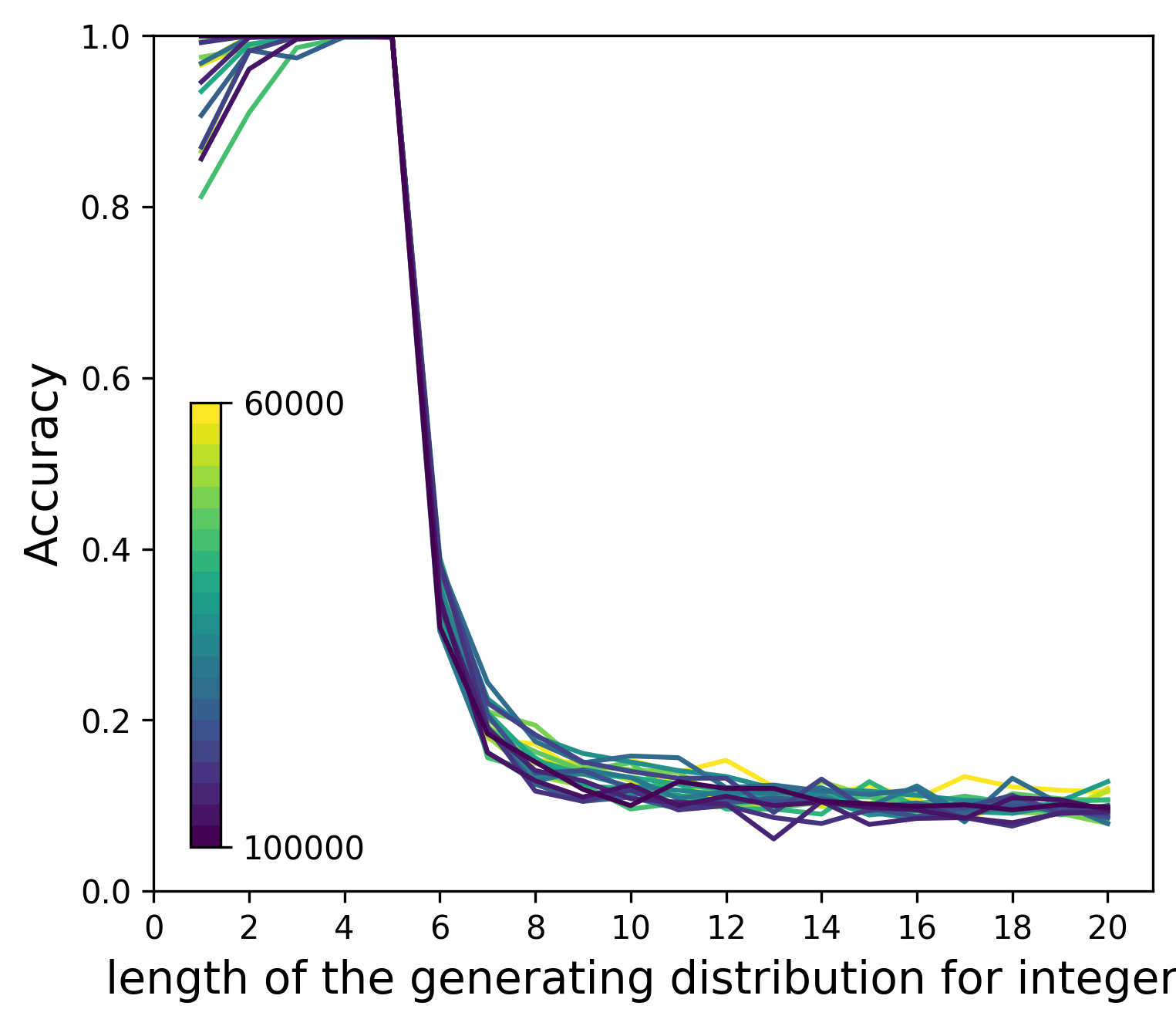}
         \caption{}
         \label{Mult_APE}
     \end{subfigure}
     %\hfill % Remove this comment if you want the figures to be pushed outward to the edges!
     % \hspace{0.5in}
    % \hspace{0.15in}
     \begin{subfigure}[b]{0.32\textwidth}
         \centering
         \includegraphics[width=1.0\linewidth]{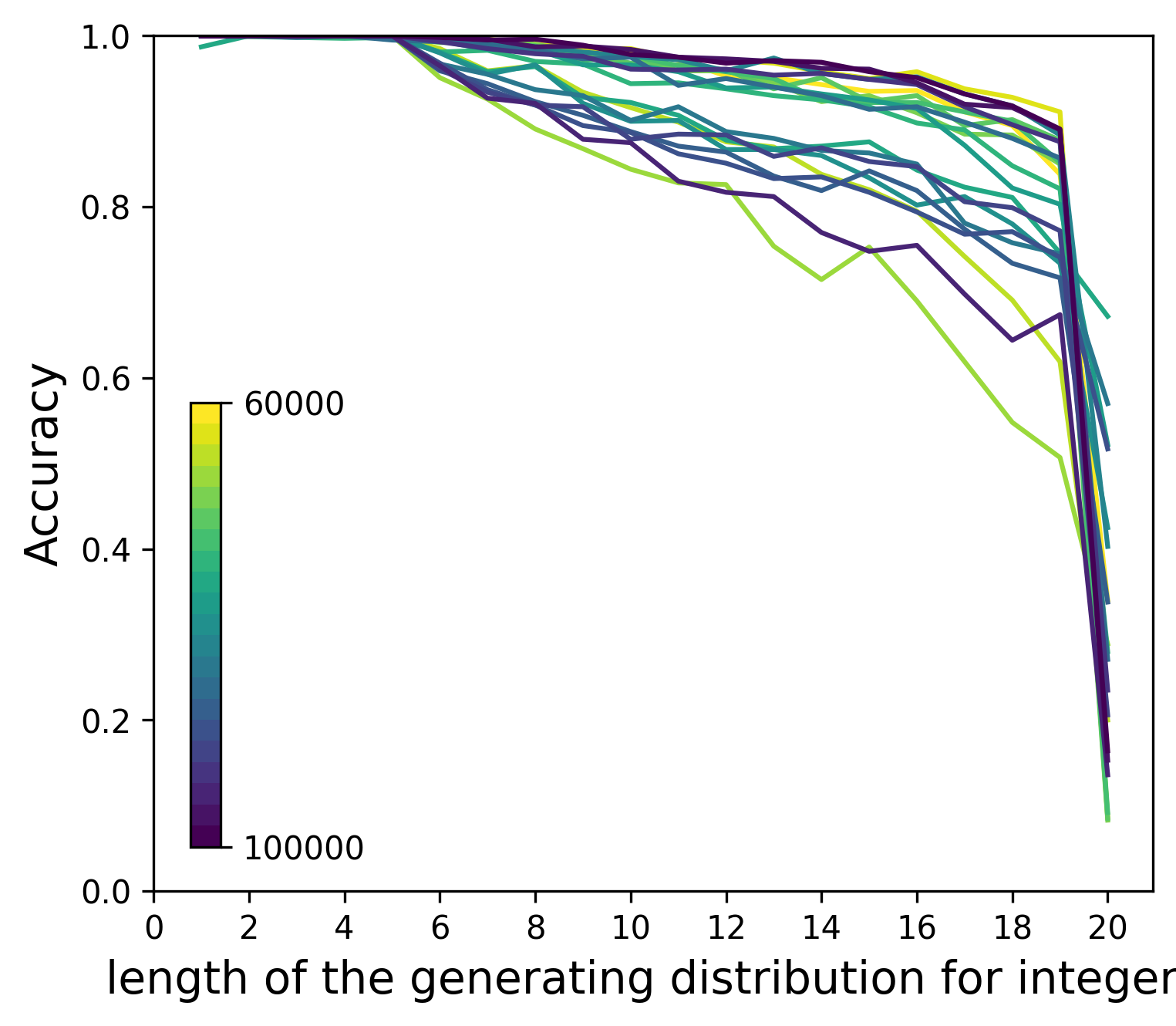}
         \caption{}
         \label{Mult_RPE}
     \end{subfigure}
    % \hspace{0.15in}
     % \\
     \begin{subfigure}[b]{0.32\textwidth}
         \centering
         % Linewidth is the now the unit for half a page.
         \includegraphics[width=1.0\linewidth]{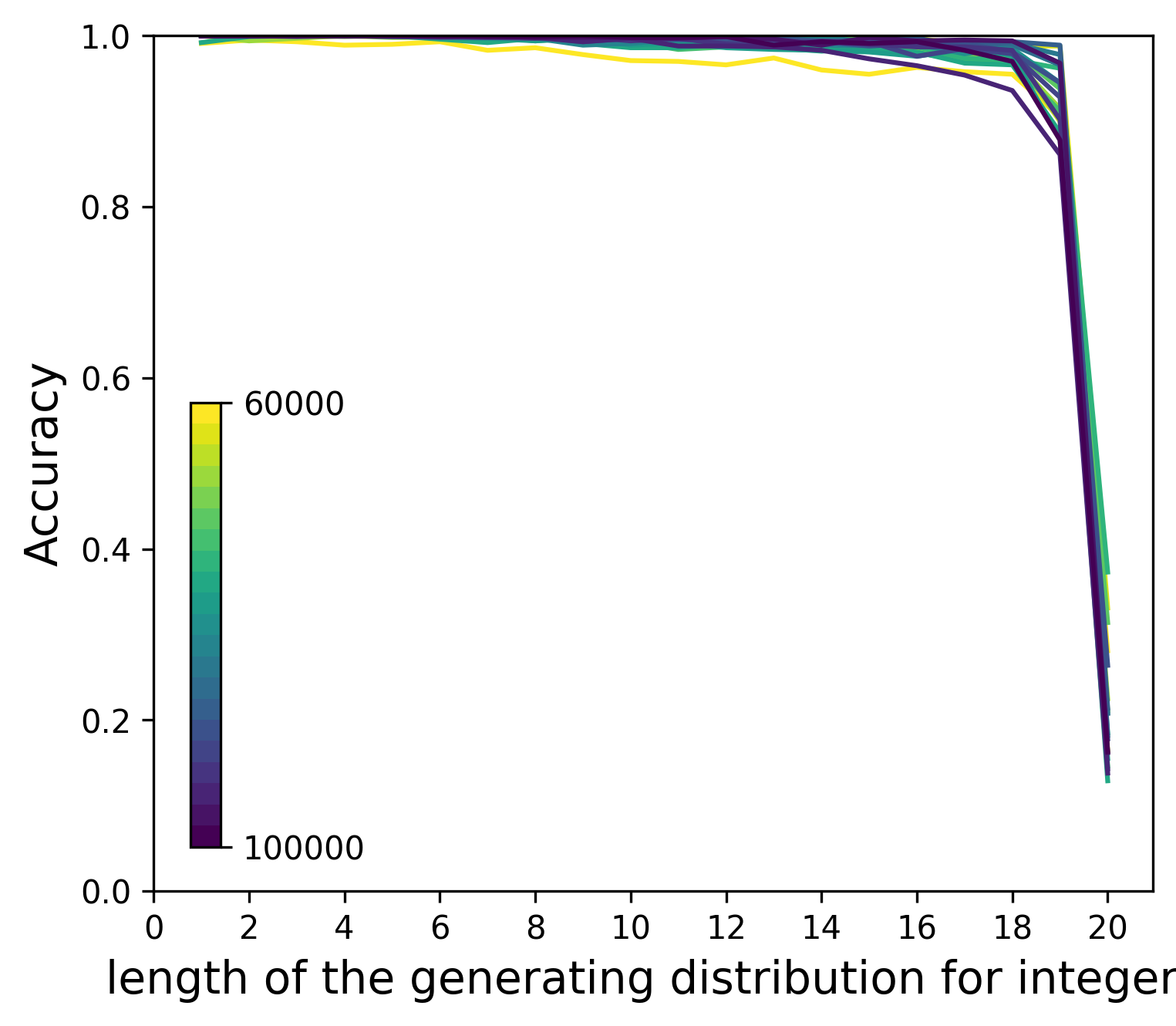}
         \caption{}
         \label{Mult_UPE_set}
     \end{subfigure}
     %\hfill % Remove this comment if you want the figures to be pushed outward to the edges!
     % \begin{subfigure}[b]{0.33\textwidth}
     %     \centering
     %     \includegraphics[width=1.0\linewidth]{figs/Mult_CPE_layers.png}
     %     \caption{}
     %     \label{Mult_UPE_layers}
     % \end{subfigure}
        \caption{\textbf{(a)} Accuracy of a BERT model with APE for single-digit $\times$ multi-digit multiplication when trained only up to 5-digit multiplicands. \textbf{(b)} Same setting as in (a) but with RPE. \textbf{(c)} Same setting as (a) but with our proposed UPE. Using the uniform symmetry naturally gives advantages over RPE. 
        % \textbf{(d)} Using a different layer for uniform positional encoding, and RPE on top of it results in pretty close performance.
        }

\end{figure}

\begin{figure}[t!]
     \centering
     \begin{subfigure}[b]{0.245\textwidth}
         \centering
         % Linewidth is the now the unit for half a page.
         \includegraphics[width=1.0\linewidth]{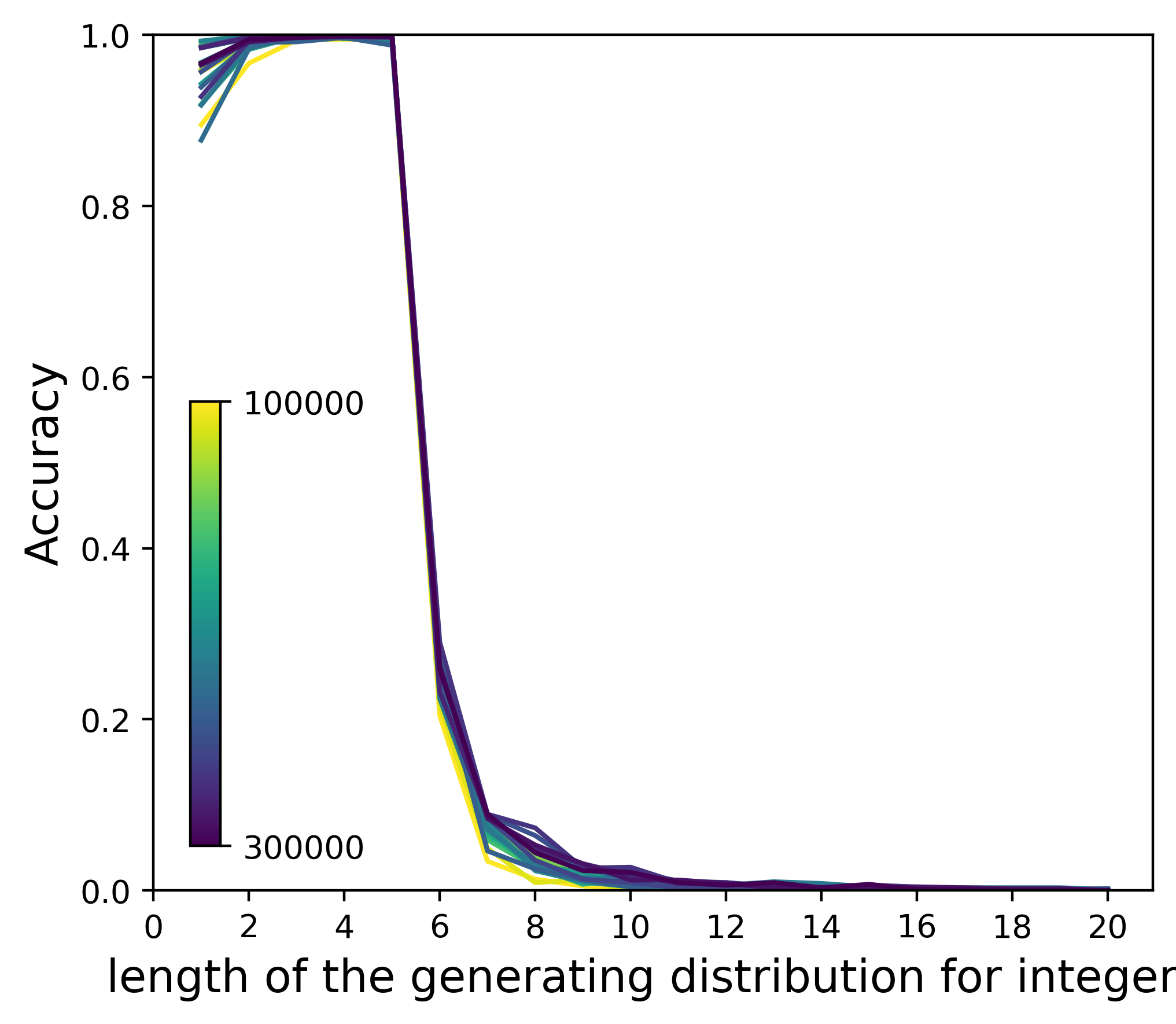}
         \caption{}
         \label{Mult3_RPE}
     \end{subfigure}
     \begin{subfigure}[b]{0.245\textwidth}
         \centering
         \includegraphics[width=1.0\linewidth]{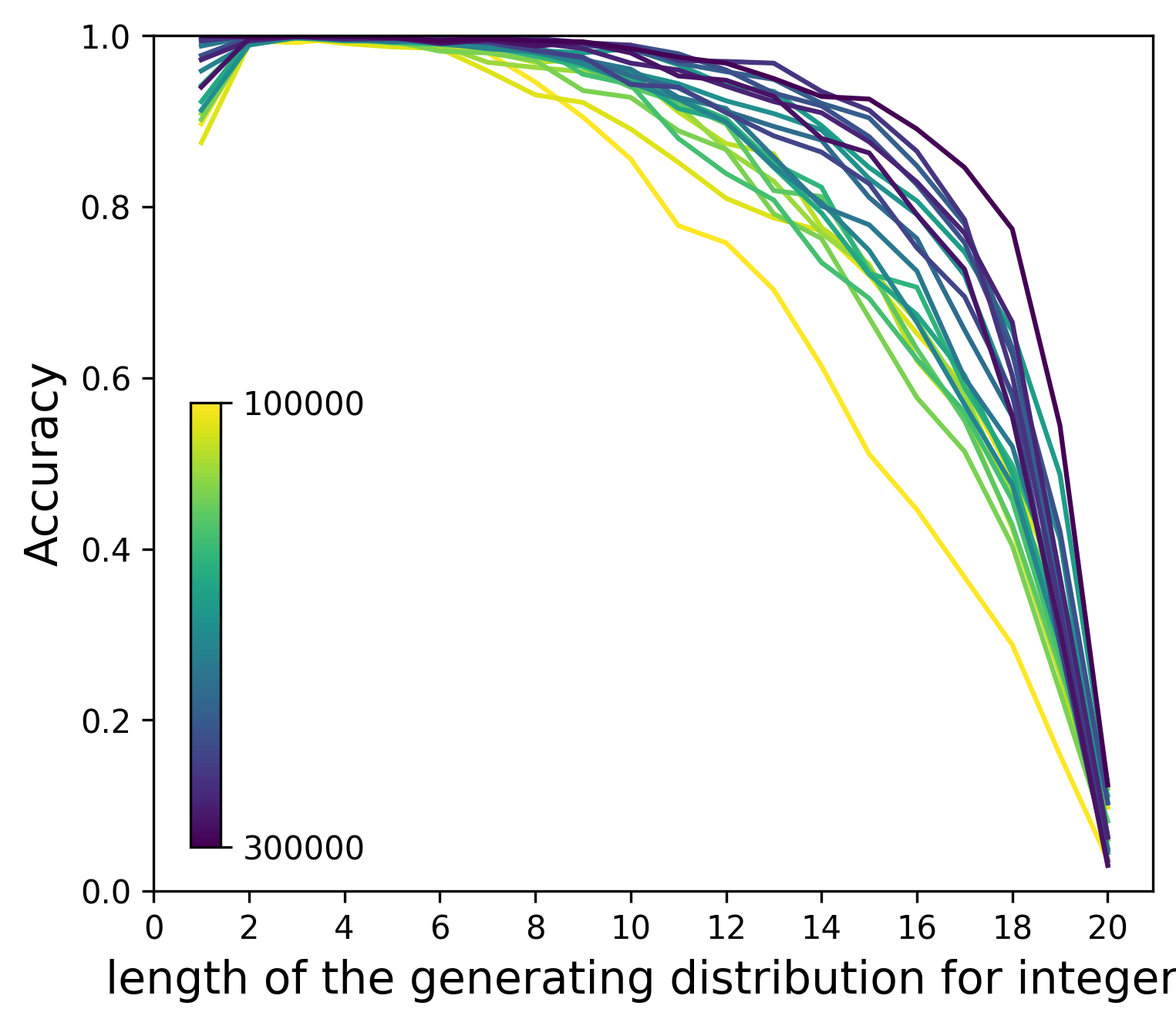}
         \caption{}
         \label{Mult3_UPE}
     \end{subfigure}
      \begin{subfigure}[b]{0.245\textwidth}
         \centering
         % Linewidth is the now the unit for half a page.
         \includegraphics[width=1.0\linewidth]{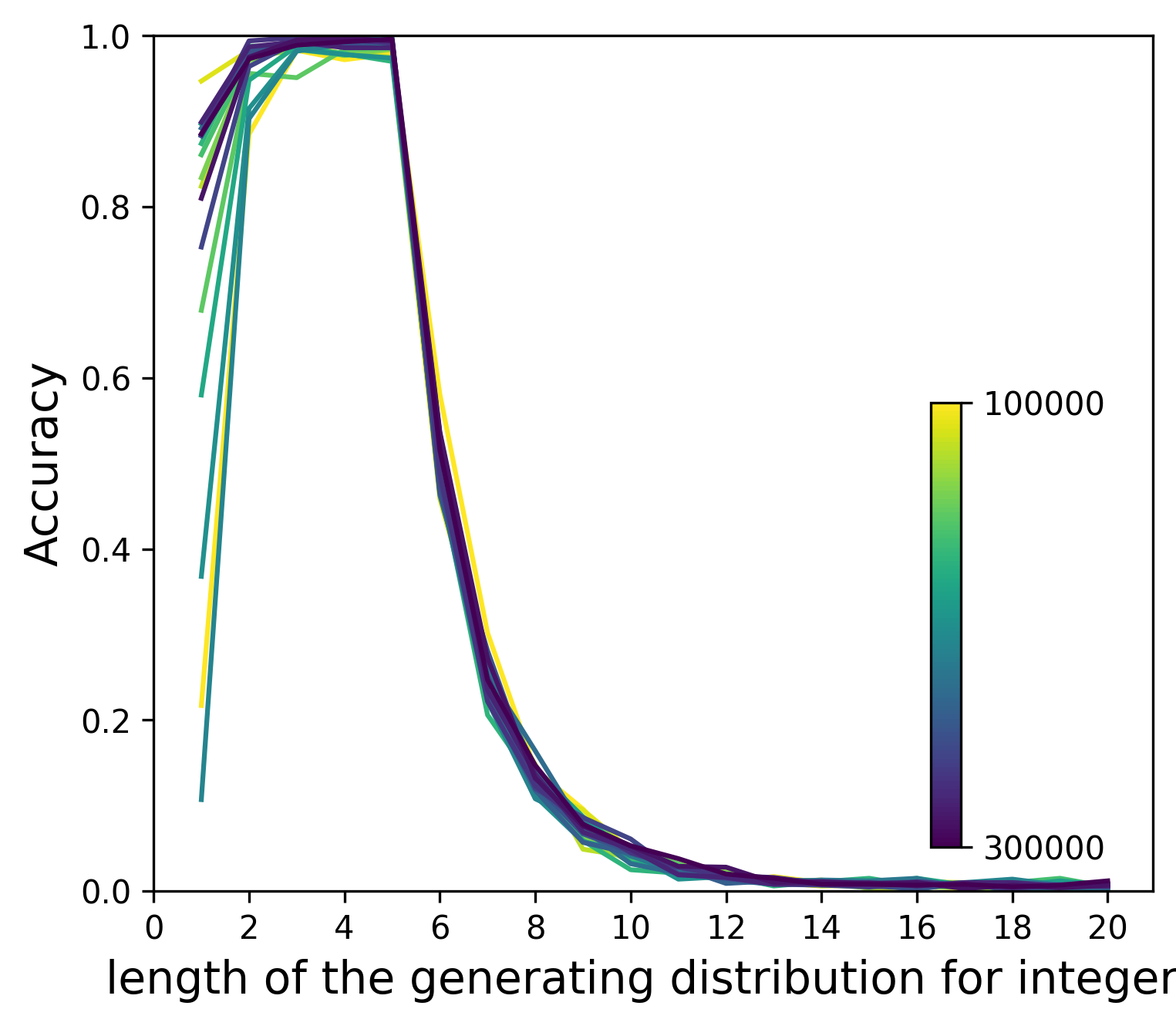}
         \caption{}
         \label{Mult3_APE}
     \end{subfigure}
     \begin{subfigure}[b]{0.245\textwidth}
         \centering
         % Linewidth is the now the unit for half a page.
         \includegraphics[width=1.0\linewidth]{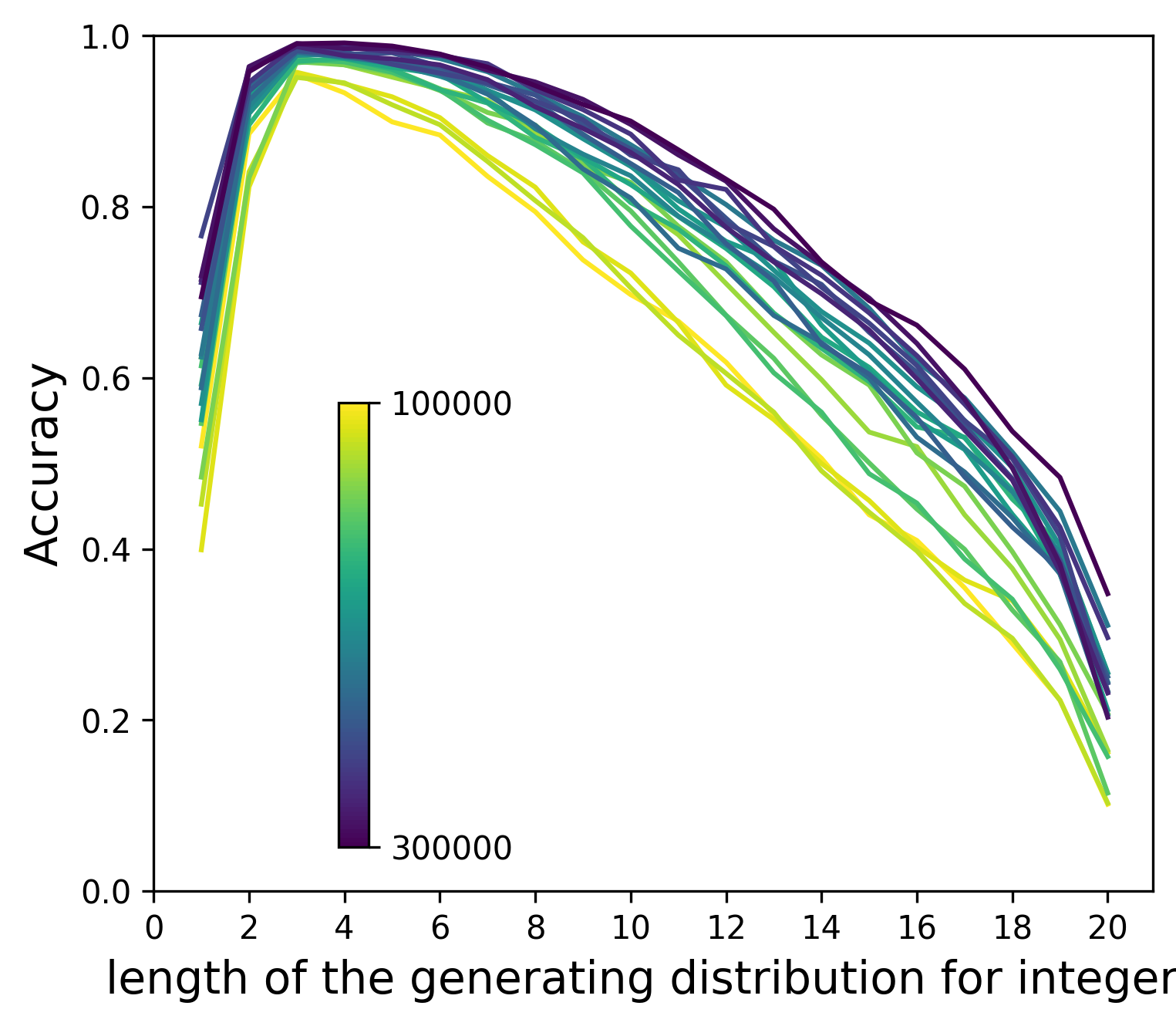}
         \caption{}
         \label{RPE3_augmented}
     \end{subfigure}
     % \begin{subfigure}[b]{0.33\textwidth}
     %     \centering
     %     % Linewidth is the now the unit for half a page.
     %     \includegraphics[width=1.0\linewidth]{figs/Multip_3digits_CPE_layers.png}
     %     \caption{}
     %     \label{Mult3_UPE_layers}
     % \end{subfigure}
        \caption{\textbf{\textbf{(a)}} Accuracy of a BERT model with RPE for 3-digit multi-digit multiplication when trained only up to 5-digit multiplicands. \textbf{(b)} Same setting as in (a) but with our new positional encoding. The difference between UPE and RPE becomes more apparent as the length of the multiplier increases. \textbf{(c)} The model using APE with the privilege of augmented data. \textbf{(d)} The model with RPE and trained with augmented data. Both models with APE and RPE are outperformed by our UPE.
        % \textbf{(c)} Using a separate layer for UPE again offers similar performance.
        }
\end{figure}

\paragraph{(M1) APE and RPE fail at length generalization.}
We evaluated both absolute and relative positional encodings, in a single-digit multiplier scenario, as provided in \Cref{Mult_APE,Mult_RPE}. As anticipated, APE fails to generalize on the unseen positions, and RPE shows a decline in accuracy. In the three-digit multiplier scenario, as carries become more challenging, RPE cannot generalize to longer multiplicands as shown in  \Cref{Mult3_RPE}, which is consistent with the results of prior work \cite{jelassi-length}. RPE is unable to preserve the functionality of the multiplier on those positions of the multiplicand padded at training.

\paragraph{(M2) UPE (along with RPE) is effective length generalization.}
In contrast, our UPE imposes this constraint and shows a significant improvement over RPE in the single-digit multiplier scenario (\Cref{Mult_UPE_set}). This advantage becomes more apparent for the three-digit multiplier (\Cref{Mult3_UPE}); Once we incorporate the three uniform positional vectors from \Cref{upe} to the multiplier's digits and RPE for the rest, they enable the model to maintain 90\% of its original accuracy at the length of 16. As far as we know, this is the first positive result for length generalization in multiplication purely from architectural modifications. For the same reason explained in \Cref{Section_RPE} for RPE, the fall in accuracy in 20-digit sums is inevitable considering our input format.

\paragraph{(M3) Augmentation fails in multiplication.}
When some zeros are added to the end of the multiplicand while keeping the multiplier unchanged, the result is a shifted answer by zeros. This highlights a symmetry of the problem, aligned with the translational symmetry of RPE for the multiplicand and the constant functionality of UPE for the multiplier. We used this to augment the training data and see if it improves results for models with absolute and relative positional encodings. However, as shown in \Cref{Mult3_APE}, the model utilizing APE fails similarly to the same attempt in the addition task. As for the model with RPE, despite extensive training data, it still falls short of the performance of our UPE. See \Cref{RPE3_augmented}.

\subsection{Extension: Applying to Data with Text + Numbers}\label{Section:text}

\begin{figure}
    \centering
    \includegraphics[width=\linewidth]{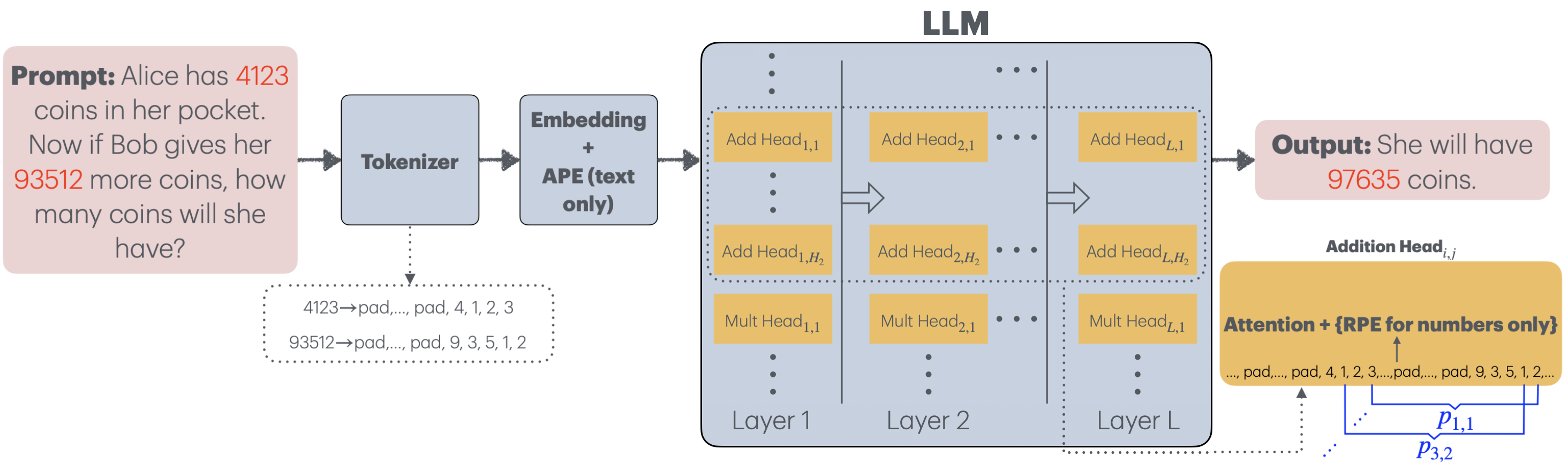}
    \caption{Extra attention heads utilized with pairwise positional encodings are integrated into the model to exploit structures across various tasks. The diagram illustrates how certain attention heads employ relative position encodings to enable length generalization for the addition task.}
    \label{fig:LLM_sketch}

\end{figure}

Our input format may initially seem limited to numeric experiments. However, we show how our proposed positional encodings can be effectively applied to a broader corpus, including text. We propose to set up the pairwise positional encodings in \Cref{pairwise} based on the significance of the digits within their integer; For instance, always the same pairwise positional vector is incorporated to the first digits of both integers regardless of their positions in the input. This approach leverages the fact that the structures in arithmetic tasks like addition and multiplication are grounded on the number system. As illustrated in \Cref{fig:LLM_sketch}, we employ separate positional encodings for text (using APE) and arithmetic calculations (using Pairwise PE). Since the latter must be implemented at the attention level, specific attention heads are designated for this purpose. For instance, \Cref{fig:LLM_sketch} demonstrates that attention labeled with "Add Head$_{i, j}$" utilizes RPE. These heads add RPE only to the pairwise interactions among the numerical parts of the input, as detailed in \Cref{pairwise}, while the non-numerical parts remain unaffected by RPE. Note that we do not mask any part of the input sequence, allowing all attention heads to access the entire input.

\begin{figure}
     \centering
     \begin{subfigure}[b]{0.32\textwidth}
         \centering
         % Linewidth is the now the unit for half a page.
         \includegraphics[width=1.0\linewidth]{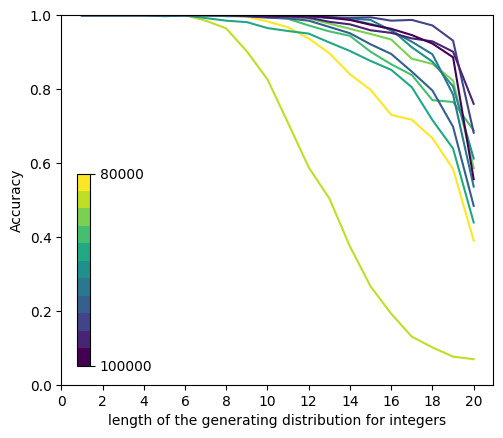}
         \caption{}
         \label{sum_text}
     \end{subfigure}
    \hspace{0.2in}
     \begin{subfigure}[b]{0.32\textwidth}
         \centering
         % Linewidth is the now the unit for half a page.
         \includegraphics[width=1.0\linewidth]{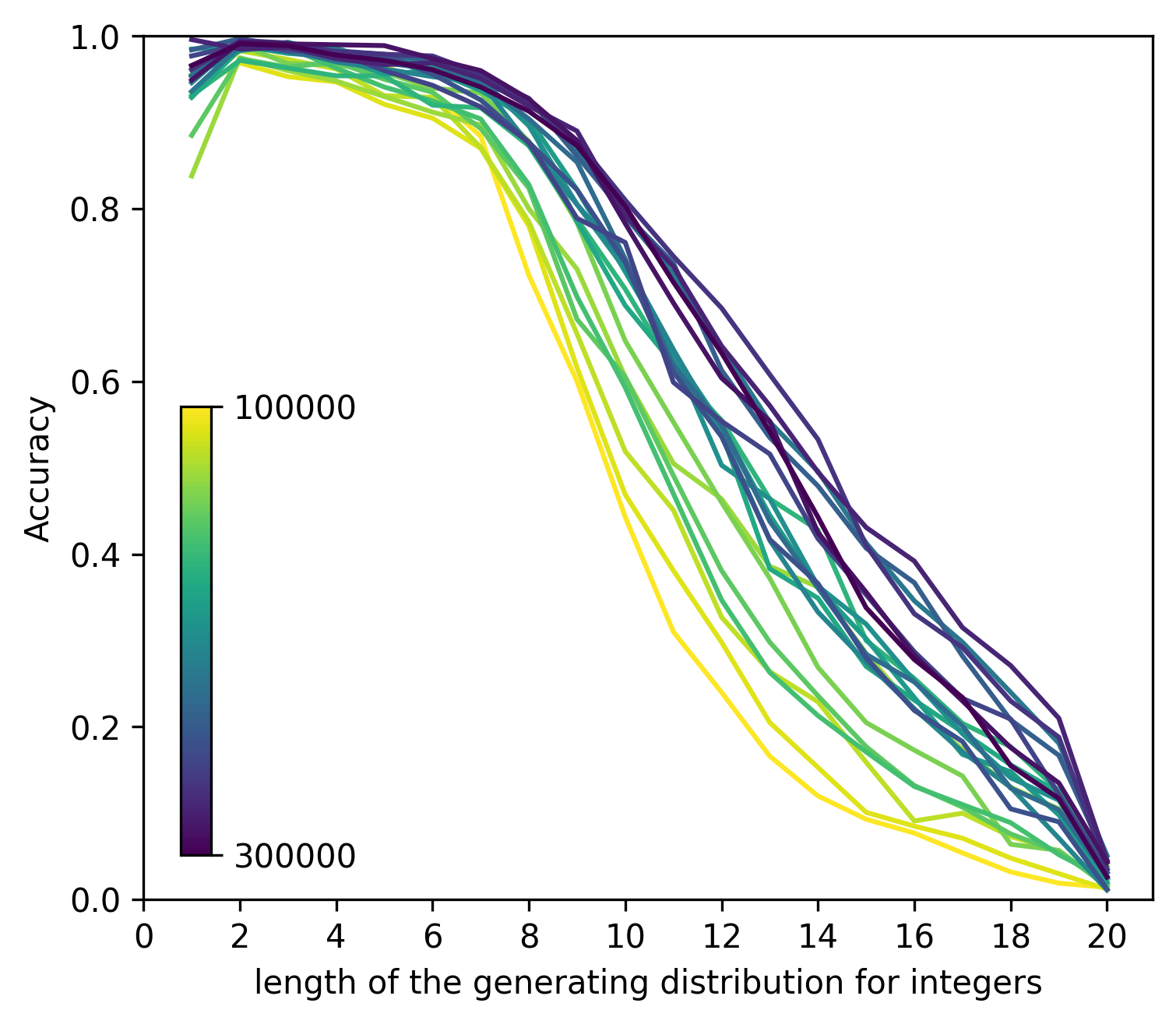}
         \caption{}
         \label{mult_text}
     \end{subfigure}
        \caption{\textbf{(a)} The generalization performance on the task of addition for a model that also allows for text with variable length is as good as the original implementation in \Cref{RPE_5to50}. \textbf{(b)} The trained model with text for the multiplication task also follows the same generalization trend as in \Cref{Mult3_UPE}.}

\end{figure}

\textbf{Addition and Multiplication in the presence of text.} While we include text in our experiments, attention heads are specifically trained to produce the correct answers for the arithmetic tasks. Although fully integrating this method into LLMs requires extensive resources, our current setting is sufficient to demonstrate its effectiveness and can be easily scaled up in future work. To test the robustness of our encodings in a more complex scenario, for both tasks of addition and multiplication, we altered formats in \Cref{eq:format_sum,eq:format_mult} to include up 20 random tokens before, between, and following the two integers, and as previously, the answer is fetched from the output of the position of the second integer. For the addition task, we modified RPE such that the same positional vector is always incorporated to the interaction of the positions of the least significant digits for both integers, and so on for other positions in spite if their variable distance in the input sequence. \Cref{sum_text} shows the method is as effective as the scenario without the text. During training, attention heads designated for arithmetic operations learn to ignore  the text, focusing solely on arithmetic tasks. In turn, for the multiplication task, analogous of this approach is applied to equip the model with UPE, labeled with "Mult Head$_{i, j}$" in \Cref{fig:LLM_sketch}, and \Cref{mult_text} depicts comparable generalization performance to \Cref{Mult3_UPE}. 

\section{Beyond Structural Symmetries: Complexity of Task}\label{Section: complexity}
So far, we have explored the importance of leveraging the structural symmetries. However, we may want the model to consistently work for all examples, not just those that are more probable. Here, we focus on analyzing the errors of these models, relating these observations to the complexity of the examples. Specifically, for the task of addition, we will assert that examples with longer dependencies across the input sequence are considerably more difficult for the model. Finally, we will show how to counteract this performance drop by adjusting the training distribution in favor of higher complexity examples. 

% % \surbhi{Title is not meaningful}
% \subsection{Quantifying Complexity} 
Typically, it is not necessary to consider all preceding digits of the two integers when calculating the result of the current position. Without carry-overs, the calculation simplifies to repeating single-digit additions across the sequence in parallel. However, as the chain of carries extends, managing them grows more challenging in parallel, as the model needs to track more positions in the input for each output position. We introduce a notion of complexity based on how far the model should look back to compute the carry-overs, suggesting that this window may not need to be large for most instances.
%This dependence is manifested through the the length of consecutive carry-overs that 
%Notice that it is the carry that induces the recurrent formulation described in the beginning of this section, without carry there in no need of taking $I_c$ into account in equation \eqref{eq:2}. 

% \begin{definition} \label{complexity_def}
%     Consider a task $T$ and a sequence ${\rm S}$ on which the task operates. Let $T_i({\rm S}_\sigma)$ be the $i$'th computed position of the output having only access to a part of the sequence, where ${\rm S}_\sigma$ means that only the positions inside $\sigma$ are observed. Also, let $T_i({\rm S})$ be the output when the entire sequence is observed. 
%     We define $\C_T({\rm S})$ as follows:
%     \begin{equation}
%         \C_T({\rm S}):= \max_i \big\{ \min\{ |\sigma|: T_i({\rm S}_\sigma) = T_i({\rm S}) \} \big\}
%     \end{equation}
    
%     Then for a domain $\D$ that ${\rm S}$ belongs to, we define $\C_T(\D):= \max_{{\rm S} \in \D} \big\{ \C({\rm S}) \big\}$.    
% \end{definition}
% For instance, for any $x^{(1)}, x^{(2)} \in \mathcal{D}$, to compute $x^{(1)}+x^{(2)}$ at each position $i$, we need to look at the values of $\C(\D)$  positions at most. 

\paragraph{Definition (\textit{complexity}).} 
\textit{For a general seq-to-seq task and a sequence ${\rm S}$ on which the task operates, 
% , meaning that for each element of the input there is a corresponding element in the output sequence,
we define 'complexity' as the maximum \textbf{dependency length}, i.e., number of positions such that their values are necessary to calculate the output of the current position, across the entire input.}

Consequently, for the task of addition, and within the seen domain that only contains samples with limited length $l_s = 5$, the complexity of any sequence is trivially upper bounded by $10$, the total number of positions. As another straightforward example, consider a scenario where no two carries occur consecutively. In this case, the required list of positions, $\sigma(i)$, would only include $i-1$, $i$, $i+l$, and $i+l+1$.

\begin{figure}
     \centering
     % You have to set the textwidths to sum to under 1 because Latex doesn't like it when the two are exactly 1. :(
     % This splits the figure into it to two half pages width figures.
     \begin{subfigure}[b]{0.32\textwidth}
         \centering
         % Linewidth is the now the unit for half a page.
         \includegraphics[width=1.0\linewidth]{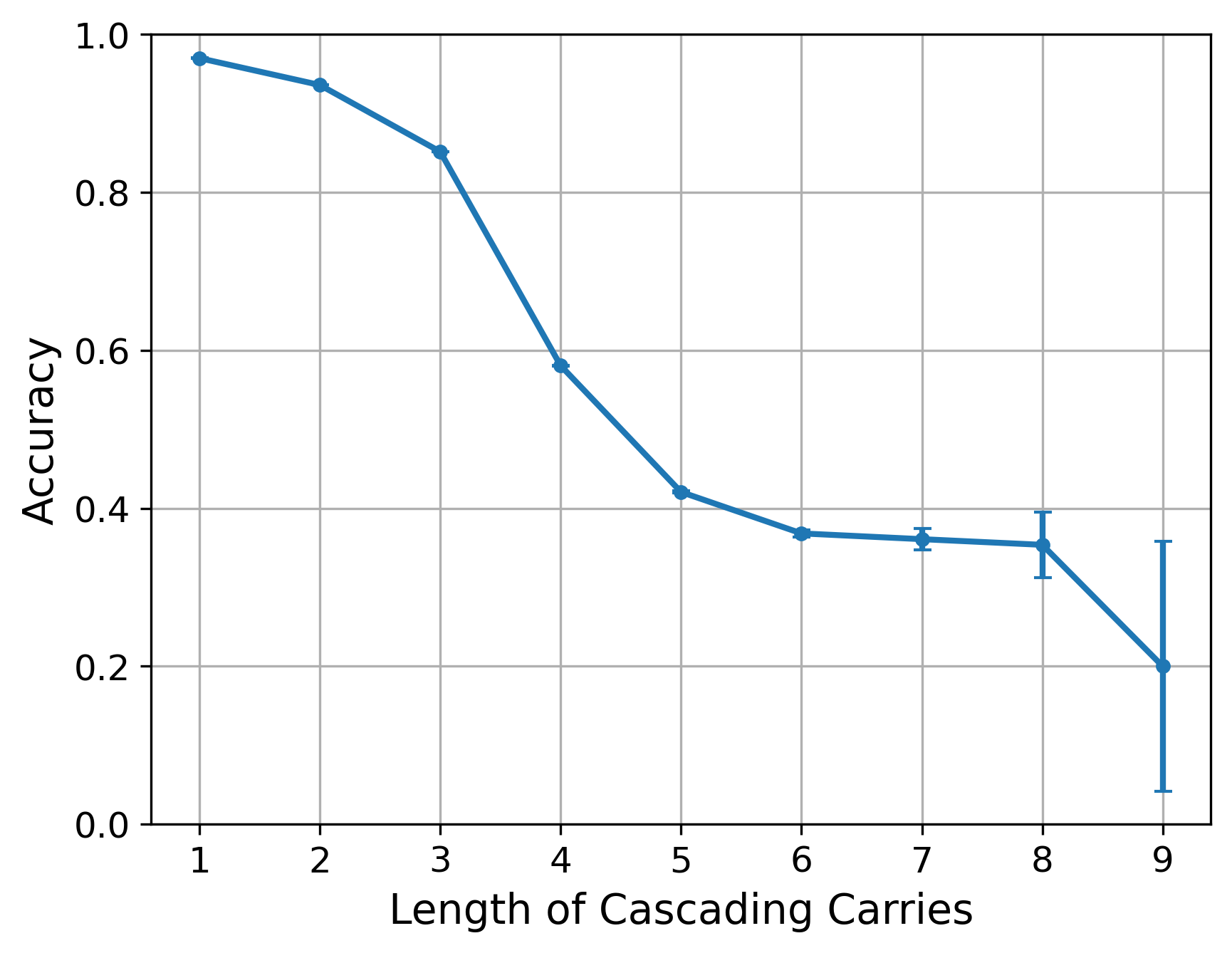}
         \caption{}
         \label{cascade}
     \end{subfigure}
     %\hfill % Remove this comment if you want the figures to be pushed outward to the edges!
     \begin{subfigure}[b]{0.32\textwidth}
         \centering
         % Linewidth is the now the unit for half a page.
         \includegraphics[width=1.0\linewidth]{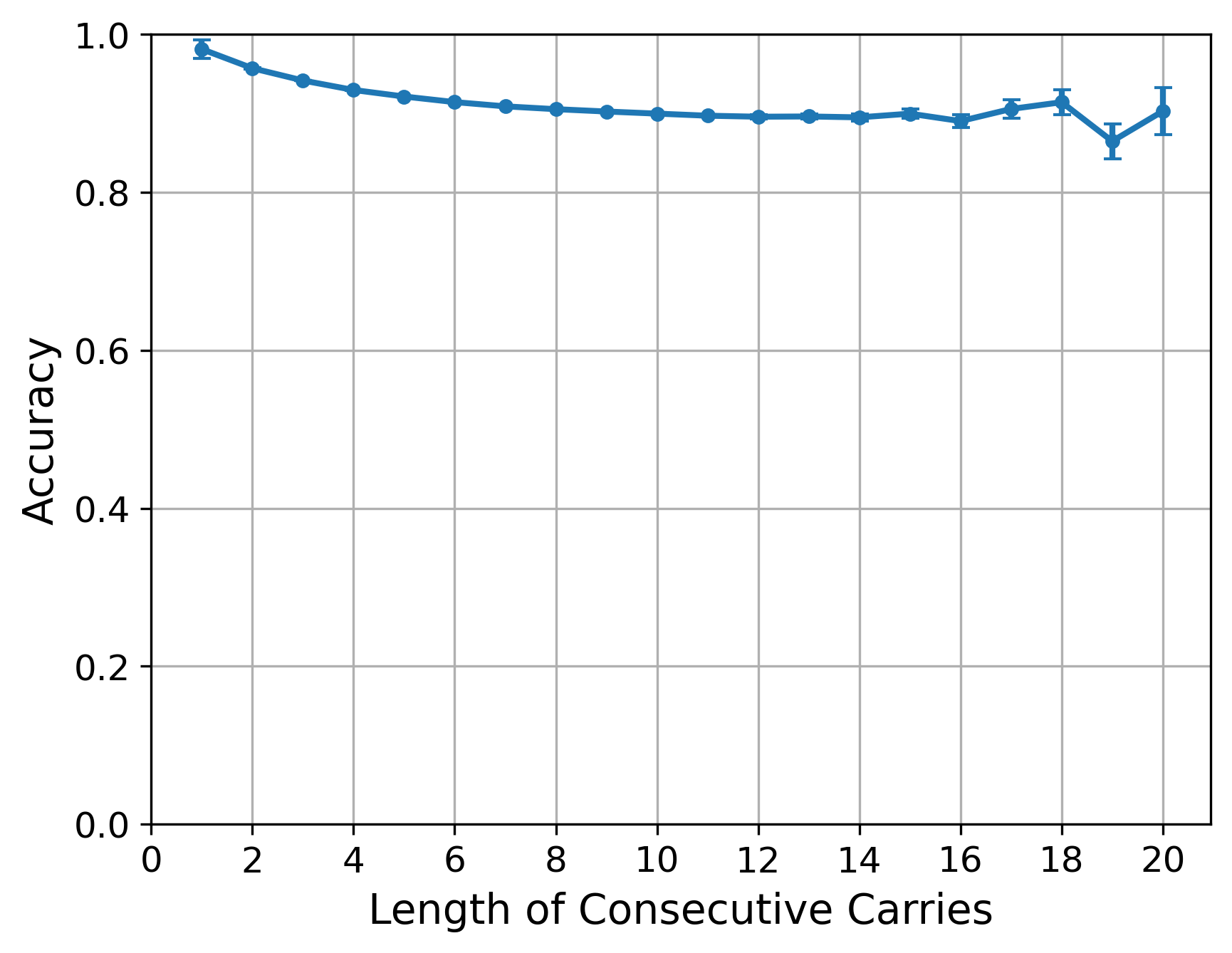}
         \caption{}
         \label{consecutive}
     \end{subfigure}
     \begin{subfigure}[b]{0.32\textwidth}
         \centering
        \includegraphics[width=1.0\textwidth]{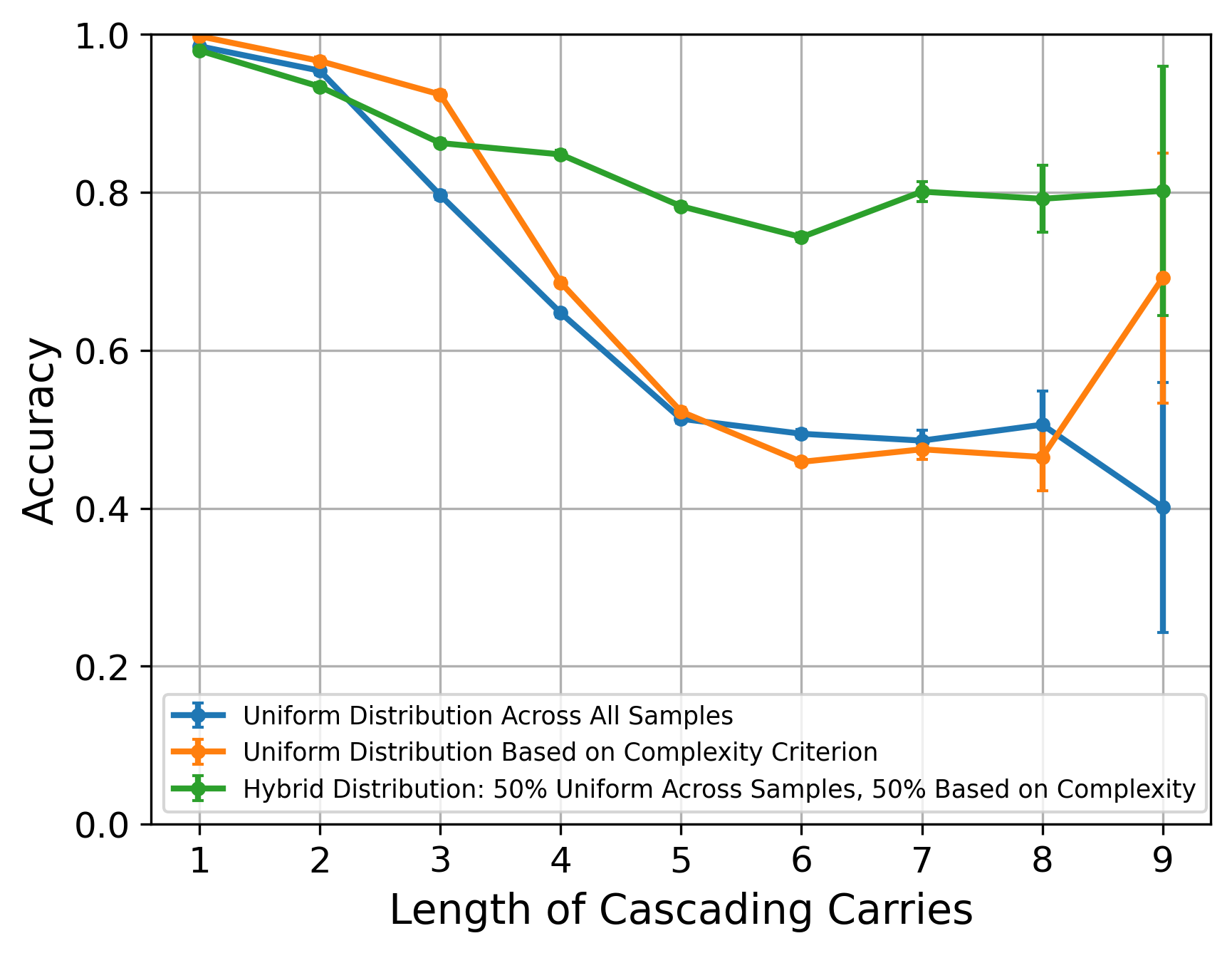}
         %\hfill % Remove this comment if you want the figures to be pushed outward to the edges!
         %\hfill % Remove this comment if you want the figures to be pushed outward to the edges!
        \caption{}
        \label{comp_40}
     \end{subfigure}
    \caption{\textbf{(a)} Cascading carries are noticeably hard samples for the model trained with RPE. \textbf{(b)} If the cascade effect is not taken into account, focusing only on a series of carries in a row fails to obtain harder samples. \textbf{(c)} The performance of the model trained with 100k samples chosen based on different mixtures of easy and hard examples. Exposure to more complex samples improves the accuracy on higher complexities.}
\end{figure}

\paragraph{(C1) Consecutive carries are not representative of complexity.} Last example might prompt one to think that the mere consecutive occurrence of carries qualifies for the long dependency complexity, and will lead to performance drop for the model. \Cref{consecutive} shows the performance of the model for consecutive carries of different lengths. As seen from the figure, the performance does not drop significantly as the length of the sequence of consecutive carries increases. This signifies the fact that consecutive carries do not accurately capture the level of difficulty for this problem.

Now we will quantify the complexity rule for addition. Consider a sample $(x^{(1)}, x^{(2)})$ and the sum at position $i$. If $x^{(1)}_{i-1} + x^{(2)}_{i-1} \geq 10$, the output at position $i$ is $x^{(1)}_i + x^{(2)}_i + 1$  regardless of the values of all previous positions. Similarly, when $x^{(1)}_{i-1} + x^{(2)}_{i-1} \leq 8$, the output at position $i$ is $x^{(1)}_i + x^{(2)}_i$. Thus, the output at position $i$ depends on previous positions when the carry cannot be determined without knowing the prior carries. This is the number of consecutive indices which sum to 9 that are triggered by a carry, leading to a \textit{cascade effect}. \Cref{sum_example} shows an example where carries cascade for three stages, from the first to the fourth position.

% We omit the argument of $\C$ when discussing the complexity of the addition. Assuming that a domain $D_s$ has complexity $\C(\D_s)$,
% and for an arbitrary domain $\D$ that the samples are instantiated from,
% \begin{align*}
%     \sigma_d(i) = &\big( a^{(1)}_i = i, a^{(2)}_i = i + l + 1, {\rm C}_i = (i - d, \cdots, i - 1, i + l - d+1, \cdots, i + l) \big)
% \end{align*}

\paragraph{(C2) Cascading carries quantify failure to higher complexity.}
In \Cref{fixed_map}, we show that RPE learns a fixed list of relative positions for computing every digit of the output under a restricted complexity constraint, which is met during training (within the seen domain that only contains samples with limited length $l_s = 5$, the complexity of any sequence is trivially upper bounded by $10$). Nevertheless, this does not address the scenario where test samples exceed the constraint. If the model has seen cascades up to a certain length at training, it may not be able to generalize to samples with larger cascade sizes.
% For example, if training is done over sample of length at most 5, and hence the maximum cascade size seen is also 5, then at test time the model may not be able to generalize to samples that have cascades of length 6 or more.
\Cref{cascade} demonstrates that an increase in cascade length causes a significant decline in the performance. However, as will discuss in the next part and in \Cref{carry_dist}, higher complexity samples are rare; in fact, most of the cascade lengths are covered by a small number and thus lower complexity. Consequently, given that our evaluation are based on the uniform distribution over the unseen domains, higher complexity samples do not tangibly affect the performance in expectation.

\begin{figure}

\end{figure}

\paragraph{(C3) Models trained with more complex samples generalize better.} In \Cref{comp_40}, in a bid to expose the models to more complex samples during training, we adjusted the training distributions. 
Beyond the uniform distribution across all samples, we trained another model using samples that are selected equally for their complexity (namely, uniform distribution on the complexity). We further trained a third model based on a hybrid approach, where each sample is chosen based on a 50/50 probability from the first two distributions. These models were then tested on 40-digit sums with varying cascade lengths. Allowing the model to observe more samples from higher complexities aids in preserving its accuracy even on cascading lengths above 5 (the maximum length observed in-distribution). Notably, when a sufficient number of samples from higher complexities are observed (approximately 16k of samples with a cascade length of 5 in the uniform complexity scenario, and 8k in the mixed distribution scenario), the uniform complexity loses its advantage in higher complexity areas. The model trained with the mixed distribution outperforms the other two at most lengths. We conclude that smaller cascades are in fact more beneficial for learning the positional aspect of the task. This aligns with the findings in \cite{shen2023positional} about assigning higher weights to shorter examples. Moreover, samples with higher complexity are  necessary to enhance the performance on higher complexities.

% \begin{figure}[h!]
    
% \end{figure}

% \begin{figure}[h!]
%     \centering
%     \begin{subfigure}[b]{0.4\textwidth}
%          \centering
%          \includegraphics[width=1.0\linewidth]{figs/complexity_1000.png}
%          \caption{}
%          \label{comp_10}
%     \end{subfigure}
%     \begin{subfigure}[b]{0.4\textwidth}
%         \centering
%         \includegraphics[width=1.0\textwidth]{figs/complexity_100000_all_40.png}
%         \caption{}
%         \label{comp_40}
%     \end{subfigure}
%      %\hfill % Remove this comment if you want the figures to be pushed outward to the edges!
%     \caption{\textbf{(a)} The result of the three models trained with 1000 samples chosen based on different criteria. The exposure to more complex samples improves the accuracy on higher complexity, in both in-distribution and out-of-distribution regions. \textbf{(b)} The result of the three models trained with 100k samples chosen based on different criteria. This figure underscores the importance of observing sufficient number of less complex samples before jumping into more complex ones for a better overall performance.}
% \end{figure}

\section{Theoretical Results} \label{label:theory}
We theoretically validate our findings using a one-layer transformer model trained on a simple regression task that captures translational symmetries. We show that in the population regime, i.e., having access to infinite amounts of data, when the transformer is trained by gradient descent (with weight decay), encoding these symmetries into the structure of the positional vectors through RPE will provably lead to length generalization on the unseen positions. However, in the same setting, the model with APE, which is oblivious to translational symmetries, will fail to generalize on the unseen positions. We further show that APE fails even if we use data augmentation to help the model learn the translational symmetries during training. We defer the proofs to \Cref{app:proofs}.

\paragraph{Task description.} We consider a seq-to-seq regression task consisting of an input sequence $(x_1, x_2, \cdots, x_n)$ and an output sequence $(y_1, y_2, \cdots, y_n)$, where for $i \in [n]$:

% \begin{equation} \label{regression_y_x}
% y_i = \sum_{r = -\tau}^{\tau} c_r \langle \theta, x_{i+r} \rangle.
% \end{equation}

\begin{equation}\label{regression_y_x}
    y_i = \alpha \langle \theta, x_i \rangle + \beta \langle \theta, x_{i-1} \rangle + \beta \langle \theta, x_{i+1} \rangle
\end{equation}

Here, $\alpha$ and $\beta$ are some constants, and we assume $\| \theta \| = 1$. 
% To keep the task simple, we constrain the output at each position to depend only in the input at the same position as well as the immediate neighbors. 
The translational symmetries should now be clear from \Cref{regression_y_x}; if we apply a (circular) shift to the input, then the output will undergo the same type of shift.

We consider a length-generalization scenario where, for a given integer $n_1 \leq n$, the training set will include sequences $(x_1, x_2, \cdots, x_n)$ where the values at the first $n_1$ positions are generated according to the Gaussian distribution, i.e. for $i \in [n_1]: x_i \stackrel{\rm{i.i.d.}}{\sim} \calN(0, \mathbb{I}_d)$, and the rest are zero-padded, i.e. for $i > n_1$ $x_i = \bzero \in \mathbb{R}^d$. At test time, however, the trained model will be evaluated according to the inputs $(x_1, x_2, \cdots, x_n)$ where the values at \emph{all} the positions are generated according to the Gaussian distribution. We can view the 0 padded positions as the "unseen" positions. To measure the performance, we consider the MSE loss $\ell_i = (\hat{y}_i - y_i)^2$, where $\hat{y}_i$ is the model's prediction at position $i$. 
We will train a simplified linear Transformer on this task:

\paragraph{Linear Attention model.} One layer of a transformer architecture, specifically the \textit{linear} modification (no Softmax bottleneck), has been deployed as the main tool in some recent work \cite{vonoswald2023transformers, mahankali2023step, ahn2024linear} to analyze the dynamics of training. A Linear Transformer is defined by the following parameters: three matrices including a query matrix $W_Q \in \R^{d \times d}$, a key matrix $W_K  \in \R^{d \times d}$, and a value matrix $W_V  \in \R^{d \times d}$. Given the sequence of tokens $\{x_1, \cdots, x_n\}$ the output from the attention head is then passed through a linear projection $h \in \mathbb{R}^d$ to obtain the scalar outputs of the linear attention:

$$ \hat{y}_i = h^T \sum_{r=1}^n (W_V x_r) (W_K x_r)^T (W_Q x_i)  = h^T \sum_{j=r}^n (W_V x_r) (x_r^T W_k^T W_Q x_i) $$

To encode the positional information, positional vectors $\{p_1, \cdots, p_n\}$ are added to the tokens: $x_i \leftarrow x_i + p_i$. Furthermore, for our analysis, we study a simplified one-layer transformer with \textit{factored linear} attention, a model commonly studied in prior work \cite{stat_bhattacharya, stat_Goldt, jelassi_vision}:
\begin{equation}\label{factored}
    \hat{y}_i = h^T W_V \sum_{r=1}^n x_r (p_r^T W_K ^ T W_Q p_i)
\end{equation}

In this model, the positional vectors and the input are decoupled. This is to isolate the attention to only depend on the position, as to be consistent with the task. This simplification abstracts the learning of position encodings which is the main focus of our study
We also will demonstrate in \Cref{RPE_proof} that a modified version of this model, expressed as: $\hat{y}_i = h^T W_V \sum_{r= 1}^n x_r (p_{i-r}^T W_K^T W_Q u)$, where $u$ is a constant vector, is the appropriate abstraction for RPE.

For simplicity, we assume that $W_K = W_Q = W$. Besides, as we aim to study the structures of the learned positional vectors, we will absorb the attention weights into the positional vectors without altering the representational capabilities of these models. This approach is equivalent to fixing the attention weights in the model, thereby training solely on the positional encodings. Note that for the model to achieve generalization on all positions, $h^T W_V$ must get aligned with $\theta$, $p_i^T W^T W p_i \propto \alpha$, $p_i^T W^T W p_{\{i-1, i+1\}} \propto \beta$, and zero otherwise. For RPE, this similarly requires $p_0^T W^T W u \propto \alpha$, $p_{\{-1, +1\}} W^T W u \propto \beta$. Having this said, we will prove the following proposition:

\begin{proposition}[informal] \label{alltheory}
For the seq-to-seq regression task, the transformer model after training with GD with infinitesimal weight-decay on the positional vectors in the population regime:
\begin{itemize}[leftmargin=3ex,noitemsep]
    \item (APE) After training we have $p_k \approx 0$ for all $k \not \in [n_1]$. Hence, APE fails at generalization.
    \item (APE with augmentation) For $\beta = 0$, after training $|p_i^\top p_j|$ is large for all $i \neq j$ and $p_i^\top p_i = \alpha$.  Hence, APE with augmentation fails to generalize.
    \item (RPE) After training $ \langle v, \theta \rangle p_0^T W^T W u= \alpha, \langle v, \theta \rangle p_{-1}^T W^T W u = \langle v, \theta \rangle p_{1}^T W^T W u = \beta$, and $p_j \approx 0$ for all $j \not \in \{-1, 0, 1\}$, thus the model generalizes.
\end{itemize}
\end{proposition} 

The key idea is that APE does not receive any useful signals on the padded positions during training, and therefore, does not learn how to predict on those positions. For APE with augmentation, while the inner products of a positional vector with its adjacent vectors match the underlying task, those further apart are unsupervised and thus may behave unpredictably. Finally, making the model relative through RPE resolves the issue of undesirable trajectory of inner products occurring for the model with APE, leading to full generalization. We refer the reader to \Cref{app:proofs} for additional details of the setup and full proofs.

\begin{figure}[t!]
     \centering
     \begin{subfigure}[b]{0.33\textwidth}
         \centering
         % Linewidth is the now the unit for half a page.
         \includegraphics[width=1.0\linewidth]{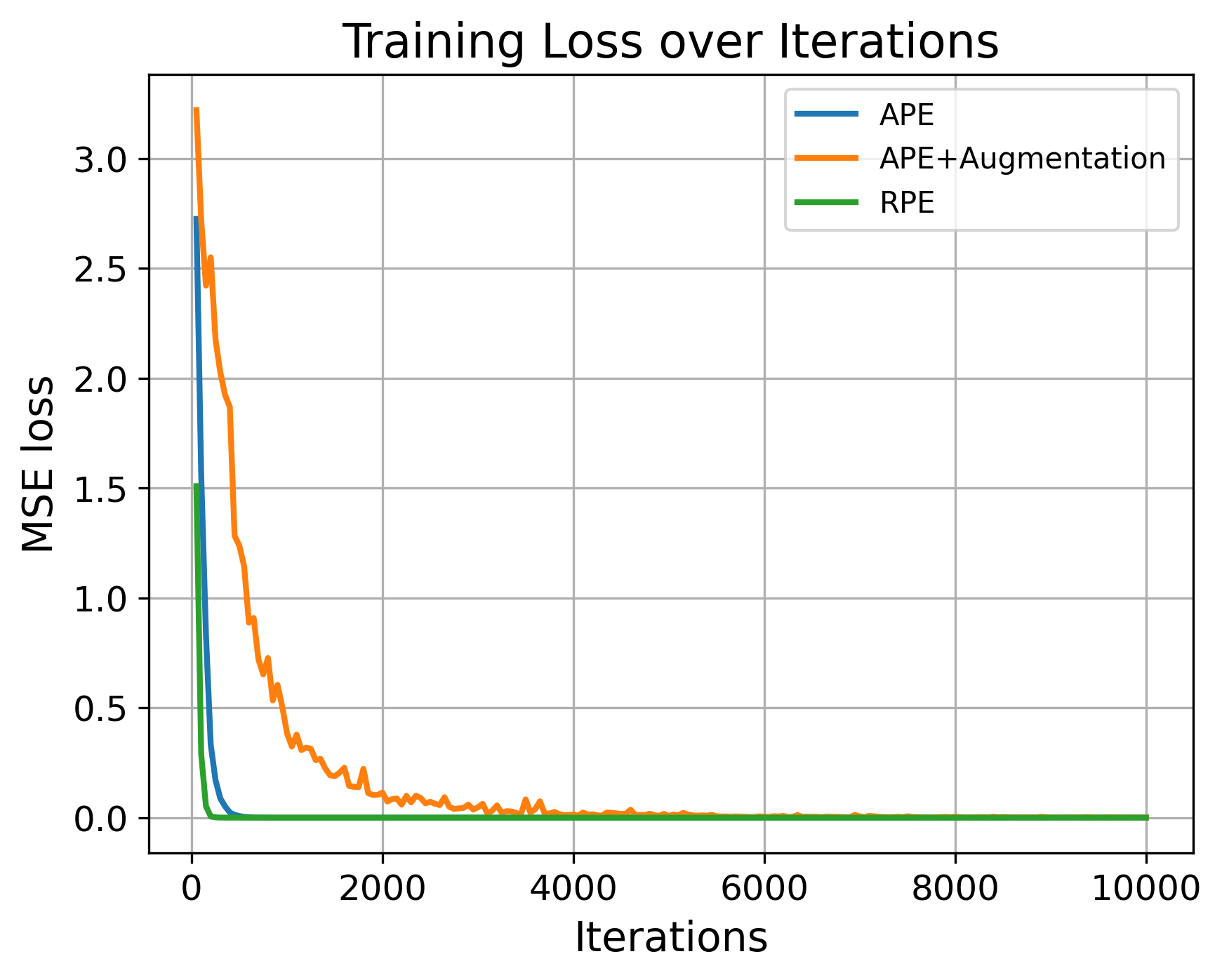}
         \caption{}
         \label{Loss_iter_m1}
     \end{subfigure}
    \hspace{0.5in}
     \begin{subfigure}[b]{0.33\textwidth}
         \centering
         % Linewidth is the now the unit for half a page.
         \includegraphics[width=1.0\linewidth]{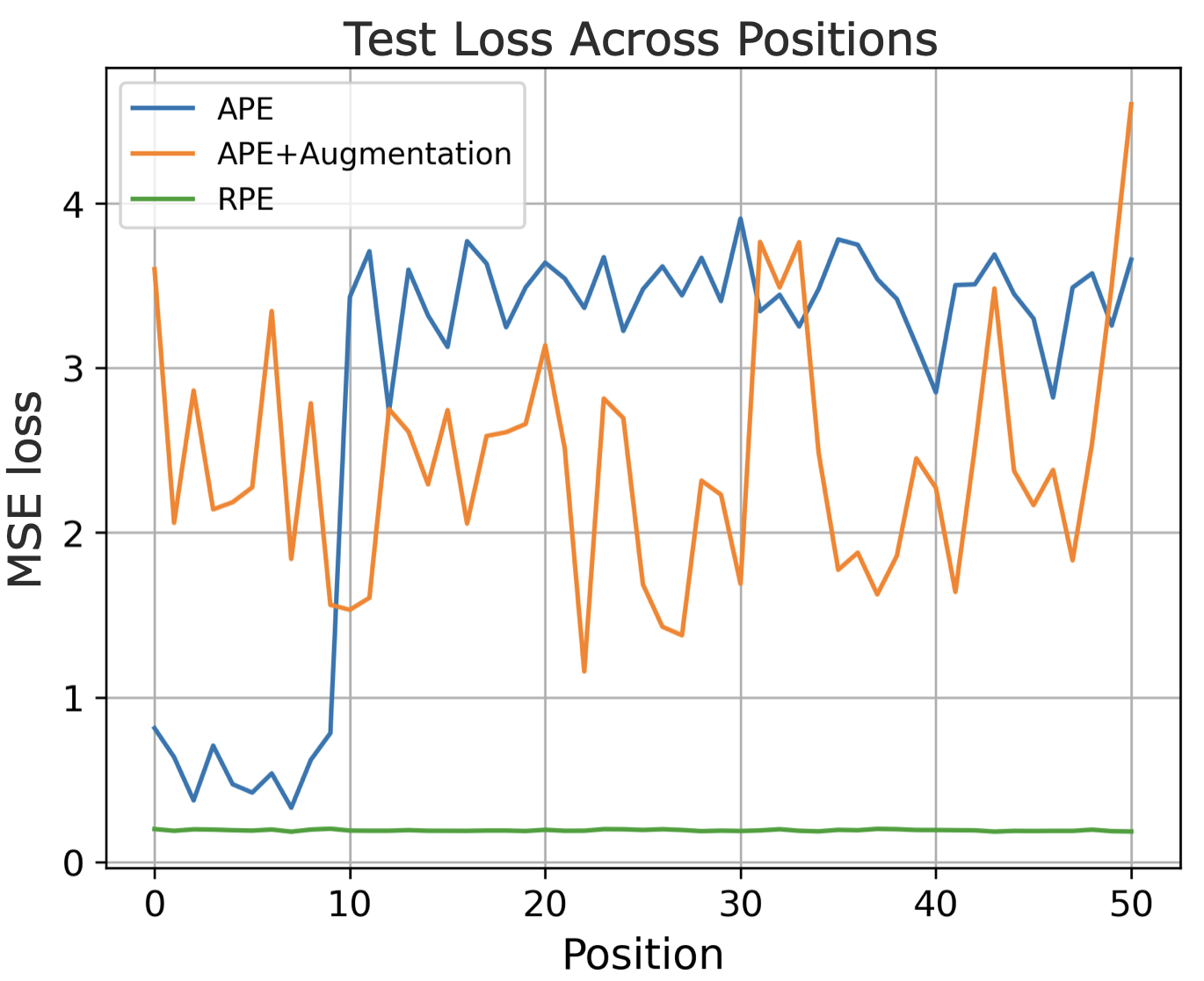}
         \caption{}
         \label{Gener_pos_m1}
     \end{subfigure}
        \caption{\textbf{(a)} All three scenarios perform perfect in-distribution, when only a part of the sequence is not padded. At test time, all positions get non-padded values. \textbf{(b)} The test loss at each position in the sequence. The model with RPE outperforms the other two. The inherent noise present in APE and APE with augmentation is a characteristic of their generalization capabilities, The plots are generated by averaging over 5000 samples at each position.}
\end{figure}

\paragraph{Experimental Validation.} To remain consistent with our assumptions, we conducted an experiment that positions are on a ring and for the task in \Cref{regression_y_x}. Here, $n= 51$, $n_1 = 10$, and $d= 200$. \Cref{Loss_iter_m1} shows that all three settings (APE, APE with augmentation, and RPE) converge to zero in their training loss. However, when we compute the test loss at each position in \Cref{Gener_pos_m1}, that is, $\E\{l_i\}$ according to the normal distribution at all positions and not only the allowed windows in training, both APE and APE with augmentation cases cannot generalize.

\section{Conclusion}
Our work provides a general approach for capturing structural symmetries in arithmetic tasks. This allows us to provide the first length generalization result for multiplication that relies only on minor architectural modifications. Exploring how to induce such structural symmetries beyond these basic arithmetic tasks, and in tandem with text data is a natural next direction. Future work will explore extending our approach to handle more than two integers and to combine multiple arithmetic tasks at once. Another possibility is exploring the compatibility of our methods with text-formatted arithmetic tasks. Orthogonally, our understanding of the current models is that, without seeing longer dependencies, the model cannot generalize to more complex samples. Next steps would be to understand the limitations of generalization abilities of these models on the complexity axis, and suggest potential architectural, data, and algorithmic fixes to improve their abilities.
%Representations learned  the model are not sufficient for a full generalization, and acquiring a deeper understanding of the task at more advanced levels requires exploring beyond the seen domain. 
% , the ability to handle more complex samples can be enhanced. We propose that the next step should be aligned with this strategy.

% \newpage
% \paragraph{Impact Statement.} This paper presents work whose goal is to advance the field of Machine Learning. There are many potential societal consequences of our work, none which we feel must be specifically highlighted here.

% \newpage
\bibliography{bibliography}
\bibliographystyle{alpha}

\newpage
\appendix
\section{Additional Related Work} \label{app:related-work}
\paragraph{Length generalization.} It has long been known that AI models struggle to fully grasp the underlying concept from limited experience \cite{FODOR19883, smolensky2022neurocompositional, li2022systematic}. For the special case of transformers, there is ongoing effort to understand to what capacity these models can learn the simple "algorithms" \cite{li2022systematic, liu2023transformers, liu2023exposing}. Large models tend to learn map functions rather than master the underlying tasks, failing at generalizing to variable lengths \cite{lee2023teaching}. Even fine-tuning pre-trained models and scratch-padding only provides marginal benefits \cite{nye2021work, anil2022exploring}. To mitigate this, recent work has suggested modifications to architectural components or data formats \cite{jelassi2023length, shen2023positional}, along with providing frameworks to argue when generalization may be feasible \cite{zhou2023algorithms} or infeasible \cite{abbe-lengen, mahdavi2023better}. In particular, \cite{zhou2023algorithms} show feasibility of length generalization in Transformers for tasks that use positions only via certain operations such as comparison, successor, predecessor, etc., and ignore the role of positional encodings.  

\paragraph{Solving arithmetic tasks with transformers.} Arithmetic tasks are among the tasks LLMs struggle with. \cite{yang2023gpt} studies the performance of GPT-4 \cite{openai2023gpt4} at solving arithmetic problems -- GPT-4 gets zero accuracy on 11-digit arithmetic operations, and the best methods have achieved $40 \sim 50 \%$ accuracy. In this regard, a growing body of work has focused on improving the representations of the numbers fed to the model \cite{wu2016google, sennrich-etal-2016-neural, thawani2021representing}, showing the sub-optimality of the current sub-word methods \cite{wallace2019nlp}. Another line of work has introduced the concept of intermediate steps by using chain-of thoughts reasoning \cite{wei2022ChainofThought, ouyang2022training}, or by gradually increasing the complexity of tasks \cite{wang2021survey, yang2023gpt}. Despite all these works, transformers continue to face challenges with high-complexity tasks \cite{dziri2023faith}. Authors in \cite{dziri2023faith} have investigated these problems in detail, and there are many causes for this including error-propagation, learning shortcut solutions and insufficient features, etc. It has also been observed that the distribution of the samples can greatly affect generalization \cite{lee2023teaching, shen2023positional}. For instance, \cite{shen2023positional} states that favoring smaller numbers more that larger numbers leads to better representations. Moreover, \cite{jelassi-length} claims that adding a few samples from the unseen domain can significantly improve generalization for longer lengths. Here, we maintain the uniform distribution over the data and contend that this approach requires a large amount of data to better generalize, but without entailing any longer samples. Our investigation centers on the two primary arithmetic tasks, addition and multiplication. In contrary to intermediate step approaches, we tackle this in an end-to-end manner. Another line of work has focused on the mechanistic interpretability  of transformers \cite{elhage2021mathematical} for the task of addition \cite{quirke2024understanding}, trying to understand the mechanism that a single-layer transformer implements this task in parallel. In the same vein, we study which input positions are necessary for each output position when doing the task in parallel, and on this basis, we build up a new approach that interprets our models.

\paragraph{Positional encodings.} In tackling the length generalization problem, various studies have altered the positional vectors by introducing new structures to them such as restiveness in \cite{rpe, huang2018music}. Other work has devised new schemes to leverage the positional information such as \textit{RoPE} \cite{su2023roformer}, and adding scalars to the attention scores in \cite{raffel2023exploring}. Some works have even suggested removing positional vectors \cite{kazemnejad2023impact}, or using randomized positional encoding \cite{ruoss2023randomized} for better generalization. Another line of work has focused on adding structures to attention maps, eliminating the need for positional vectors \cite{parikh2016decomposable, Press22ALiBi}.

\section{Experiment Details}\label{exp_det}
%and to compensate for the hardness of the task we increase of a attention layers from 6 to 9.
% We observed that providing more samples of in-distribution data improves the performance of out-of-distribution samples, so all the plots in figure 10 were trained with five times the amount of data compared to previous examples. 
For all the experiments we used Huggingface \cite{wolf2019huggingface} implementation of Bert \cite{devlin2018bert}. Throughout the addition section we used models with the same specifications: $6$ attention layer, $H= 8$ heads, embedding dimension $d_z= 768$, and dropout = $0.1$. A model with the same specification was also employed for the single-digit multiplier setting. In the 3-digit multiplier scenario, to compensate for the hardness of the task, we increase the number of attention layers from 6 to 9. We observed that we need more coverage for this setting , so all the plots in Figure 6 were trained with five times the amount of data compared to previous settings, i.e. 100k number of samples for the addition and the single-digit multiplier settings vs 500k number of samples for the 3-digit multipliers setting. Each element of the input is in the one-hot format, and the model will print the score for each digit of the output. All models are trained on the Cross-Entropy loss using AdamW \cite{loshchilov2019decoupled}, and from the scratch, with batch-size = $64$, learning-rate = $1{\rm e}-4$, and weight-decays in $\{3{\rm e}-6, 1{\rm e}-5, 3{\rm e}-5, 1{\rm e}-4\}$. We used one NVIDIA RTX A5000 to run our experiments.

\section{Additional Experiments}

\subsection{Spurious correlations learned with augmentation}\label{Aug:table}
As we explained the experiment for \Cref{APEshifted_5to20}, despite that the model does well in single-digit sums, it cannot maintain this performance on the first digit of multi-digit sums. The accuracy on every position of the output is listed in \Cref{Aug:table}. We conclude that augmentation based on the translational symmetry leads to spurious correlations between positions, which is in agreement with our findings in theory described in \Cref{label:theory}.

\begin{table}[h!]
     % \vskip 0.15in 
    \centering
    
    \begin{tabular}{lcccr}
    \toprule
    Test Case & Accuracy \\
    \midrule
    Shifted integers & 99.7 $\pm$ 0.1\\
    Uniform distribution & 81 $\pm$ 1\\
    Accuracy on the 1st digit & 90 $\pm$ 1 \\
    Accuracy on the 2nd digit & 96 $\pm$ 1\\
    Accuracy on the 3th digit & 98.8 $\pm$ 0.1\\
    Accuracy on the 4th digit & 98.0 $\pm$ 0.1\\
    Accuracy on the 5th digit & 98.8 $\pm$ 0.1\\
    Accuracy on 6:11 digits & 96 $\pm$ 1\\
    \bottomrule
    \end{tabular}
    \label{tab1} 
    \caption{10-digit sum's accuracy of the augmented model for in-distribution, out of distribution (uniform integers in $[1, 10^{10} -1]$), and isolating the performance at every coordinate of the output. Despite the acceptable accuracy of the model on most digits, the first digit errors hurt the model.} 
\end{table}

\subsection{Addition with no carries}
\begin{figure}
    \centering
    \includegraphics[width=0.45\linewidth]{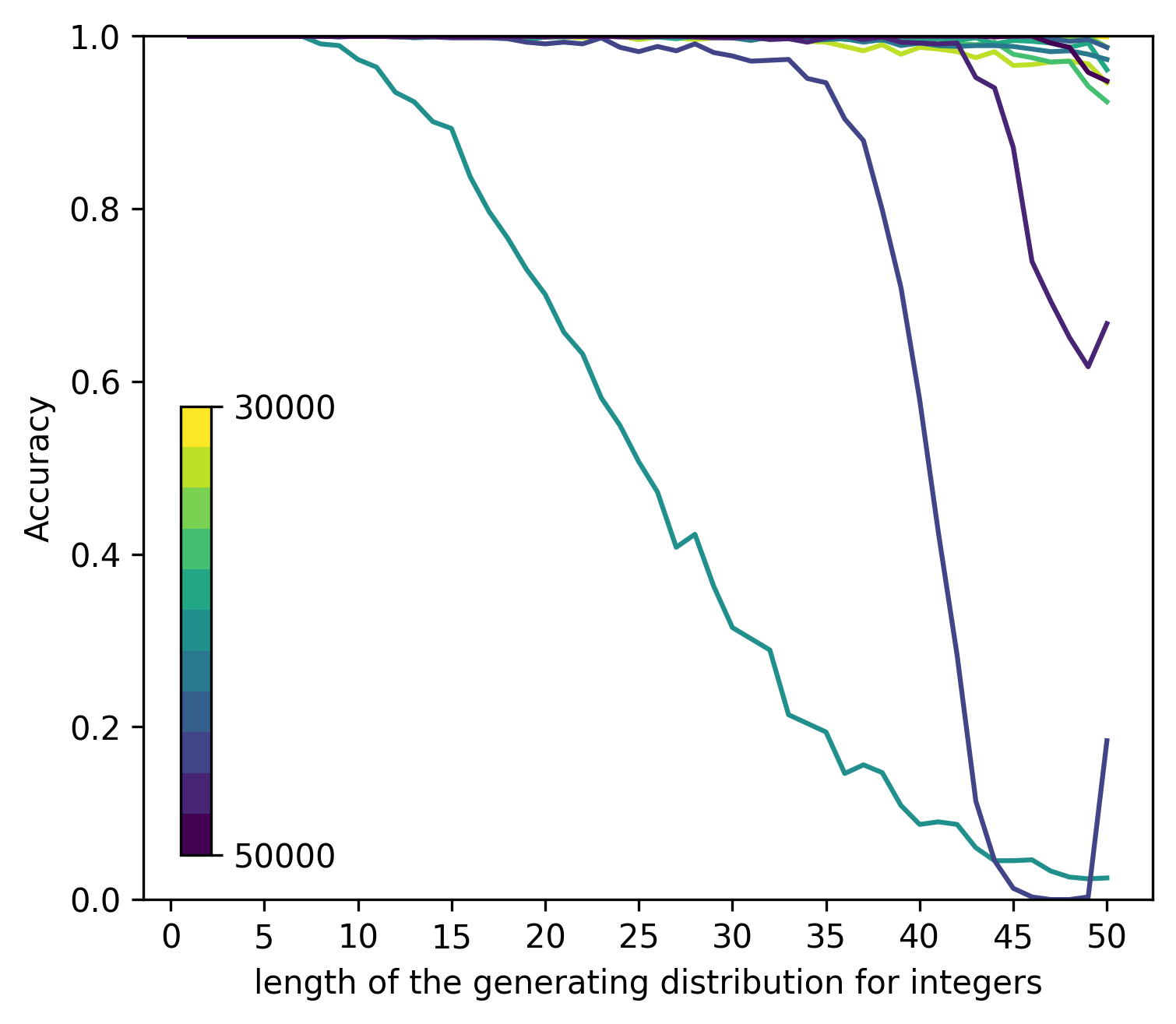}
    \caption{A relative model easily learns addition where there is no carry. Here, training samples are up to 5-digit sums.}
    \label{nocarry}
\end{figure}

One of the experiments we did to establish that the main issue in the addition task is the existence of carries, was to try a setting with no carry, every step's output being independent of former calculations. In this case, addition is just a digit-wise operation that can be implemented in parallel for a sequence. Therefore, with our input's format and relative positional encoding, the model at each position has to only look at itself and $l+1$ (Addition in parallel Section), and calculate the sum of those tokens. To this end, Figure \ref{nocarry} approves that this problem does not cause difficulty unless carries are included.

\begin{figure}
     \centering
     \begin{subfigure}[b]{\textwidth}
         \centering
         % Linewidth is the now the unit for half a page.
         \includegraphics[width=1.0\linewidth]{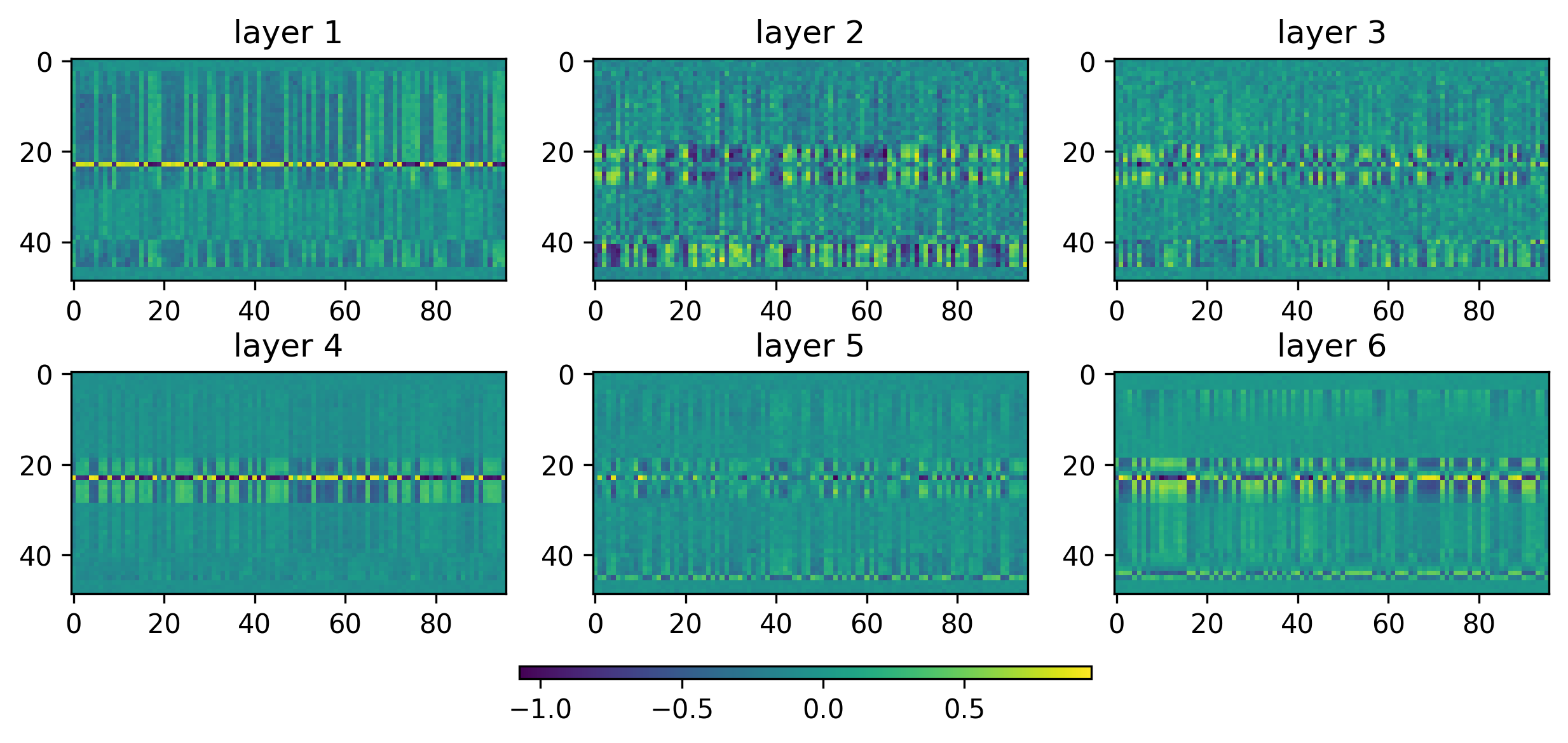}
         \caption{}
         \label{RPE_maps}
     \end{subfigure}
    \\
     \begin{subfigure}[b]{\textwidth}
         \centering
         % Linewidth is the now the unit for half a page.
         \includegraphics[width=1.0\linewidth]{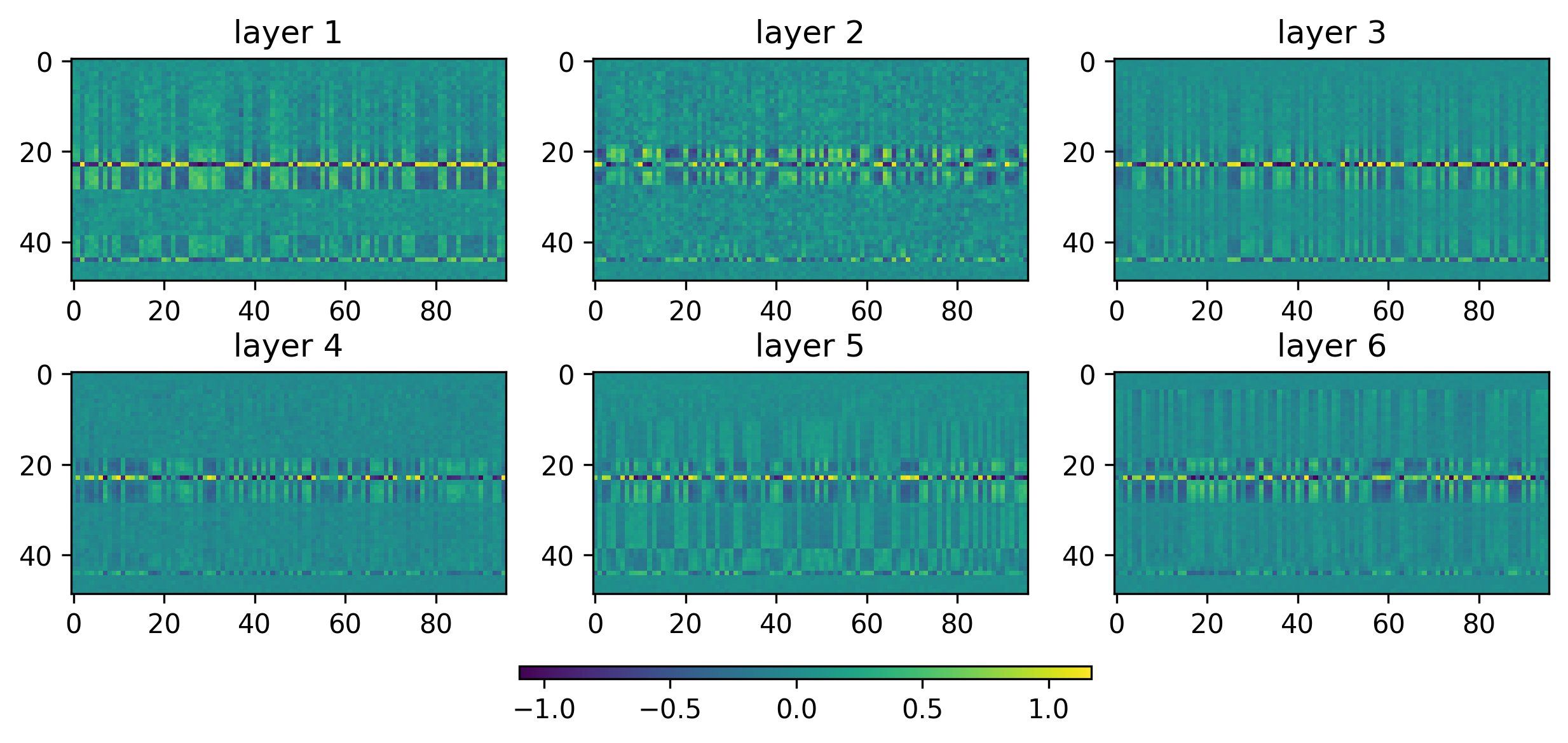}
         \caption{}
         \label{CPE_maps}
     \end{subfigure}
    \caption{\textbf{(a)} Plotting the pairwise positional vectors for each layer of the model with RPE in the single-digit multiplier scenario. The main signal is in the 23rd vector, meaning that the model mainly looks at only one position back at each layer. \textbf{(b)} The model with UPE in the same setting as (a) exhibits more interpretable positional maps.}
\end{figure}

\begin{figure}
     \centering
     % You have to set the textwidths to sum to under 1 because Latex doesn't like it when the two are exactly 1. :(
     % This splits the figure into it to two half pages width Figures.
     \begin{subfigure}[b]{0.8\textwidth}
         \centering
         % Linewidth is the now the unit for half a page.
         \includegraphics[width=1.0\linewidth]{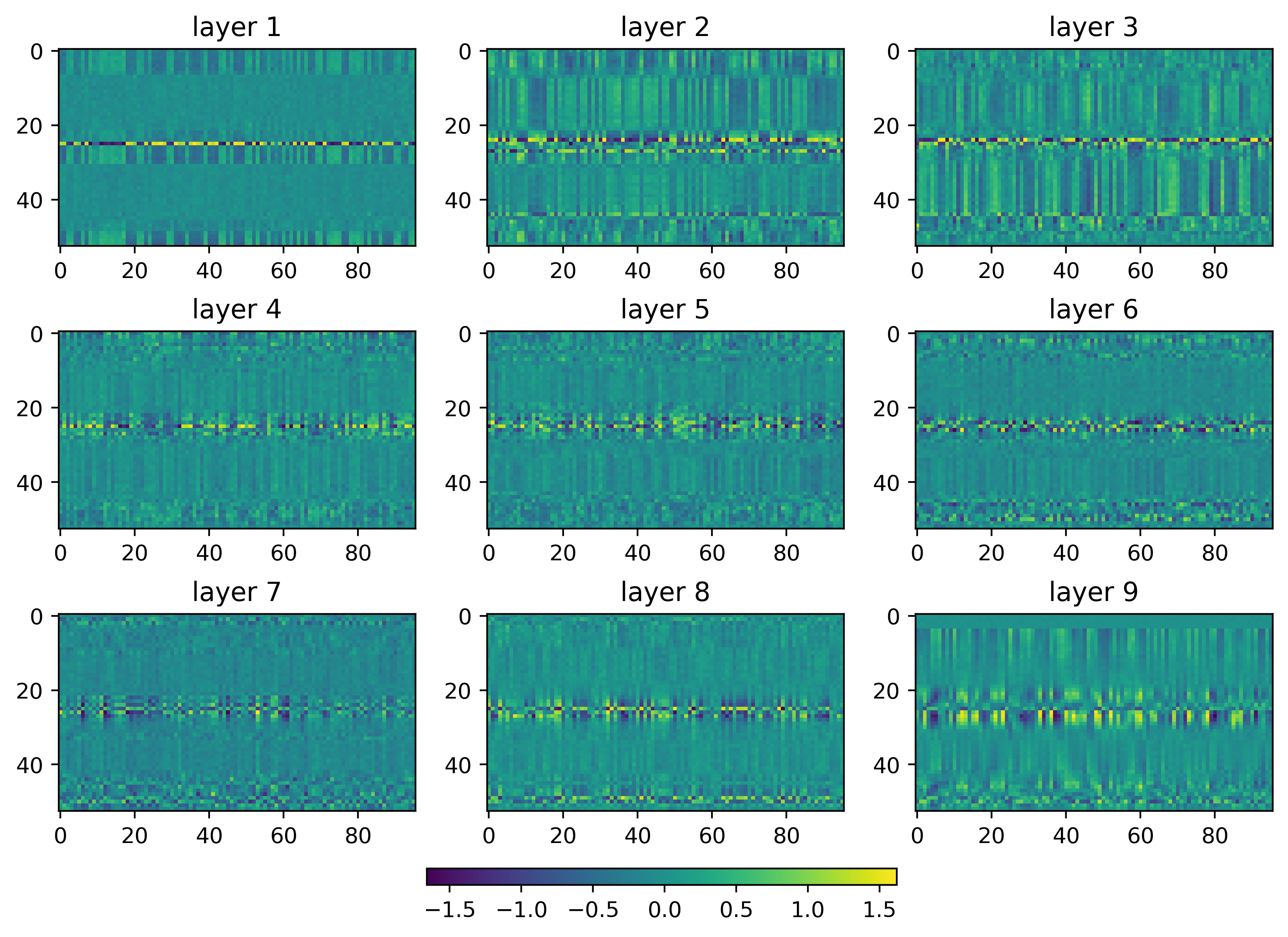}
         \caption{}
         \label{RPE3_maps}
     \end{subfigure}
    \\

     \begin{subfigure}[b]{0.8\textwidth}
         \centering
         % Linewidth is the now the unit for half a page.
         \includegraphics[width=1.0\linewidth]{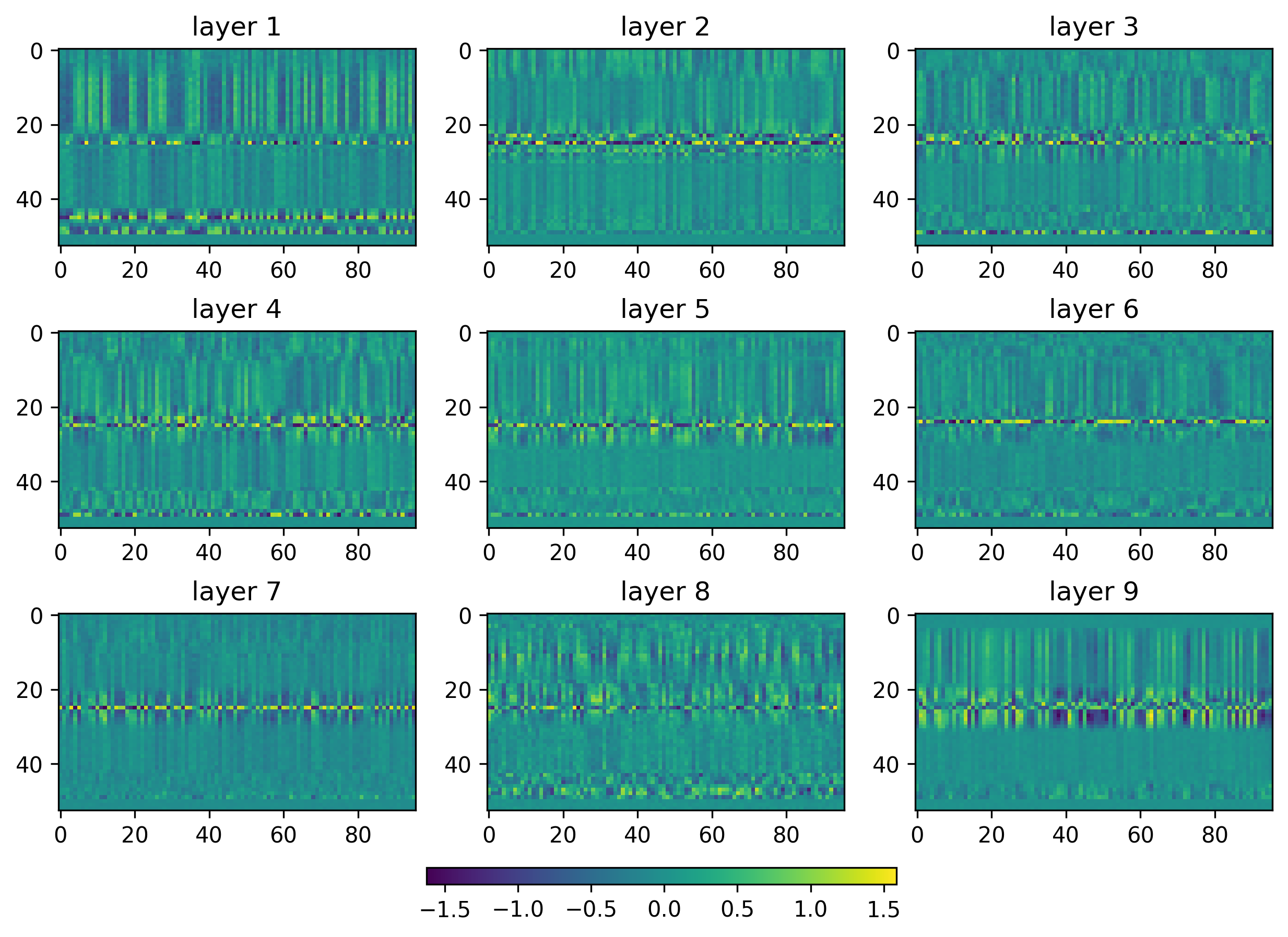}
         \caption{}
         \label{CPE3_maps}
     \end{subfigure}
     \vspace{-0.05in}
    \caption{\textbf{(a)} Plotting the pairwise positional vectors for each layer of the model with RPE in the 3-digit multiplier scenario. The model has to look at more positions in addition to the previous one. \textbf{(b)} The model with UPE in the same setting as (a) is still more interpretable.}
\end{figure}
\subsection{Positional maps}
Using weight-decay in training causes unnecessary parts of the network to vanish over time. Thus, plotting the positional vectors is informative about what positions the model is looking at. Figures \ref{RPE_maps} and \ref{CPE_maps} are the positional plots for models with RPE and UPE trained in the single-digit multiplier scenario (accordingly for models in \ref{Mult_RPE} and \ref{Mult_UPE_set}). y-axis accounts for the index of the plotted vectors: $i - j + (\text{max-length} -1)$ for $P_{i - j}$ so that in all the plots of \ref{RPE_maps} and \ref{CPE_maps} the 24th is $P_0$. And x represents elements of a positional vector here. Because the attention dimension is 768 in our setting and the number of heads is 8, positional vectors are 86-dimensional. We want to highlight one similarity and one major difference from the plots. In both models and in every layer, the 23rd vector is the most noticeable, which means looking at the previous position. Considering our task, this seems very intuitive, as carry comes only from the previous position in the single-digit multiplier case. The difference is that attention is less dispersed in UPE, especially in layers 2 to 6, which will avoid overfitting to in-distribution data. We think this is one of the reasons for a better out-of-distribution performance of the model with UPE. 

We plot the positional vectors for the 3-digit multiplier setting in Figures \ref{RPE3_maps} and \ref{CPE3_maps}. This time max-length = 27, and therefore, $P_0$ is the 26th row of every plot.  This time besides $P_{-1}$, $P_{-2}$ and $P_{-3}$ are also noticeable (this can be seen at layer 2 of \ref{Mult3_UPE}), which again is reasonable since carries jump to at most 3 latter positions for a 3-digit multiplier. Although less conspicuous here, the model with UPE still depicts less dispersed attention than the RPE.

\subsection{Distribution of the Complexity}\label{carry_dist}
% \textbf{Complexity of 50-digit addition}
% As mentioned in Section \ref{complexity}, longer dependencies, which correspond to more complex samples, can only occur when there exists a series of consecutive $j < i$ such that $x^1_j + x^2_j = 9$, and a carry coming before them. In Figure \ref{sum_carries_50}, after length of 4, the likelihood of cascading carries becomes extremely low. In fact, the probability of cascading carries up to length 4 covers 0.998 of the total probability. Now,  Figure \ref{sum_carries_5} shows the distribution of the cascading length for 5-digit sums. The probability of 4 is 8.3e-4 in this plot. By ensuring a good coverage, that is, using a high number of training samples from $\D_s$, samples with complexity of 4 will be covered and observed by the model, enabling the model to generalize well. It is crucial to have sufficient coverage over the most complex samples in $\D_s$ in order to achieve generalization across $\D_u$.

% \begin{figure}[h!]
%      \centering
%      % Linewidth is the now the unit for half a page.
%      \includegraphics[width=0.4\textwidth]{figs/Sum_carries_5.png}
%      \caption{Distribution of cascading carries in 5-digit sums.}
%      \label{sum_carries_5}
% \end{figure}

\begin{figure}[t!]
    \centering
    \begin{subfigure}[b]{0.4\textwidth}
         \centering
         \includegraphics[width=1.0\linewidth]{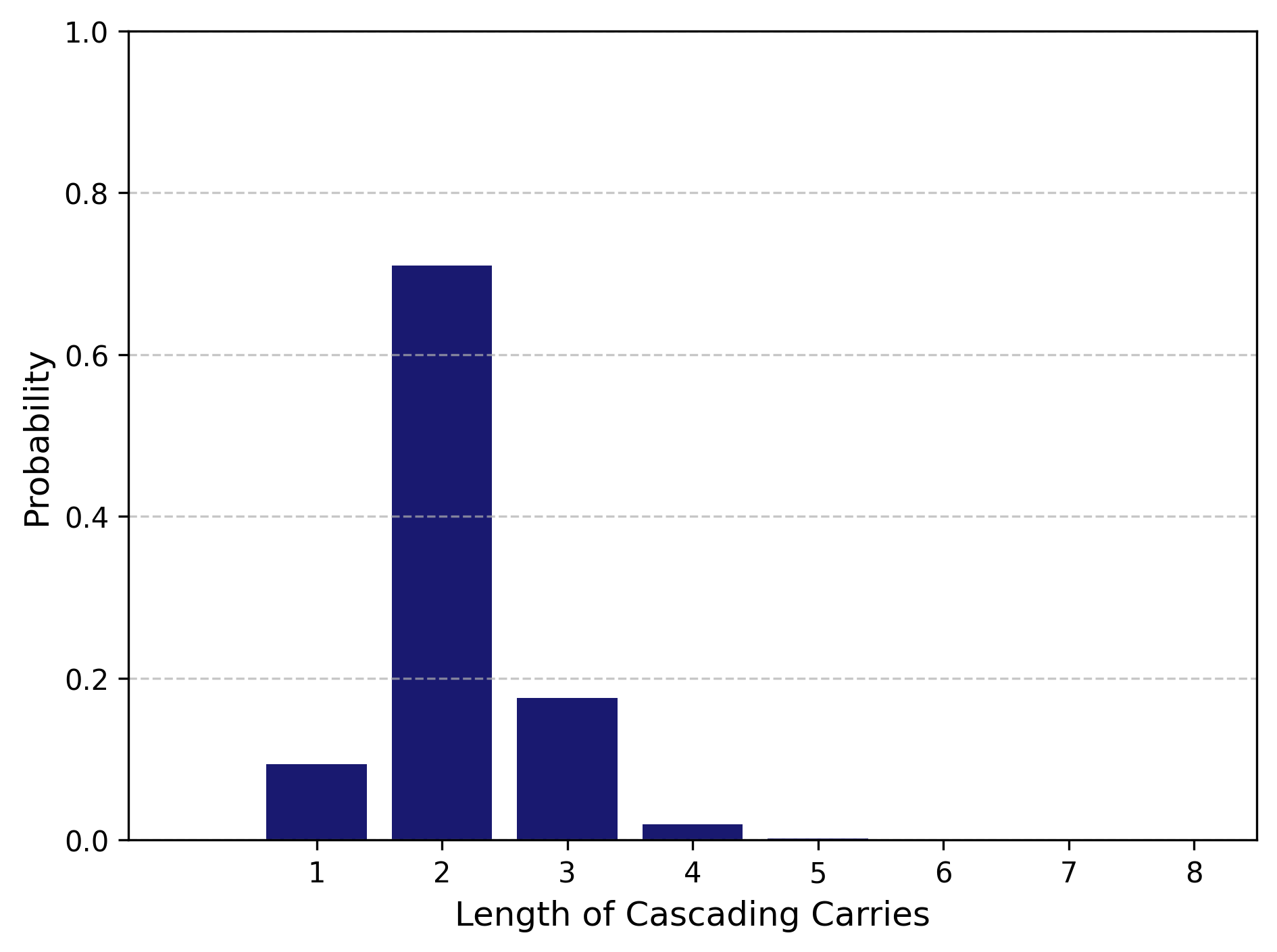}
         \caption{}
         \label{sum_carries_50}
    \end{subfigure}
    \begin{subfigure}[b]{0.4\textwidth}
        \centering
        \includegraphics[width=1.0\textwidth]{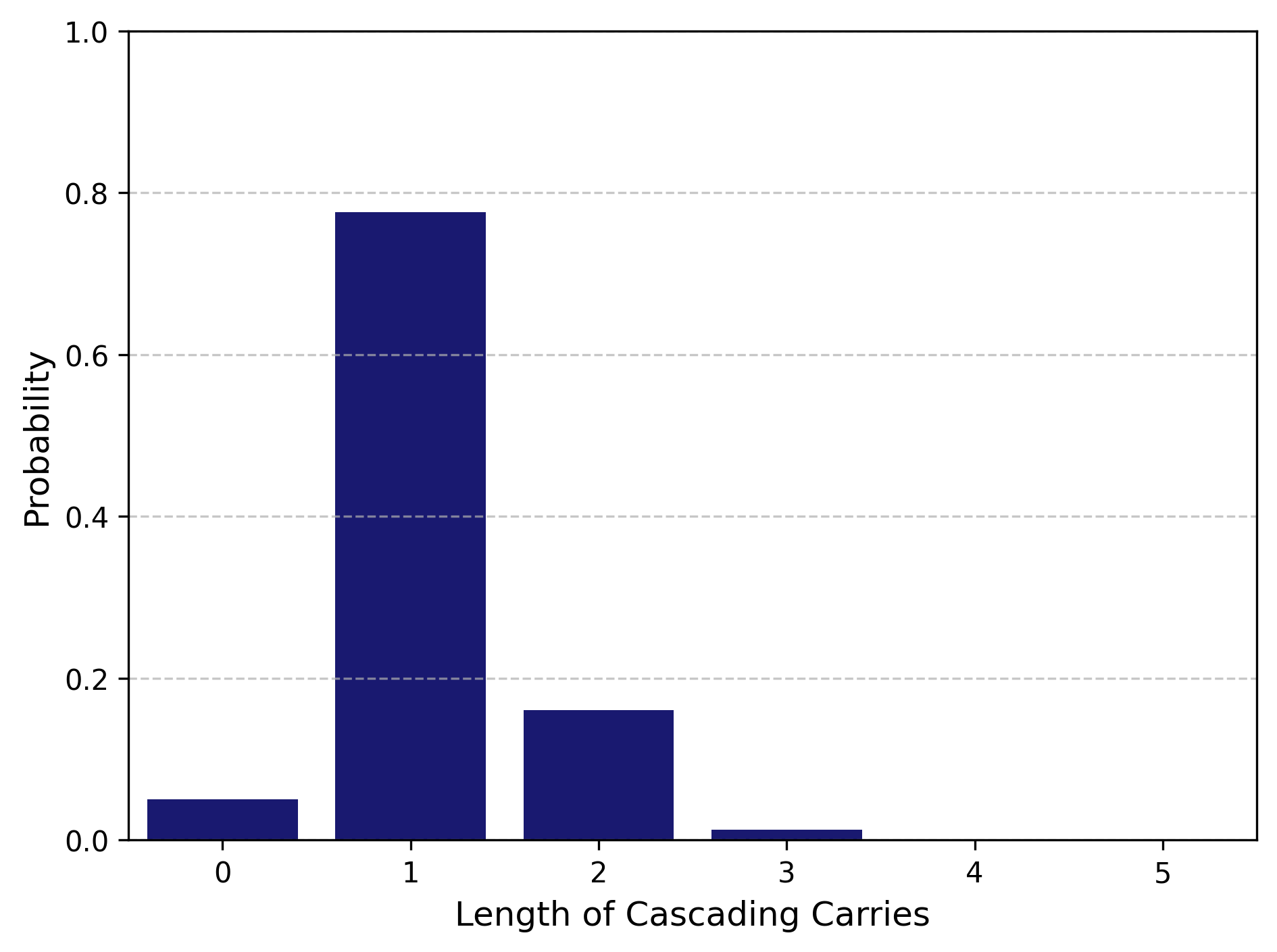}
        \caption{}
        \label{sum_carries_5}
    \end{subfigure}
     %\hfill % Remove this comment if you want the figures to be pushed outward to the edges!
    \caption{\textbf{(a)} Distribution of cascading carries for 50-digit sums. Most of the mass lies within 1 to 4. \textbf{(b)} Distribution of cascading carries in 5-digit sums.  Increasing the coverage of high-complexity samples of 5-digit sums can help offset the disparities in distributions.}
\end{figure}

\noindent\textbf{Distribution of cascade carries.} \Cref{sum_carries_50,sum_carries_5} show the distribution of the cascade length for randomly sampled 50-digit and 5-digit sums. As can be seen, the cascade length probability is very small beyond length 4. In fact, the probability of cascading carries up to length 4 covers 0.998 of the total probability. Training on 5-digit addition can provide us with cascades of length up to 5, however, extensive training samples are required for the model to cover and observe high-complexity samples (the probability of length 4 cascades in random 5 digit numbers is roughly 8.3e-4). This explains why the performance drops with higher cascade length. Note that despite never seeing cascades of length higher than 5, the model still gets non-trivial performance on samples with cascade length more than 5.

\begin{figure}[t!]
     \centering
     \begin{subfigure}[b]{0.4\textwidth}
         \centering
         \includegraphics[width=1.0\linewidth]{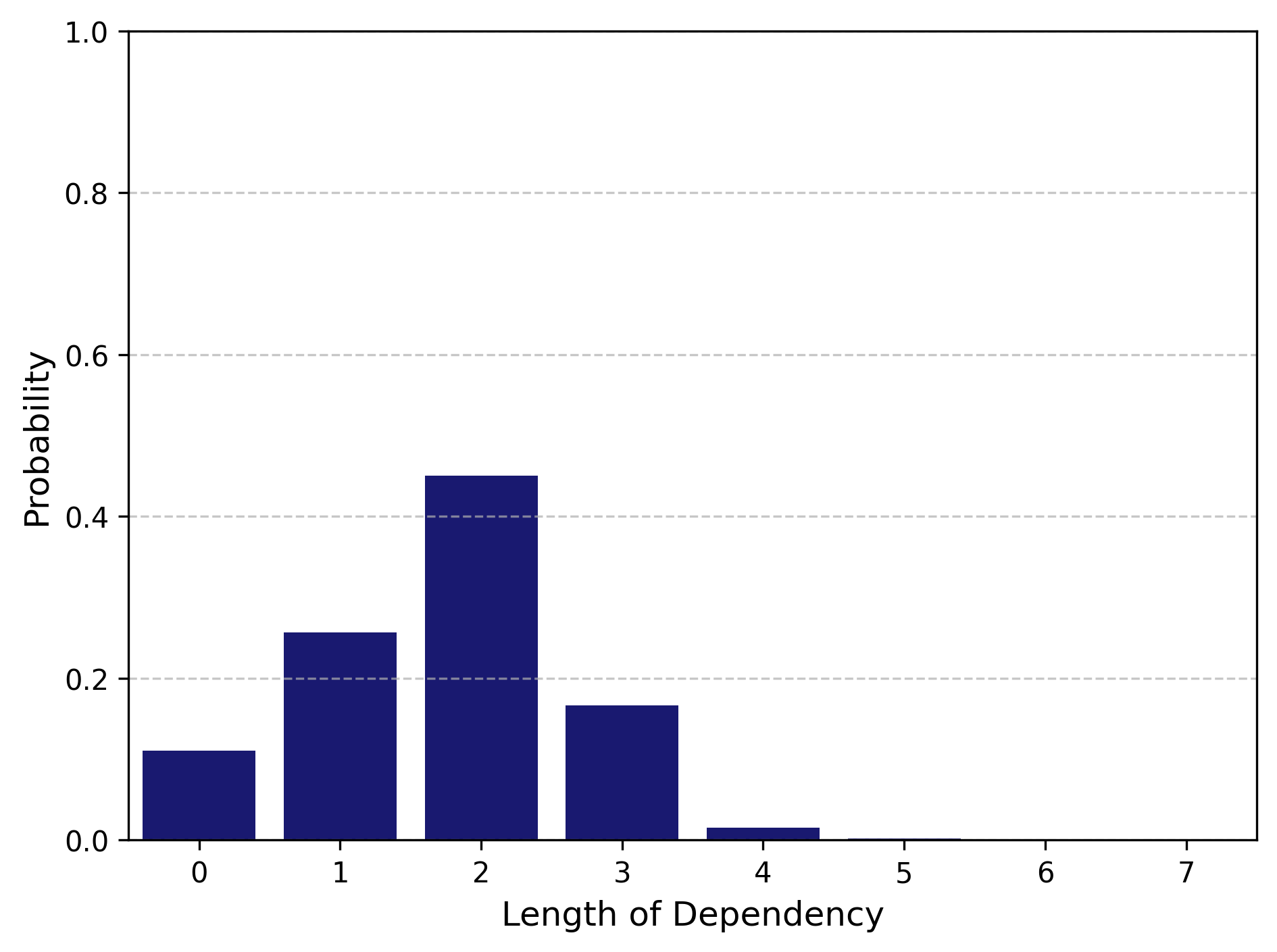}
         \caption{}
         \label{smult_20}
     \end{subfigure}
    \centering
    \begin{subfigure}[b]{0.4\textwidth}
         \centering
         \includegraphics[width=1.0\linewidth]{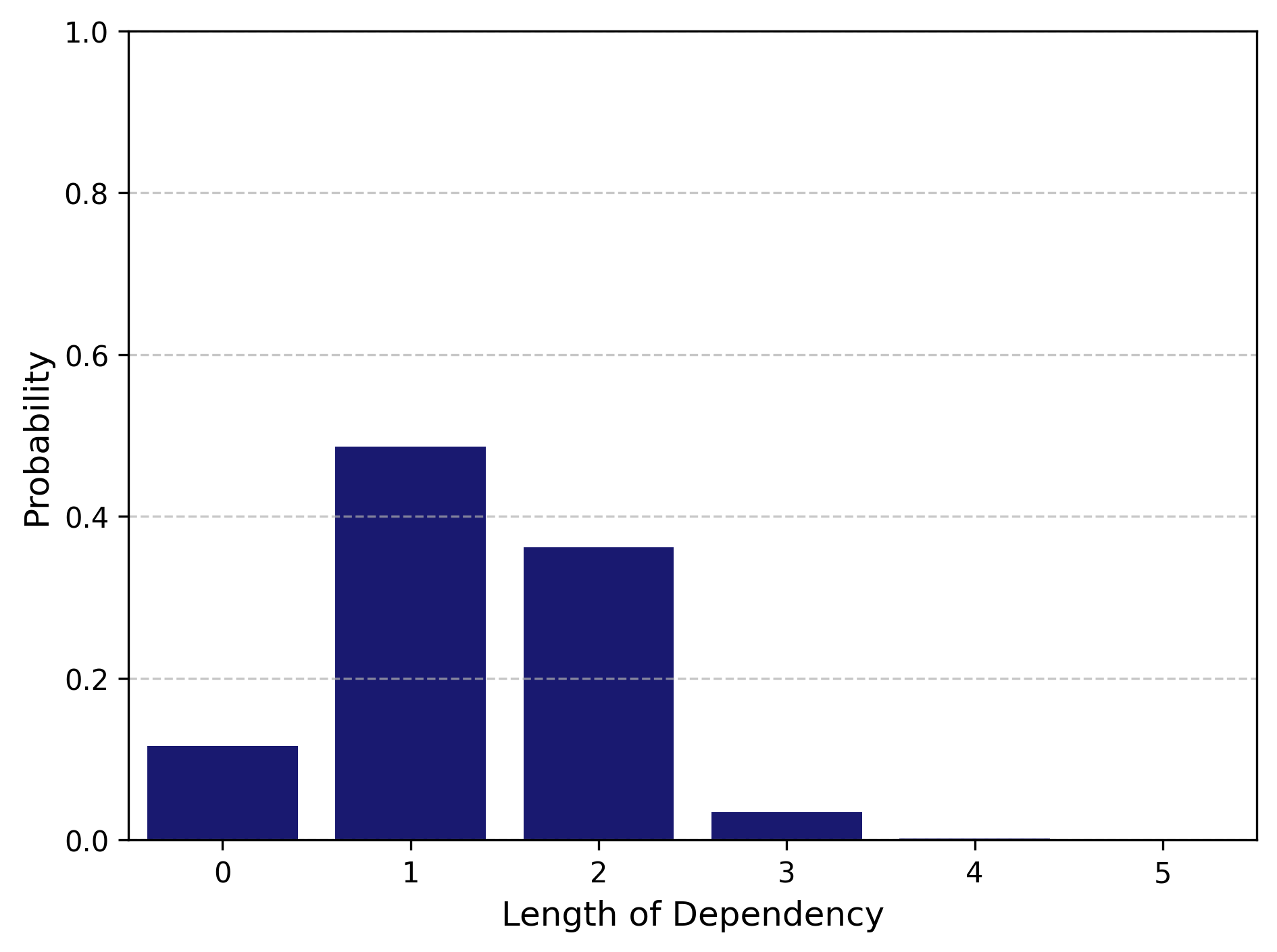}
         \caption{}
         \label{smult_5}
     \end{subfigure}
     \caption{\textbf{(a)} Distribution of dependency for 20-digit multiplicand. \textbf{(b)} Distribution of dependency for the 5-digit multiplicand scenario.}
\end{figure}

\paragraph{Complexity of the single-digit multiplier scenario} Unlike addition, there is no direct way to identify samples with higher complexities in multiplication since the carry's value is not limited to 1. However, we can use the fact that length of the maximum dependency in a sequence is equal to the total number of levels in Figure \ref{mul_example}. By the last level, every position's value will be a single digit. Figures \ref{smult_20} and \ref{smult_5} show the dependency distribution for the 20-digit multiplicand and the 5-digit multiplicand accordingly, when the multiplier is uniformly sampled from 1 to 9. Similarly to addition, up to length of 4 covers 99.8 percent of the distribution in Figure \ref{smult_20}, and the probability of 4 is 1.9e-3 in Figure \ref{smult_5}. Hence, a sufficient number of training samples from $\D_s$ will ensure generalization on the unseen part.

\section{Proofs} \label{app:proofs}

\subsection{Proof of \Cref{alltheory}: APE}\label{APE_proof}
After absorbing $W$ into the positional vectors, i.e.  $p_i \leftarrow  W p_i$, and letting $v:= W_V^T h$, we obtain:

\begin{equation}\label{linear_ape}
    \hat{y}_i = \sum_{r=1}^n p_i^T p_r \langle v, x_r \rangle
\end{equation} 

During training only the first $n_1$ positions are variable. Thus, the total loss function at every step is: 
\begin{equation}\label{eq:loss_noaug}
    \ell(\{p_1, \cdots, p_n\}) = \sum_{i =1}^{n_1} (\hat{y}_i - y_i)^2
\end{equation}

Before diving into the proofs, it is important to note that we write the equations on a ring, where index "-1" points at the last element in the sequence; i.e. index $n$. This makes it simpler to handle the boundary conditions.

To make the convergence problem easier, we set $v = \theta$ at initialization such that it does not need to explore any other direction. This alignment assumption only holds for our proofs of negation for absolute positional encoding (APE) and APE with augmentation, and the extra assumption makes our proof of negation even stronger. We will prove with all of these simplifications that APE will not still learn the structure for padded positions, hence failing to generalize on the unseen. 

\begin{unumproposition}[Formal statement of \Cref{alltheory}-- First part]
The solution found by the gradient descent with infinitesimal weight-decay in the population regime on the loss function \eqref{eq:loss_noaug} satisfies $p_k = 0$ for all $k \not \in [n_1]$. Therefore, the expected loss at each position, i.e., $\E \{\ell_i\} = \E \{(\hat{y}_i - y_i)^2 \}$ for $i>n_1$ converges to $\alpha^2 + 2 \beta^2$.
\end{unumproposition}

\begin{proof}
\begin{align*}
    \ell(P) :&= \ell(\{p_1, \cdots, p_n\}) \\
            & =  \sum_{i =1}^{n_1} (\hat{y} - y)^2 \\
            &= \sum_{i =1}^{n_1} \big(\sum_{r=1}^{n_1} p_i^T p_r \langle \theta, x_r \rangle + \sum_{r=n_1 + 1}^n p_i^T p_r \langle \theta, \bzero\rangle - \alpha \langle \theta, x_{i} \rangle -\beta \langle \theta, x_{i-1} \rangle -\beta \langle \theta, x_{i+1} \rangle \big)^2 \\
            &= \sum_{i =1}^{n_1} \big(\sum_{r=1}^{n_1} p_i^T p_r \langle \theta, x_r \rangle - \alpha \langle \theta, x_i \rangle -\beta \langle \theta, x_{i-1} \rangle -\beta \langle \theta, x_{i+1} \rangle \big)^2 
\end{align*}

In the second equality we have separated the contribution of padded elements from the rest. Because $x_r \sim \calN(0, I_d)$ and $\| \theta \| = 1$ their inner product is a normal random variable that we call it $s_r$ for $r \leq n_1$. Taking the gradient of the loss function with respect to an arbitrary vector $p_k$, we will have:
\begin{align*}
    \nabla_{p_k} = & \sum_{i =1}^{n_1} 2 \big(\sum_{r=1}^{n_1} p_i^T p_r s_r  - \alpha s_i - \beta s_{i-1} \mathbbm{1}(i \neq 1) - \beta s_{i+1} \mathbbm{1}(i \neq n_1) \big) \\
    & \quad \times \Big(\mathbbm{1}(k \in [n1])\big(\mathbbm{1}(k = i)(2 p_k s_k + \sum_{r \neq k}^{n_1} p_r s_r) + \mathbbm{1}(k \neq i) (p_i s_k) \big)+  \mathbbm{1}(k \notin [n_1]) \big(0) \Big)\\
\end{align*}

% Now we want to compute the gradient flow of every vector, meaning that we will work with expectation of the above expression. So for $k \in [n_1]$:

% \begin{align*}
%     \E \{ \nabla p_k\} = & 2 \E \Big\{ \big(\sum_{r=1}^{n_1} p_k^T p_r s_r - s_k\big) \big(2 p_k s_k + \sum_{r \neq k}^{n_1} p_r s_r)\Big\} + 2 \sum_{i \neq k}^{n_1} (p_i^T p_k) p_i +  2 \sum_{i= n_1+1}^{n} (p_i^T p_k) p_i \\
%     = & 4 \sum_{i = 1}^ {n_1} (p_i^T p_k) p_i - 4 p_k +  2 \sum_{i= n_1+1}^{n} (p_i^T p_k) p_i
% \end{align*}
In which $\mathbbm{1}$ function gives 1 if inside condition holds, 0 otherwise. Therefore the gradient for $k \notin [n_1]$ is zero. As a result, the updating rule under SGD with weight-decay rate equal to $\epsilon$ for vectors outside the window is:
$$p_k^{(t+1)} = p_k^{(t)} - \epsilon p_k^{(t)}$$

And for $\epsilon$ being sufficiently small, we will have: 
$$p_k \to_t \bzero \quad \forall k \notin [n_1]$$

Meanwhile generalization over elements outside of $[n_1]$ requires $p_k^T p_k \to \alpha$ simply by matching the form of $\hat{y}$ to $y$. Thus, APE cannot learn the structure outside of the seen window. 
\end{proof}

\subsection{Proof of \Cref{alltheory}: APE + Augmentation}\label{Aug_proof}
Consider the same task \eqref{regression_y_x} and the model \eqref{factored}. However, we consider training with data augmentation in which we apply translational shifts to the training data. Before, the training data had Gaussian-generated values in the first $n_1$, and the other positions are zero-padded. Here, for each $i \in [n]$ we shift the training data by $i$ positions. The shifted sequences will have Gaussian values at positions $i+1, i+2, \cdots, i+n_1$ and are zero-padded elsewhere (note that we're considering circular shifts). The shifted data will then be augmented to the training set. The augmented data is thus generated according to the following distribution: For each $i \in [n]$: 
$$
\begin{cases}
    x_r \stackrel{\rm{i.i.d.}}{\sim} \calN(0, \mathbb{I}_d) \quad &\text{for} \quad i+1 \leq r \leq i+n_1, \\
    x_r = \bzero_d & r \notin \{i+1, \cdots, i+n_1\}. 
\end{cases} 
$$

The overall loss function for the case where data is shifted by $i$ positions is: 
\begin{equation}\label{eq:loss_aug_i}
    \ell_i(\{p_1, \cdots, p_n\}) = \sum_{t = 1}^{n_1} (\hat{y}_{i+t} - y_{i+t})^2
\end{equation}

By summing over all the shifts, the total augmented loss becomes:

\begin{equation}\label{eq:loss_aug}
    \ell(P) := \ell(\{p_1, \cdots, p_n\}) =  \sum_{i =1}^{n} \sum_{t=1}^{n_1} (\hat{y}_{i+t} - y_{i+t})^2
\end{equation} 

\begin{unumproposition}[Formal statement of \Cref{alltheory}-- Second part]
The solution found by the gradient flow with infinitesimal weight-decay in the population regime on the loss function \eqref{eq:loss_aug} satisfies $ \E \{\ell_i\} = \E \{(\hat{y}_i - y_i)^2 \} =  \Omega(n/d)$ when $x_i \stackrel{\rm{i.i.d.}}{\sim} \calN(0, \mathbb{I}_d) \ \forall i \in [n]$.
\end{unumproposition}

\begin{proof}
Inserting $y$ and $\hat{y}$ into this will give: 

\begin{align*}
    \ell(P) &=\sum_{i =1}^{n}  \sum_{t = -w}^{w} \big(\sum_{r \in W_i} p_{i+t}^T p_r \langle \theta, x_r \rangle - \alpha \langle \theta, x_{i+t} \rangle -\beta \langle \theta, x_{i+t-1} \rangle \mathbbm{1}(t \neq -w) \\
    & \hspace{2.6in} - \beta \langle \theta, x_{i+t+1} \rangle \mathbbm{1} (t \neq w) \big)^2
\end{align*}

Borrowing our definition from last part $s_i = \langle \theta, x_i \rangle$ for those positions that are not padded, we compute the expected gradients:

\begin{align*}
    \E\{\nabla_{p_k}\}= & \E \Big\{ \sum_{i =1}^{n}  \sum_{t = -w}^{w} 2 \Big(\sum_{r \in W_i} p_{i+t}^T p_r s_r - \alpha s_{i+t} - \beta s_{i+t-1} \mathbbm{1}(t \neq -w) - \beta s_{i+t+1} \mathbbm{1}(t \neq w)  \Big)  \\
    & \qquad \qquad \quad \ \times \Big(\mathbbm{1}(k \in W_i)\big(\mathbbm{1}(k = i+t)(2 p_k s_k + \sum_{r \in W_i, r \neq k} p_r s_r ) \\
    & \hspace{0.95 in} + \mathbbm{1}(k \neq i+t) (p_{i+t} s_k) \big) \Big) \Big\} \\
    = & 2 \sum_{i =1}^{n}  \sum_{t = -w}^{w} \Big\{ \mathbbm{1}(k \in W_i) \big\{ \mathbbm{1}(k = i+t) \big( 2 (p_k^T p_k) p_k - 2 \alpha p_k + \sum_{r \in W_i, r \neq k} (p_k^T p_r) p_r \\
        & \qquad \qquad \qquad \qquad \qquad \qquad \qquad \qquad - \beta p_{k+1} \mathbbm{1}(t \neq w)
          - \beta p_{k-1} \mathbbm{1}(t \neq -w) \big)  \\
          & \hspace{1.3in} + \mathbbm{1}(k \neq i+t) \big((p_{i+t}^T p_k)p_{i+t} - \beta \mathbbm{1}(k = i+t -1) \mathbbm{1}(t\neq w) p_{k+1} \\
          & \hspace{2.99in} - \beta \mathbbm{1}(k = i+t+1) \mathbbm{1}(t \neq -w) p_{k-1} \big) \big\} \Big\} \\
    = &  2 \sum_{i =1}^{n} \Big\{ \mathbbm{1}(k \in W_i) \big\{2 \sum_{r \in W_i} (p_k^T p_r) p_r - 2 \alpha p_k  -  \mathbbm{1}(k \neq i+w) 2 \beta p_{k+1} \\
    & \hspace{2.47in} - \mathbbm{1}(k \neq i-w) 2 \beta p_{k-1} \big\} \Big\}
\end{align*}

In the last line, we have taken the some over $t$ inside, where for each $k \in W_i$ there is only one $t$ that $k = i+t$, and at most one such that $k = i+t-1$, or $k = i+t+1$. To apply the sum over $i$ on the first component of the sum, we have to count the total number of times that $k$ and $r$ fall on the window when varying $i$. The answer is $(2w+1 - |i - r|)$ for $|i - r| \leq 2w+1$ and zero otherwise. For other terms, every k happens $(2w + 1)$ times, therefore: 

$$\E\{\nabla_{p_k}\} = 4 \sum_{r = -2w}^{2w} (2w+1 -|r|) (p_k^T p_{k+r}) p_{k+r} - 4 (2w +1) \alpha p_k  -  8w \beta p_{k+1} - 8w \beta p_{k-1} $$

We insert this gradient in the gradient flow equation with weight-decay: 
\begin{align}
    \frac{d}{dt}p_k =  -  \nabla_{p_k} - \epsilon p_k =& - 4   \sum_{r = -2w}^{2w} (2w+1 -|r|) (p_k^T p_{k+r}) p_{k+r} + (4   (2w +1) \alpha - \epsilon) p_k   \nonumber\\
    & +  8   w \beta p_{k+1} + 8   w \beta p_{k-1}
\end{align}\label{p_derivative}
Since $\epsilon$ is small and it appears in a component with non-zero factor, we can neglect it. 

It is not where each positional vector converges to that controls the generalization, but rather their inner products that show up in \eqref{linear_ape}. Let's define $A_{k,l}:= p_k^T p_l$, we compute how elements of $A$ evolve over time: 
\begin{align}\label{A_derivative}
    \frac{d}{dt} A_{k, l} = & (\frac{d}{dt}p_k)^T p_l + p_k (\frac{d}{dt} p_l) = -  \{ (\nabla_{p_k})^T p_l + p_k^T \nabla_{p_l}\} \nonumber\\
    = & -4   \sum_{r = -2w}^{2w} (2w+1 -|r|) \big((p_k^T p_{k+r}) (p_l^T p_{k+r}) + (p_l^T p_{l+r}) (p_k^T p_{l+r}) \big) +  8   (2w +1) \alpha p_k^T p_l \nonumber \\
    & + 8   w \beta (p_l^T p_{k-1} + p_l^T p_{k+1} + p_k^T p_{l-1} + p_k^T p_{l+1}) \nonumber \\
    = & 4   \sum_{r = -2w}^{2w} (2w+1 -|r|) \big(A_{k, k+r} A_{l, k+r} + A_{l, l+r} A_{k, l+r}) +  8   (2w +1) \alpha A_{k, l} \nonumber \\
    & + 8   w \beta (A_{l, k-1} + A_{l, k+1} + A_{k, l-1} + A_{k, l+1}) 
\end{align}

This means that elements of $A$ determine their own dynamics, and it suffices to track them only. Next step is to put assumptions on the initialization of positional vectors. Final model must acquire the same translational symmetries as the underlying task for it to generalize at test time. We seed these conditions at initialization to put it on the right track, and we will show that even in this case off diagonal elements of A do not converge to desired values. Applying the translational invariance to initial inner products of $p_k$'s:
\begin{equation}\label{initialization}
    p_k^{(0)T} p_{k+j}^{(0)} = p_l^{(0)T} p_{l+j}^{(0)} \quad \forall k, l, j
\end{equation}

\begin{lemma}
    Condition \eqref{initialization} holds for all $t$ if positional vectors start under condition \eqref{initialization} and evolve by \eqref{p_derivative}.
\end{lemma}
\begin{proof}
    In other words the lemma says $A_{k, k+j} = A_{l, l+j}$. By induction, it only suffices to show if condition holds at one time t, their derivatives are equal. Using equation \eqref{A_derivative}:
    \begin{align*}
        \frac{d}{dt} A_{k, k+j} = & - 4   \sum_{r = -2w}^{2w} (2w+1 -|r|) \big(A_{k, k+r} A_{k+j, k+r} + A_{k+j, k+j+r} A_{k, k+j+r})\nonumber \\
        &  +  8   (2w +1) \alpha A_{k, k+j}  + 8   w \beta (A_{k+j, k-1} + A_{k+j, k+1} + A_{k, k+j-1} + A_{k, k+ j+1}) 
    \end{align*}
    In every term of above expression $k$ appears on both indices of A, hence it can be replaced by l everywhere under condition \eqref{initialization}. Thus:
    $$\frac{d}{dt} A_{k, k+j} = \frac{d}{dt} A_{l, l+j}$$
\end{proof}

As a result, we can address the elements of A by the difference of its two indices: 
$$A_j \leftarrow A_{l, l+j}$$

$A_0$ stands for all diagonal elements of $A$, and $A_j$ for $j \geq 1$ addresses off diagonal elements with respect to their distance from the main diagonal. Note that due to the circular constraint, the number of independent variables is equal to $\lfloor \frac{n}{2} \rfloor$, and $A_{\lfloor \frac{n}{2} \rfloor + 1} = A_{\lfloor \frac{n-1}{2} \rfloor}$. This brings us to a system of coupled differential equations for elements of A: 

\begin{align*}
    \frac{d}{dt} A_j = &  -4   \sum_{r = -2w}^{2w} (2w+1 -|r|) A_r \big(A_{r -j} +  A_{j+r}\big) +  8   (2w +1) \alpha A_j \\
    & + 8   w \beta (A_{-j-1} + A_{1-j} + A_{j+ 1} + A_{j - 1})
\end{align*}

Using the fact that $A_j = A_{-j}$:
\begin{align}\label{diff}
    \frac{d}{dt} A_j =  8   (2w+1) (\alpha - A_0) A_j - 8   \sum_{r = 1}^{2w} (2w+1 -|r|) A_r \big(A_{j-r} +  A_{j+r}\big) + 16   w \beta (A_{j+ 1} + A_{j - 1}) \nonumber\\
    \quad j= 0, 1, \dots, \lfloor \frac{n}{2} \rfloor
\end{align}

We will go through this system case by case: 

\textbf{Case 1}. $w= 0$, $\beta  = 0$\\
This means for each sample only one element is supervised and the rest of elements are zero in that sample. The most basic task we can consider in this setup is one where the output of each position is solely dependent on its own input. So \eqref{diff} simplifies to:
\begin{equation}\label{wzero}
    \frac{d}{dt} A_j =  8   (\alpha - A_0) A_j
\end{equation}

Writing it for $j = 0$:
$$ \frac{d}{dt}A_0 = 8   (\alpha - A_0) A_0 $$

Which is a non-linear ODE with the following solution:
$$A_0^{(t)} = \frac{\alpha}{1+ (\frac{\alpha}{A_0^{(0)}}-1) \exp{(-8   \alpha t)}} $$

$A_0$ converges monotonically to $\alpha$, either from below or above depending on the values of $\alpha$. If $\alpha > A_0^{(0)}$, which holds with high probability if we choose $\alpha > 1$ and initialize $p_k$ with an isotropic Gaussian vector \cite{Vershynin_2018}, then $a - A_0^{(t)}$ is always greater than 0, and according to equation \eqref{wzero}:
$$\begin{cases}
    \frac{d}{dt}A_j > 0 \quad &A_j >0,\\
    \frac{d}{dt}A_j < 0 \quad &A_j <0.\\
\end{cases}$$

Absolute value of all off diagonal elements increases over iterations. So if at initialization $A_j = \theta(\frac{1}{\sqrt{d}})$, the final test loss for position $i$ is:
\begin{align}
    \E\{\ell_i\} = & \E \{ \big(\sum_{j = 1}^n (p_i^T p_j) s_j -  \alpha s_j \big)^2\} = \E \{ \big(\sum_{j = 1}^n (A_{i-j}) s_j -  \alpha s_j \big)^2\} = \E \{ \big(\sum_{j = 1, j \neq i}^n (A_{i-j}) s_j \big)^2\} \nonumber \\
    = & \sum_{j = 1}^{n-1} |A_r|^2  = \Omega (\frac{n}{d}) > 0 
\end{align}
\end{proof}
%\textbf{Case 2}. $w > 0$, $\beta  \neq 0$\\

To sum up, the use of APE in a model does not enable generalization, even when all samples in the training domain are augmented.

\subsection{Proof of \Cref{alltheory}: RPE}\label{RPE_proof}
We consider the same task as in \eqref{regression_y_x}, and investigate a factored attention model similar to \eqref{factored}, with the difference that positional vectors are relatively structured according to RPE. Note that the positional vectors are applied only to the key (and this is consistent with the models considered in \eqref{pairwise}). The query does not include positional information and must be treated as a constant vector in this scenario. We denote this constant vector by $u$. The model thus becomes \footnote{Note that if we had used $p_{i-r}$ instead of $u$, it would have made no difference in our as the effective parameter is the same. This becomes clear later on.}: 

\begin{equation}\label{factored_rel}
    \hat{y}_i = h^T W_V \sum_{r= 1}^n x_r (p_{i-r}^T W_K^T W_Q u)
\end{equation}
The formula presented here involves indices of $p$ from $-(n - 1)$ to $(n - 1)$. Only a subset of these indices are used at each position. For example, for position 1, only $p_0$ to $p_{-n_1 + 1}$ are considered. Absorbing the key and query matrices into p and u, and letting $v = W_V^T h$, we obtain the following equation:
\begin{equation}\label{linear_rpe}
    \hat{y}_i =  \sum_{r= 1}^n (p_{i-r}^T u) \langle v, x_r \rangle
\end{equation}

Consequently, we introduce a set of scalars that represent the model's parameters effectively: $a_{i-r} = p_{i-r}^T u$. The total loss function is:
\begin{align}\label{linear_rpe_opt}
    \ell(a, v) = &\sum_{i =1}^{n_1} (\hat{y}_i - y_i)^2
            = \sum_{i =1}^{n_1} \big(\sum_{r=1}^{n_1} a_{i-r} \langle v, x_r \rangle \nonumber\\
            &- \alpha \langle \theta, x_{i} \rangle - \beta \langle \theta, x_{i-1} \rangle -\beta \langle \theta, x_{i+1} \rangle \big)^2
\end{align}

At the first step, we prove that $v$ gets aligned with $\theta$ in spite of convergence for positional vectors:

\begin{lemma}
    For any initialization of v and any value for $p$ in equation \ref{linear_rpe}, running SGD with small enough step-size and infinitesimal $\epsilon$ on optimization problem \eqref{linear_rpe_opt} will align $v$ to $\theta$. 
\end{lemma}
\begin{proof}
    Taking the gradient of loss function in \eqref{linear_rpe_opt} with respect to $v$:
    \begin{align*}
    \nabla_v = & \sum_{i =1}^{n_1} 2 \big(\sum_{r=1}^{n_1} a_{i-r} v^T x_r  - \alpha \theta^T x_i - \beta \theta^T x_{i-1}  \mathbbm{1}(i \neq 1) - \beta \theta^T x_{i+1}  \mathbbm{1}(i \neq n_1) \big) \\
    & \quad \times \big( \sum_{r=1}^{n_1} a_{i-r} x_r \big)
    \end{align*}
    
    And in expectation:
    \begin{align*}
    \E\{\nabla_v\}= & \sum_{i =1}^{n_1} 2 \big(\sum_{r=1}^{n_1} a_{i-r}^2 v - \alpha a_0 \theta - \beta a_{1} \theta  \mathbbm{1}(i \neq 1) - \beta a_{-1} \theta \mathbbm{1}(i \neq n_1) \big) \\
    = &  (2 \sum_{i =1}^{n_1} \sum_{r=1}^{n_1} a_{i-r}^2) v - (2 \alpha n_1 a_0 + 2 \beta (n_1 -1) (a_{1} + a_{-1})) \theta
    \end{align*}
    
    Decomposing $v$ to one aligned vector with $\theta$, and another orthogonal element: 
    $$v = v_{\parallel} + v_{\perp}$$
    
    If the step-size is $\eta$ and weight-decay equal to $\epsilon$, then the updating rule for $v_{\perp}$ will be:
    $$v_{\perp}^{(t+1)} = v_{\perp}^{(t)} - \eta (2 \sum_{i =1}^{n_1} \sum_{r=1}^{n_1} a_{i-r}^2) v_{\perp}^{(t)} - \epsilon v_{\perp}^{(t)}$$
    $\epsilon$ is sufficiently small, if we take $\eta$ small enough such that $\eta (2 \sum_{i =1}^{n_1} \sum_{r=1}^{n_1} a_{i-r}^2)$ is smaller than 1 at all times, the orthogonal term vanishes over time. 
\end{proof}

% And even without augmentation this will accomplish a better generalization both for $i \in \{1, \dots, n_1\}$ and $i \in \{n_1 + 1, \dots, n\}$. 

After demonstrating that v becomes aligned, we will move its norm into $a_i$'s, resulting in $\ v\  = 1$ and $v = \theta$. It is time to concentrate on positional encodings:

\begin{unumproposition}[Formal statement of \Cref{alltheory}-- Third part]
The solution found by the gradient descent with infinitesimal weight-decay in the population regime on the loss function \eqref{linear_rpe_opt} will align $v$ with $\theta$ and satisfies $\langle v, \theta \rangle a_0= \alpha, \langle v, \theta \rangle a_{1} = \langle v, \theta \rangle a_{-1} = \beta$, and $a_j = 0$ for all $j \not \in \{-1, 0, 1\}$, and thus $ \E \{\ell_i\} = \E \{(\hat{y}_i - y_i)^2 \} = 0$ when $x_i \stackrel{\rm{i.i.d.}}{\sim} \calN(0, \mathbb{I}_d) \ \forall i \in [n]$.
\end{unumproposition}

\begin{proof}
Let's compute the gradient of \eqref{linear_rpe_opt} with respect to a's. Letting $s_r= \theta^T x_r$, which is a normal random variable:
\begin{align*}
    \nabla_{a_k} = & \sum_{i =1}^{n_1} 2 \big(\sum_{r=1}^{n_1} a_{i-r} s_r  - \alpha s_i - \beta s_{i-1}  \mathbbm{1}(i \neq 1) - \beta s_{i+1}  \mathbbm{1}(i \neq n_1) \big) \\
    & \quad \times \big(\mathbbm{1}(k \in [i-1, i-n_1] s_{i-k})
\end{align*}

So that in expectation:
\begin{align*}
    \E \{\nabla_{a_k} \} = & \sum_{i =1}^{n_1} 2 \mathbbm{1}(k \in [i-1, i-n_1]) \big( a_{k}  - \alpha \mathbbm{1}(k = 0) - \beta \mathbbm{1}(i \neq 1) \mathbbm{1}(k= 1)\\
    & \hspace{1.7in} - \beta \mathbbm{1}(i \neq n_1) \mathbbm{1}(k=-1) \big) \\
    = & 2 (n_1 - |k|)a_k - 2 n_1 \alpha \mathbbm{1}(k=0) - 2 (n_1 - 1) \beta \mathbbm{1}(k=1) - 2 (n_1 - 1) \beta \mathbbm{1}(k= -1)
\end{align*}

Where in the last line to calculate the sum, we use the fact that $a_k$ is observed $n_1 -|k|$ times when traversing from $1$ to $n_1$. For values of $k$ outside the range $[-n_1, n_1]$, the gradient is zero. Hence with weight-decay they will diminish over time. For values of $k$ between $-n_1$ and $n_1$, excluding -1, 0, or 1, the update equation is given by:
$$a_k^{(t+1)} = a_k^{(t)} - 2 \eta (n_1 - |k|) a_k^{(t)} - \epsilon a_k^{(t)}$$
An again, for small values for the step-size and weight-decay they converge to zero: $$a_k \to_t \bzero \quad \forall k \notin [-1, 1]$$

The other three variable go on a different route:
$$\begin{cases}
    a_0^{(t+1)} = a_0^{(t)} - 2 \eta n_1 (a_0^{(t)} - \alpha) - \epsilon a_0^{(t)}, \\
    a_1^{(t+1)} = a_1^{(t)} - 2 \eta (n_1-1) (a_1^{(t)} - \beta) - \epsilon a_1^{(t)}, \\
    a_{-1}^{(t+1)} = a_{-1}^{(t)} - 2 \eta (n_1-1) (a_{-1}^{(t)} - \beta) - \epsilon a_{-1}^{(t)}.
\end{cases}$$
For infinitesimal values of weigh-decay and small step-size, they accordingly converge to $\alpha$, $\beta$ and $\beta$. So at test time:

\begin{align}
    \E\{\ell_i\} = & \E \{ \big(\sum_{j = 1}^n (a_{i-j}) s_j -  \alpha s_j - \beta s_{i-1} - \alpha \beta s_{i+1} \big)^2\} = 0
\end{align}

Using RPE the model will be able to achieve zero test loss. 
\end{proof}

% \subsection{Experiments for linear attention}
% To further investigate our claims in the theory part, we simulated the factored attention model both for its absolute form in \eqref{}
\subsection{Experiments for the linear attention model}\label{app:lin_att}

\paragraph{Experiments for a more general setting.} We conducted another experiment for a symmetric task more generalized than \Cref{regression_y_x} (here the dependency length is $m$ instead of 1), establishing that out claims hold beyond the scope of our assumptions:

\begin{equation*}
    y_i = \alpha \langle \theta, x_i \rangle + \beta_1 \langle \theta, x_{i-1} \rangle + \beta_1 \langle \theta, x_{i+1} \rangle + \dots +  \beta_m\langle \theta, x_{i-m} \rangle + \beta_m \langle \theta, x_{i+m} \rangle
\end{equation*}

\begin{figure}[t!]
     \centering
     \begin{subfigure}[b]{0.38\textwidth}
         \centering
         % Linewidth is the now the unit for half a page.
         \includegraphics[width=1.0\linewidth]{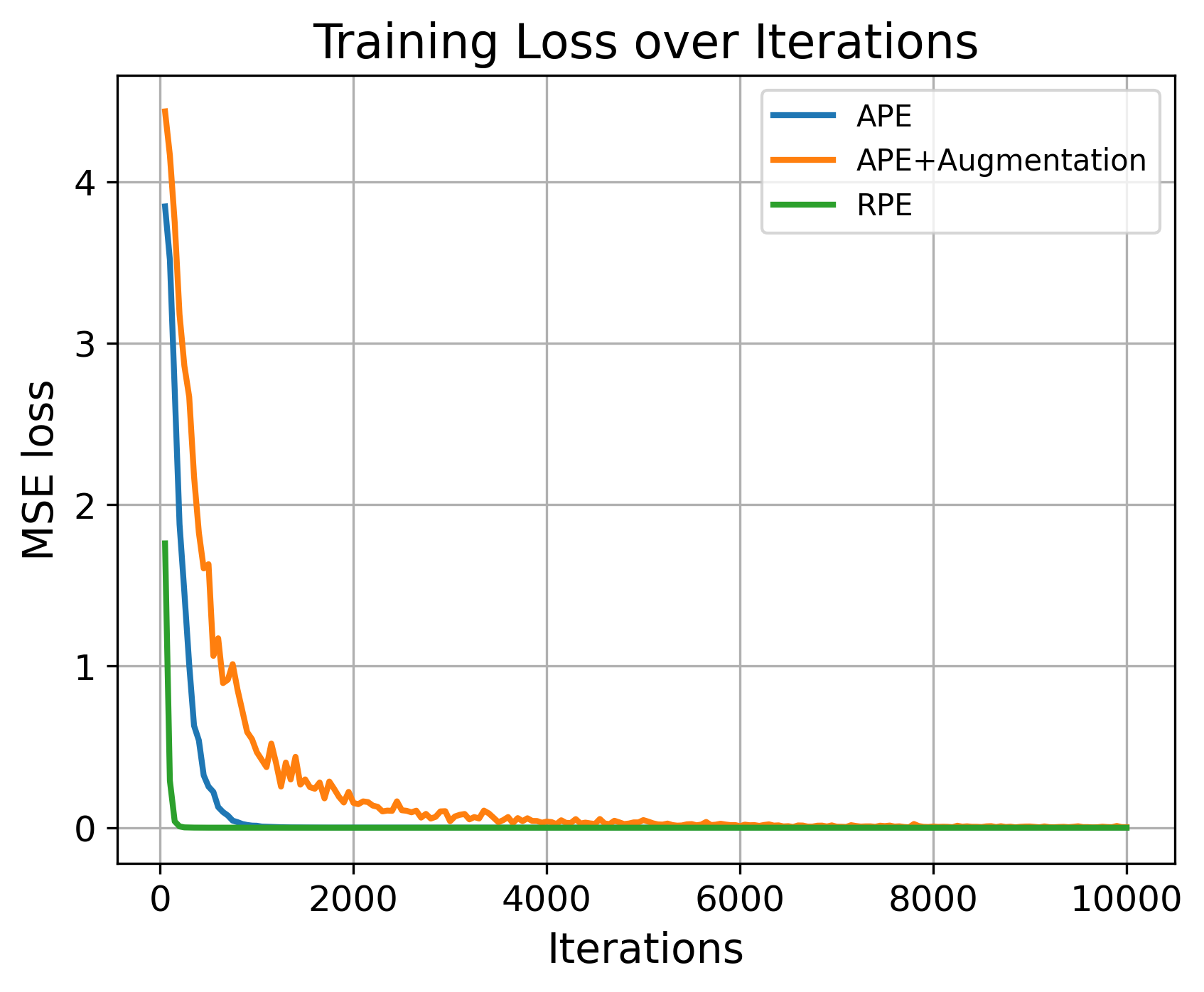}
         \caption{}
         \label{Loss_iter}
     \end{subfigure}
    \hspace{0.5in}
     \begin{subfigure}[b]{0.38\textwidth}
         \centering
         % Linewidth is the now the unit for half a page.
         \includegraphics[width=1.0\linewidth]{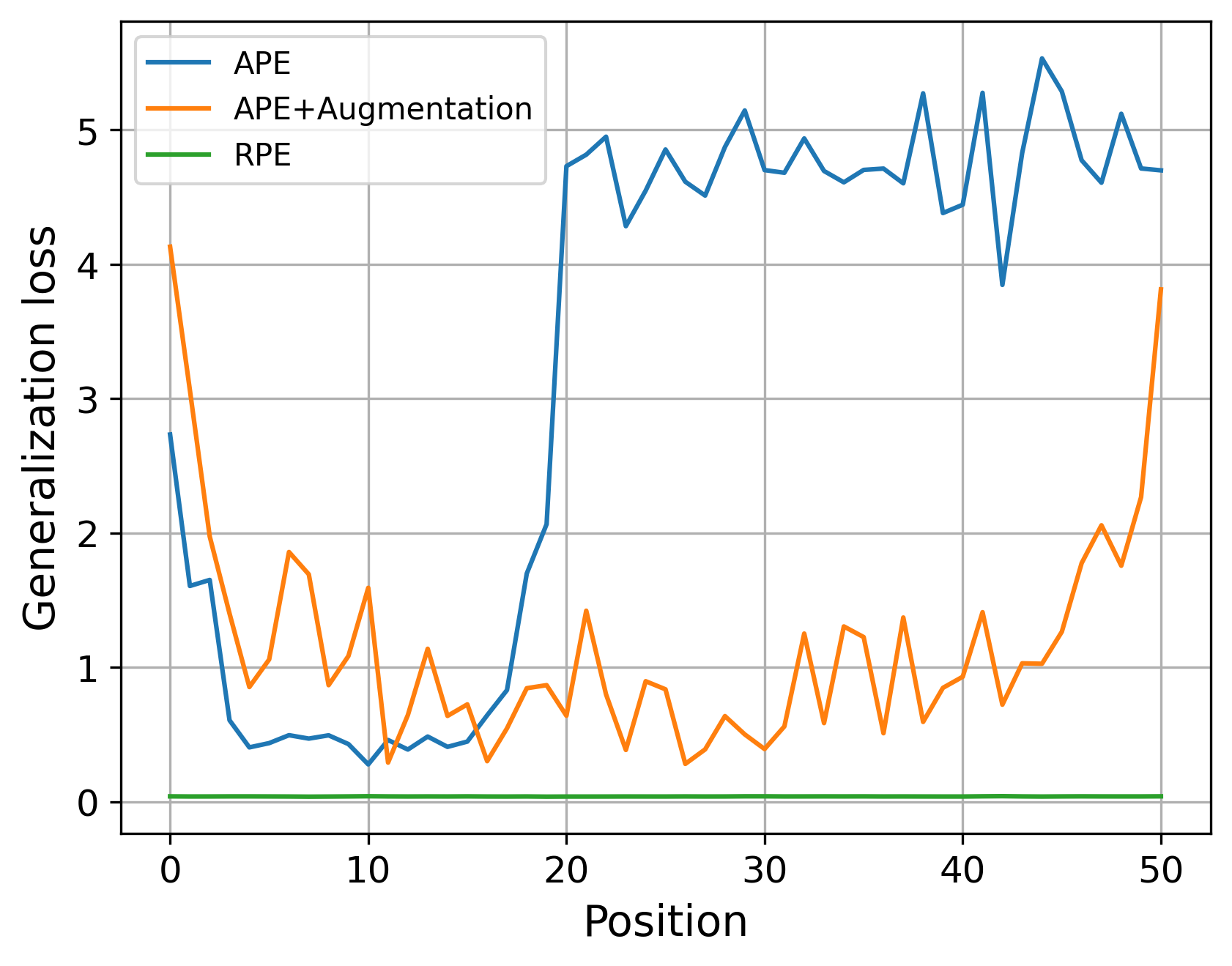}
         \caption{}
         \label{Gener_pos}
     \end{subfigure}
        \caption{\textbf{(a)} Similarly to \Cref{Gener_pos_m1}, all three scenarios reach zero in the training loss. At test time, all position get non-padded values. \textbf{(b)} The generalization loss at each position in the sequence. Again, the model with RPE greatly outperforms the other two.}
\end{figure}

\begin{figure}[t!]
     \centering
     \begin{subfigure}[b]{0.365\textwidth}
         \centering
         % Linewidth is the now the unit for half a page.
         \includegraphics[width=1.0\linewidth]{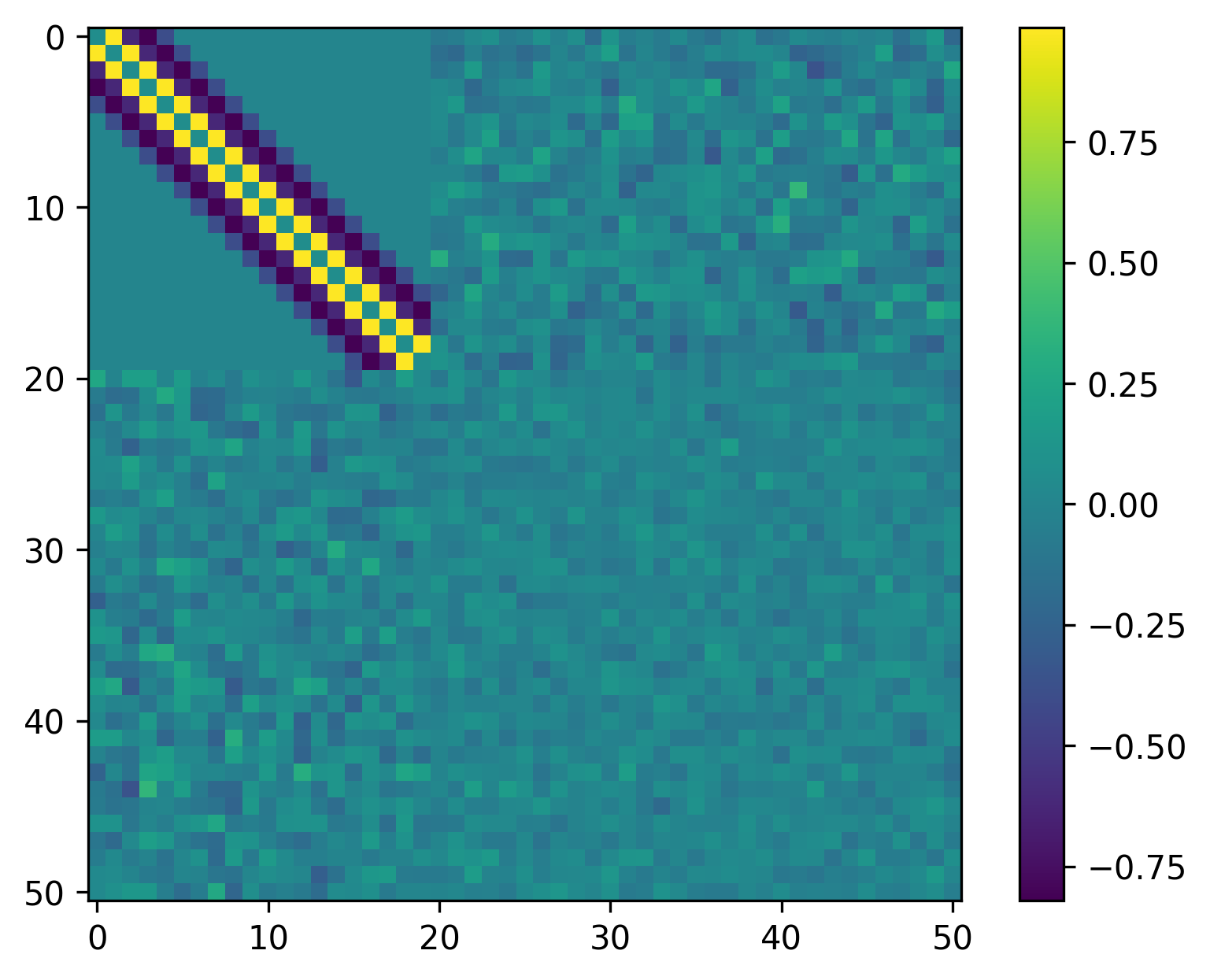}
         \caption{}
         \label{P_ape}
     \end{subfigure} \hspace{0.05\textwidth}
     \begin{subfigure}[b]{0.3\textwidth}
         \centering
         % Linewidth is the now the unit for half a page.
         \includegraphics[width=1.0\linewidth]{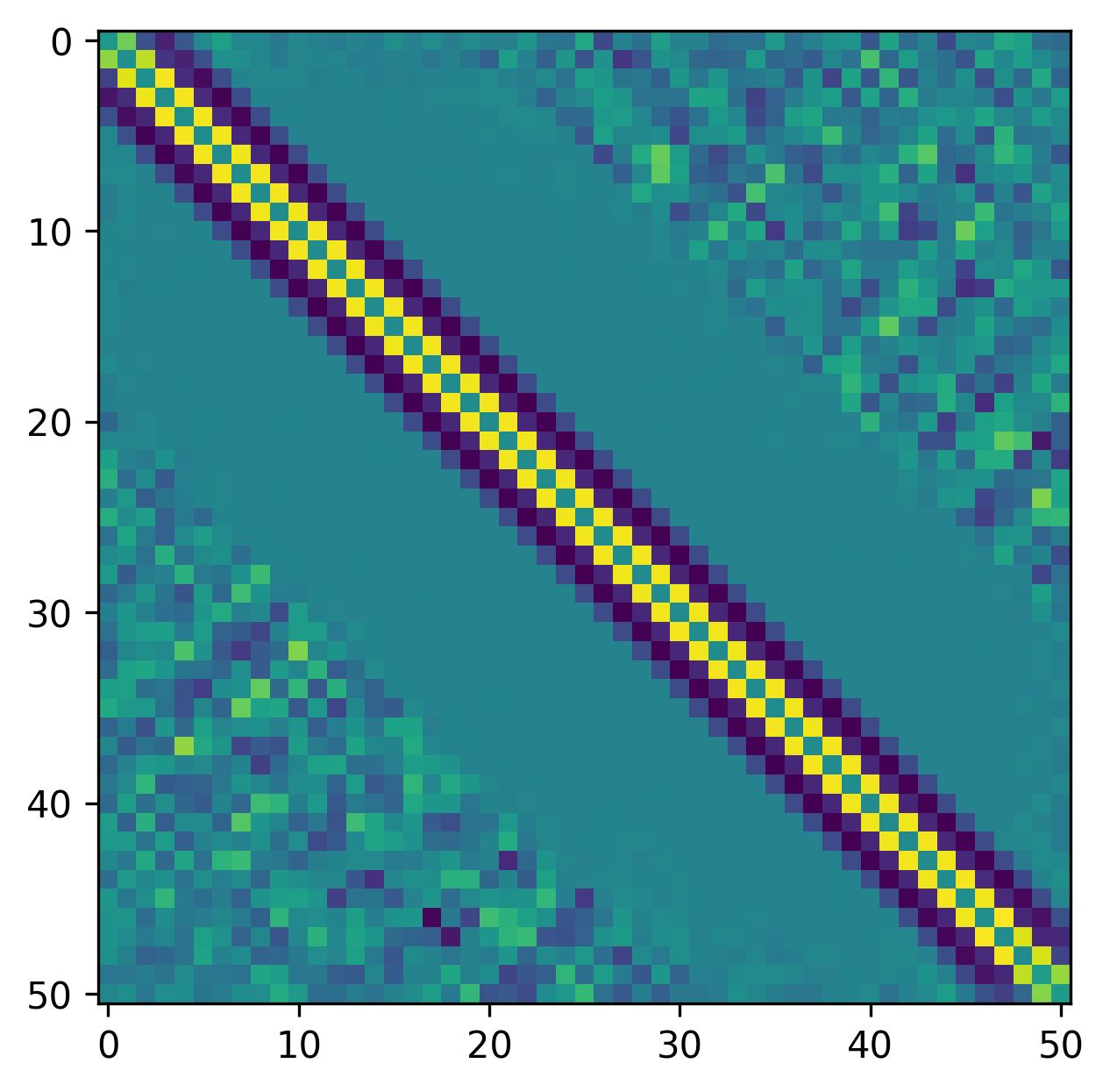}
         \caption{}
         \label{P_aug}
     \end{subfigure}
     \\
      \begin{subfigure}[b]{0.5\textwidth}
         \centering
         % Linewidth is the now the unit for half a page.
         \includegraphics[width=1.0\linewidth]{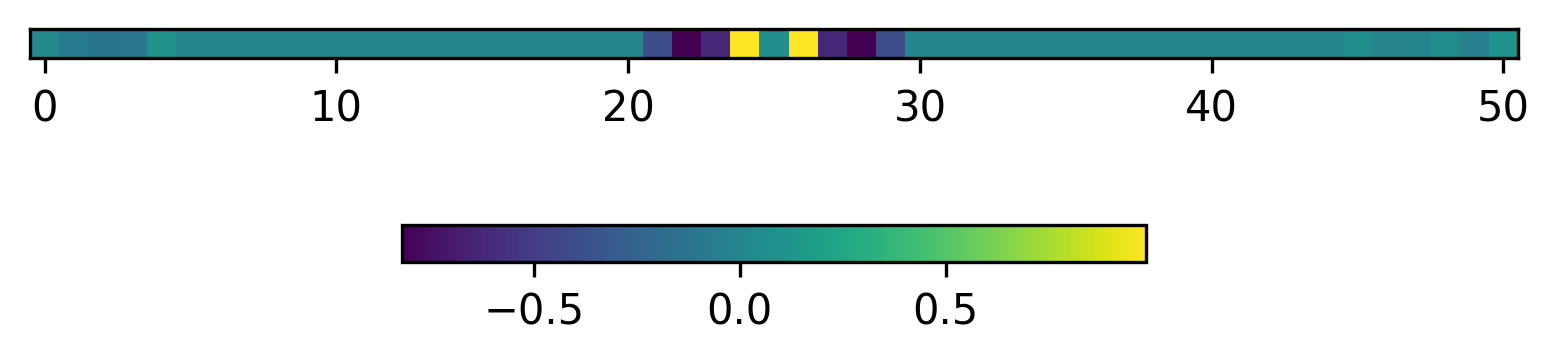}
         \caption{}
         \label{P_rpe}
     \end{subfigure}
        \caption{\textbf{(a)} The set of effective parameters are $[p_i^T W^T W p_j]_{i \leq n, j \leq n}$. The model with APE does not learn the structure for $i, j > n_1$. \textbf{(b)} Even though augmentation helps to learn the main structure of the task every where, the two off-diagonal sides remain nonzero and damage the generalization. \textbf{(c)} Shows $p_{i-j}^T W_K^T W_Q u$ where the index on this graph represents $i-j + (n-1)/2$. Note that $\frac{1}{2}$ factor appears due to the ring condition, e.g. $i = 1, j = n$ will have $p_{-1}$ associated to it.}
\end{figure}

For the result of \Cref{Loss_iter}, we set $n= 51$, $n_1 =20$, $m = 5$, and $d= 200$. As previously seen in \Cref{Gener_pos_m1}, although APE and APE with augmentation achieves zero training loss, they fails to generalize when all positions are non-padded. The causes of their failure are explained in \Cref{APE_proof} and \Cref{Aug_proof}. We have plotted $[p_i^T W^T W p_j]_{i \leq n, j \leq n}$ for APE and APE with augmentation in \Cref{P_ape,P_aug} accordingly. For $i, j>n_1$, APE does not learn the structure outside the focus window as described in \Cref{APE_proof}. But augmentation fails for a different reason, and \Cref{P_aug} shows that near diagonal elements of $[p_i^T W^T W p_j]$ are learned properly while the rest move in random directions and ruin the generalization.  

\Cref{Gener_pos} illustrates that we can address this issue with the help of RPE, here unwanted elements that keep us from generalizing in APE have been eliminated automatically.

\section{Fixed map for RPE}\label{fixed_map}
If the maximum complexity across the training dataset is $2 * d$ (Note that $d - 1$ stands for the maximum number of levels illustrated in \Cref{sum_example} among all training samples), the trained model will ignore those previous positions of both integers that do not contribute the current calculation. i.e., with our notation defined in \Cref{Section: parallel} for the tracked list of positions: 

$$\sigma_{\rm training}(i) = \big(i - d, \cdots, i - 1, i, i + l - d+1, \cdots, i + l, i + l + 1\big)$$

Thus, for a model utilized with RPE, the previous list will be memorized in terms of relative positions: 

$$\sigma_{\rm training}^{\rm rel}(i) = \big(-d,  \cdots, -1, 0, l -d+1, \cdots, l, l + 1 \big)$$

As a result, $\sigma_{\rm training}^{\rm rel}(i)$ is the same list for every $i$. Hence, if the model equipped with RPE learns to place its attention on correct relative positions for in-distribution samples, it will naturally be able to generalize to longer integers as long as the complexity of the samples is limited. 
% \newpage

\end{document}